\documentclass{article}

\usepackage{arxiv}

\usepackage[utf8]{inputenc} % allow utf-8 input
\usepackage[T1]{fontenc}    % use 8-bit T1 fonts
\usepackage{hyperref}       % hyperlinks
\usepackage{url}            % simple URL typesetting
\usepackage{booktabs}       % professional-quality tables
\usepackage{amsfonts}       % blackboard math symbols
\usepackage{nicefrac}       % compact symbols for 1/2, etc.
\usepackage{microtype}      % microtypography
\usepackage{lipsum}

\usepackage{graphicx}
\usepackage{subcaption}

\title{A New Deterministic Technique for Symbolic Regression}

\author{
  Daniel Rivero \\
  Centro de investigación CITIC \\
Department of Computer Science and Information Technology\\
 University of A Coruña\\
 A Coruña, Spain \\
  \texttt{daniel.rivero@udc.es} \\
   \And
 Enrique Fernandez-Blanco \\
 Centro de investigación CITIC \\
 Department of Computer Science and Information Technology\\
 University of A Coruña\\
 A Coruña, Spain \\
  \texttt{enrique.fernandez@udc.es} \\
}

\begin{document}
\maketitle

\begin{abstract}
Based on a solid mathematical background, this paper proposes a new method for Symbolic Regression which allows finding mathematical expressions from a dataset. Oppositely  to other methods, such as Genetic Programming, this is deterministic approach which does not require the creation of a population of initial solutions. Instead of it, a simple expression is being grown until it fits the data. The experiments performed show that the results are as good as other Machine Learning methods, in a very low computational time. Another advantage of this technique is that the complexity of the expressions can be limited, so the system can return mathematical expressions that can be easily analyzed by the user, in opposition to other techniques like GSGP.
\end{abstract}

% keywords can be removed
\keywords{Symbolic Regression \and Machine Learning \and Artificial Intelligence}

%%%%%%%%%%%%%%%%%%%%%%%%%%%%%%%%%%%%%%%%%%%%%%%%%%%%%%%%%%%%%%%%%%%%%%%%%%%%%%%%

\section{Introduction}

In Machine Learning, supervised learning allows to find models which represent the relationship between a series of inputs and outputs. Nowadays, there are many different techniques for finding these models such as Artificial Neural Networks \cite{haykin2009neural}. However, many of these methods, even they show good results in the modelling, do not give hints about the true relationship between inputs and outputs. In many environments, a black-box model is not enough, since the objective is to find an equation that the expert can analyse and thus increase the knowledge about the system being modelled. In this sense, the search of transparent ML models, showing a clear relationship between inputs and outputs is today a research hot topic, and nowadays there are conferences such as FAT (ACM Conference on Fairness, Accountability and Transparency) devoted to this field.

This work proposes a new method for symbolic regression: from a dataset (inputs/outputs), this method allows to find mathematical expressions that can reproduce this relationship. These expressions will be of benefit to many experts to understand the behaviour of a system.

Nowadays, there are few different methods for symbolic regression. The most used ones are based on Genetic Programming (GP) \cite{Koza:1992:GPP:138936} \cite{Poli:2008:FGG:1796422} which is a general-purpose evolutionary technique. The first and most common representation in GP is tree-shaped. However, recently other different representations have arisen, such as Linear GP \cite{Brameier2007}, Stack-based Genetic Programming \cite{StackGP}, Cartesian Genetic Programming (CGP) \cite{Miller:2015:CGP:2739482.2756571}, or Positional Cartesian Genetic Programming (PCGP) \cite{DBLP:journals/corr/abs-1810-04119}. However, trees are still the most used codification for mathematical expressions for symbolic regression tasks.

GP works from an initial population of trees, that undergo an evolutionary process with the execution of selection, crossover, mutation and replacement operators. These operators are based on randomness, and this makes the whole process a time-consuming task. For instance, the original crossover operator proposes the random combination of mathematical expressions, although there are different approaches that try to combine useful parts \cite{10.1007/978-3-642-01181-8_20}\cite{10.1007/978-3-642-29139-5_6}. On its side, mutation operator also makes random changes in a mathematical expression. All of this makes that although the evolutionary process is a search driven by the fitness function, it needs the computation of many mathematical expressions that are not going to be part of the final expression. Also, although the algorithms behind GP are well-known, its global behaviour is still to be studied, and the obtained results lack of a mathematical basis. However, as a system capable of performing symbolic regression, it has been successfully applied to real-world problems such as integrated circuit design \cite{5166638} or civil engineering \cite{GPcivil}.

Recently, a new type of GP, called Geometric Semantic Genetic Programming (GSGP) has arisen \cite{10.1007/978-3-642-32937-1_3}. This approach works in the so-called geometric space, which is conformed by the outputs of the GSGP programs (semantics). In this space, the targets are another point in the semantic space. Its performance, measured in results and time, is much higher than GP. However, a big drawback of this technique is that its result is a tree with an excessive number of nodes, usually higher than $10^{15}$ \cite{GSGPoverkill} Although there are works that try to reduce this number \cite{DBLP:journals/corr/abs-1804-06808}, the resulting program are still very complex, many times as a sum of different mathematical expressions. On another hand, it was demonstrated that the resulting expression that the final expression it is a sum on the expressions generated on the first generation \cite{GSGPoverkill}. This fact, having as result oversized and very complex expressions, has become a big drawback of this technique since the evolved expressions are not understandable by a human being. However, it has been successfully applied in many real-world environments such as financial \cite{GSGPfinancial} or biomedical \cite{6661969}.

Symbolic regression is a research field that was hardly been explored outside of GP literature, with very few works describing non-evolutionary approaches \cite{FFX}\cite{GSGPoverkill}. However, although their computation time is very low and they show good training results, they are based on the combination of different functions that work as a basis on a $L_2$ space. Thus, the resulting expression is the weighted sum of a series of expressions. As was already said, this was also demonstrated to happen in GSGP \cite{GSGPoverkill}. This leads again to having a system that returns expressions that are not understandable by a human.

Therefore, there is still the need of one technique that can generate simple mathematical expressions, easily understandable by the human, with a mathematical basis, that works in short time. This work presents a technique with all of these features. An additional advantage of this technique is that it is deterministic.

This work is not based on GP or GSGP. The only similarities between this work and GP is that the expression has shape of a tree as in traditional GP, and that we work in the semantic space as in GSGP. As opposed to GP or GSGP, the system proposed in this paper does not need to generate (and evaluate) a population of expressions. Instead of it, a single expression is being sequentially improved, and each change performed to it is guaranteed to improve its performance.

%%%%%%%%%%%%%%%%%%%%%%%%%%%%%%%%%%%%%%%%%%%%%%%%%%%%%%%%%%%%%%%%%%%%%%%%%%%%%%%%

\section{Model}

\label{sec:Model}

Usually when working in Machine Learning, the user has a dataset arranged as a matrix with dimensions NxL or LxN, being L the number of variables or features, and N the number of patters. In supervised learning, a target matrix is also needed, with dimension of NxT or TxN, accordingly with the previous, being T the number of outputs. In the case of symbolic regression, since a single equation is desired, T=1. The most common way of working with this is creating a L-dimensional space, in which each dimension corresponds to each variable. Thus, each pattern is a point in that L-dimensional space.

However, in this paper we work with a N-dimensional space, with one dimension for each pattern. Therefore, each variable corresponds to a single point in this space, and the targets are also a point in this space. Moreover, the output of a model (not limited to being a equation) gives one value for each pattern, thus making a vector in this space. Therefore, each model is represented also as a point in this space. In GSGP, this space is called "semantic space", because the output of each model is called the semantic.

Similarly to GSGP, a model has a semantic, i.e., a vector in the semantic space composed by the outputs $o_i$ for each pattern. Since the targets $t_i$ are also a point in the semantic space, the Euclidean distance between these two points can be measured with equation \ref{eq:eq1}.

\begin{equation}
\label{eq:eq1}
distance = \sqrt{\sum_{i=1}^N{(o_i-t_i)^2}}
\end{equation} 

This equation corresponds to the square root of the SSE (Sum Squared Error) of the model. Thus, finding a better model (i.e., a model with a lower SSE) is equivalent to finding a model closer to the target point. Therefore, a N-dimensional sphere is being created, and any model inside this sphere will be a better model (i.e., with a lower SSE). This sphere has as radius the square root of the SSE (Root Sum Squared Error, RSSE), with the following equation:

\begin{equation}
\label{eq:eq2}
\sum_{i=1}^N{(o_i-t_i)^2} < SSE
\end{equation} 

Also, in order to minimize the impact of having a large or low number of patterns, this number N is usually inserted into the equation, having as result the Mean Square Error equation to minimize, shown in equation \ref{eq:eq3}

\begin{equation}
\label{eq:eq3}
\frac{1}{N}\sum_{i=1}^N{(o_i-t_i)^2} < MSE
\end{equation} 

Again, a new model with outputs $o_i$ that complies with the restriction of equation \ref{eq:eq3} is a model with a lower MSE, and closer to the target. If the model being processed can undergo different improvements, the best improvement will be the one that makes the model closer to the target point. This is the improvement that makes the new outputs maximize the reduction in MSE, given by equation \ref{eq:eq4}

\begin{equation}
\label{eq:eq4}
reduction = MSE - \frac{1}{N}\sum_{i=1}^N{(o_i-t_i)^2}
\end{equation} 

This turns the problem of improving the model into an optimization problem: the more positive the result of equation \ref{eq:eq4} is, the better the performance of the new model will be. If different improvements are possible, then the one selected will be the one that maximizes equation \ref{eq:eq4}. If none of these possible improvements lead to having positive values in equation \ref{eq:eq4}, then the improvement process has finished.

\begin{figure}
    \centerline{\includegraphics[width=6cm, height=6cm]{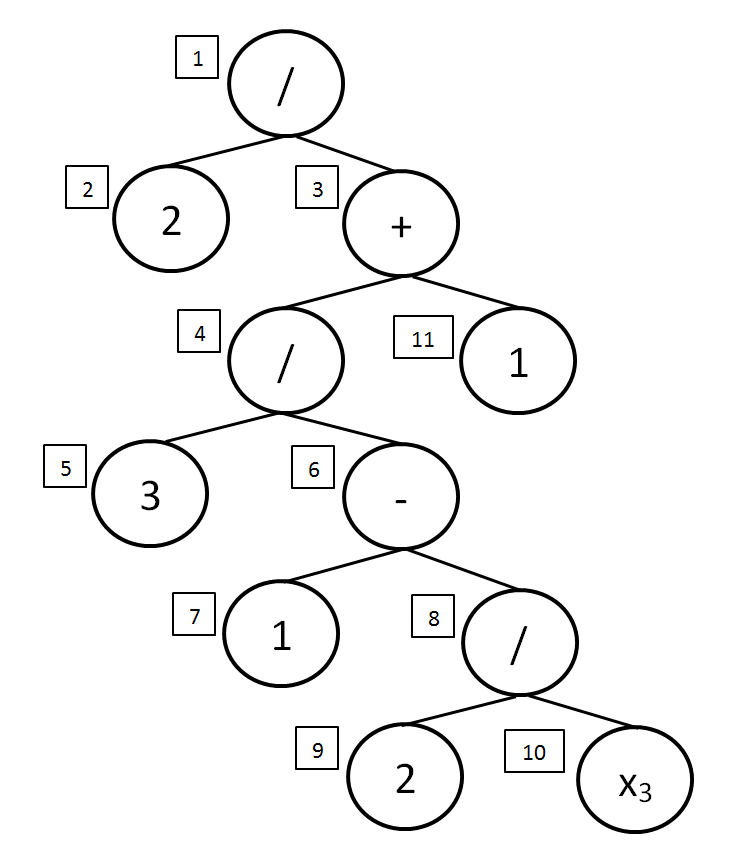}}
    \caption{Example of a tree}
    \label{fig:figExampleTree}
\end{figure}

As was previously stated, the equation being developed has the shape of a tree. As in GP, we will distinguish two types of nodes: terminal, or leaves of the tree, and non-terminal, or functions. As terminal nodes we use only those variables of the problem, and constants. As functions we use the four arithmetic operators: +,-,*,/. Note that we do not use "\%" as protected division. In GP it was necessary to protect that operation because, as a result of many different combinations, chances are that some divisions by zero will be performed. In this case, the tree being developed will always be correct, with no divisions by 0.

Figure \ref{fig:figExampleTree} shows an example of tree. This tree has 11 nodes, 4 of them are non-terminal (labeled as 1, 3, 4 and 8), and 5 are terminal (labeled as 2, 4, 5, 7, 9 and 11). This tree represents the following expression:

\begin{equation}
\label{eq:eqExample}
\frac{2}{\frac{3}{1-\frac{2}{x_3}}+1}
\end{equation}

The tree being developed will have a semantic determined by the outputs to each pattern. As has already been said, this tree can be a terminal or non-terminal. In the second case, the root of the tree will be any of the four arithmetic operators, and each of its children will be another tree, with their corresponding semantic and their corresponding points in the semantic space. Therefore, a tree with n nodes is represented in the semantic space as the semantic point of the root of the tree, but also as n-1 different points. If any of these nodes is modified, then its semantic point will be moved. This has the consequence that the overall evaluation values of the root of the tree are changed and the semantic point of the root of the tree will be moved too. Note that constants and variables (terminal nodes of the tree) also have a semantic value, representing one point in the semantic space.

Table \ref{tableExampleTree} shows, in the second column, the semantics of each node of the tree shown in fig. \ref{fig:figExampleTree}. All of the nodes, including constant and variables have their semantic, i.e., they are points in the search space. The variable $x_3$ takes the values of 1 for the first pattern, 4 for the second and -2 for the third. Note that each terminal node representing a constant $k$ has a semantic $(k,k,...,k)$, since it evaluates to $k$ for each data point.

\begin{table}
\centering
\caption{Description of the tree of the example}
{
    \begin{tabular}{@{}ccccc@{}} 
    \toprule
        Node & \multicolumn{1}{c}{Semantic} & \multicolumn{1}{c}{Equation} & \multicolumn{1}{c}{S} \\
    \midrule
        1 & (-1, 0.286, 0.8) & $\frac{1}{3}((o_i-5)^2+(o_i-4)^2+(o_i-1)^2)$ & $\emptyset$ \\
        2 & (2, 2, 2) & $\frac{1}{3}((\frac{o_i+10}{2})^2+(\frac{o_i-28}{-7})^2+(\frac{o_i-2.5}{-2.5})^2)$ & $\emptyset$ \\
        3 & (-2, 7, 2.5) & $\frac{1}{3}((\frac{-5\cdot o_i+2}{o_i})^2+(\frac{-4\cdot o_i+2}{o_i})^2+(\frac{-o_i+2}{o_i})^2)$ & $\{(0,0,0)\}$ \\
        4 & (-3, 6, 1.5) & $\frac{1}{3}((\frac{-5\cdot o_i-3}{o_i+1})^2+(\frac{-4\cdot o_i-2}{o_i+1})^2+(\frac{-o_i+1}{o_i+1})^2)$ & $\{(-1,-1,-1)\}$ \\
        5 & (3, 3, 3) & $\frac{1}{3}((\frac{-5\cdot o_i+3}{o_i-1})^2+(\frac{-4\cdot o_i-1}{o_i+0.5})^2+(\frac{-o_i+2}{o_i+2})^2)$ & $\{(1,-0.5,-2)\}$ \\
        6 & (-1, 0.5, 2) & $\frac{1}{3}((\frac{-3\cdot o_i-15}{o_i+3})^2+(\frac{-2\cdot o_i-12}{o_i+3})^2+(\frac{o_i-3}{o_i+3})^2)$ & $\{(0,0,0), (-3,-3,-3)\}$ \\
        7 & (1,1,1) & $\frac{1}{3}((\frac{-3\cdot o_i-9}{o_i+1})^2+(\frac{-2\cdot o_i-11}{o_i+2.5})^2+(\frac{o_i-2}{o_i+4})^2)$ & $\{(2,0.5,-1), (-1,-2.5,-4)\}$ \\
        8 & (2, 0.5, -1) & $\frac{1}{3}((\frac{3\cdot o_i-18}{-o_i+4})^2+(\frac{2\cdot o_i-14}{-o_i+4})^2+(\frac{-o_i-2}{-o_i+4})^2)$ & $\{(1,1,1), (4,4,4)\}$ \\
        9 & (2, 2, 2) & $\frac{1}{3}((\frac{3\cdot o_i-18}{-o_i+4})^2+(\frac{2\cdot o_i-56}{-o_i+16})^2+(\frac{-o_i+4}{-o_i-8})^2)$ & $\{(1,4,-2), (4,16,-8)\}$ \\
        10 & (1, 4, -2) & $\frac{1}{3}((\frac{-18\cdot o_i+6}{4\cdot o_i-2})^2+(\frac{-14\cdot o_i+4}{4\cdot o_i-2})^2+(\frac{-2\cdot o_i-2}{4\cdot o_i-2})^2)$ & $\{(0,0,0), (2,2,2), (0.5,0.5,0.5)\}$ \\
        11 & (1, 1, 1) & $\frac{1}{3}((\frac{-5\cdot o_i+17}{o_i-3})^2+(\frac{-4\cdot o_i-22}{o_i+6})^2+(\frac{-o_i+0.5}{o_i+1.5})^2)$ & $\{(3,-6,-1.5)\}$ \\
        \\
%    \bottomrule
    \end{tabular}
}
\label{tableExampleTree}
\end{table}

\begin{figure}
    \centerline{\includegraphics[width=6cm, height=4cm]{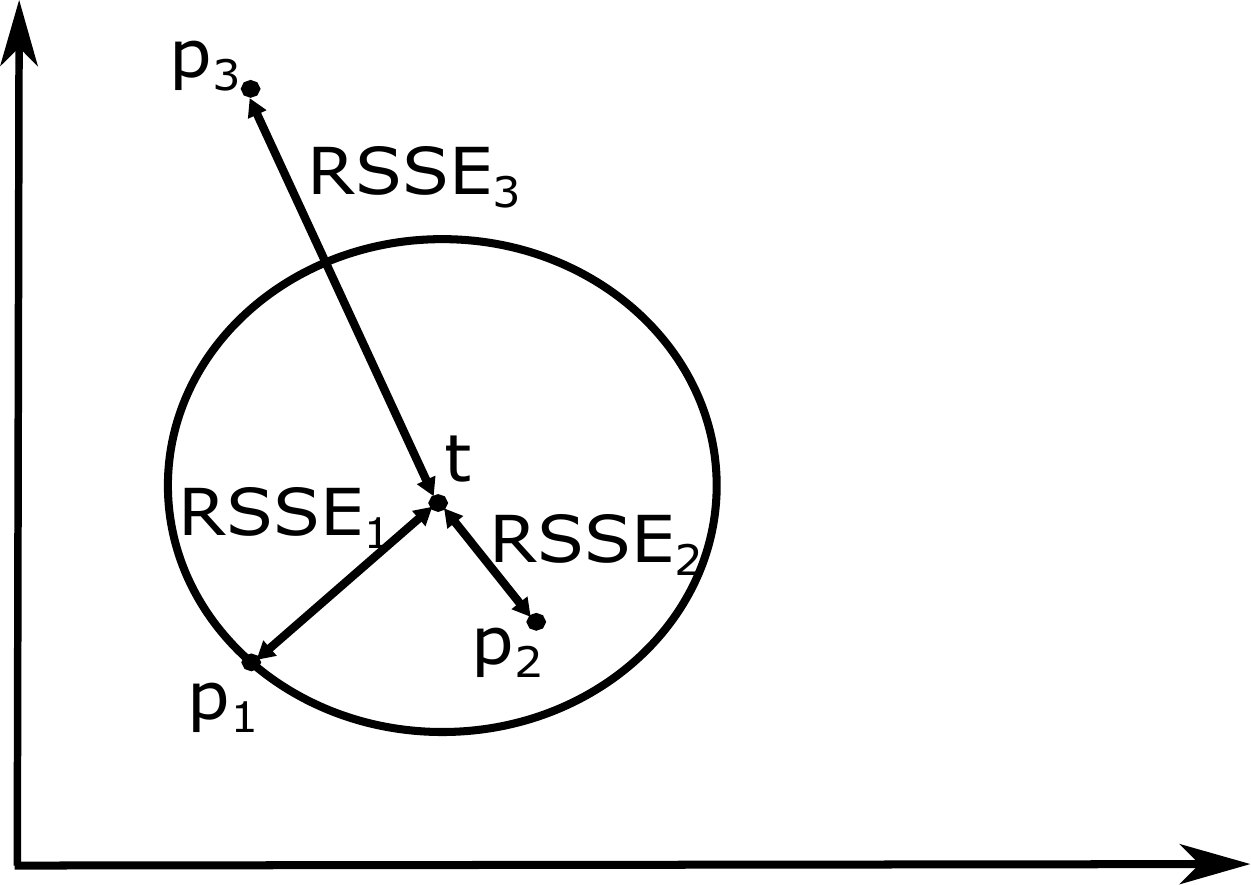}}
    \caption{Example of N-dimensional sphere generation}
    \label{fig:FigMSE}
\end{figure}

Fig \ref{fig:FigMSE} shows an example of an expression with a semantic represented by the point $p_1$, and the targets situated at the point t. The circle around the targets represent the different expressions that have the same RSSE as $p_1$. Thus, any expression inside this circle has a lower RSSE. In this case, the expression with a semantic represented by the point $p_2$ is inside the circle, so the expression $p_1$ can be replaced by $p_2$ with a reduction of the RSSE. Note that a lower RSSE leads to having a lower SSE, which also leads to having a lower MSE. This reduction can be calculated as described in equation \ref{eq:eq4}. This figure also shows another expression, $p_3$, that is outside the circle. This means that replacing $p_1$ for $p_3$ leads to having a higher RSSE, SSE and MSE. In this case, the calculation of the reduction returns a negative value. As has been already explained, only positive values in reduction lead to an improvement.

The key idea of this work is that we may not have a tree that makes a positive reduction in equation \ref{eq:eq4}, moving its semantic to get closer to the target point. However, it might be easier to modify one node in any branch in order to move the root towards the target. This modification is done by changing that subtree into another. The question is how to find the new subtree that will substitute it. In order to find it, it is necessary to calculate the MSE from the outputs of this node. In this sense, each node of the tree has associated an equation to calculate the MSE from its outputs. As happens with the root, this equation can be used to quantify the improvement of the overall result if the outputs $o_i$ of this node are modified. This equation represents a shape in the semantic space, and, for any node of the tree, has the following shape:

\begin{equation}
\label{eq:eq5}
MSE=\frac{1}{N}\sum_{i=1}^N{(\frac{a_i \cdot o_i-b_i}{c_i \cdot o_i-d_i})^2}
\end{equation}

where $o_i$ are the outputs of that node, and $a_i$, $b_i$, $c_i$ and $d_i$ are four vectors (i=1,...,N) characterizing the equation for this node. These vectors will be different for each node of the tree, thus having a different equation on each node. However, given a tree, the result of each equation of each node applied to its outputs will be the same MSE value. Therefore, a first step of this algorithm is to calculate these vectors for each node.

This process is recursively done from each non-terminal node to its child nodes: each non-terminal node, from the outputs of their child nodes (whether they are terminal or non-terminal) and its $a_i$, $b_i$, $c_i$ and $d_i$ vectors, calculates the vectors for each of its children, and then this process is repeated for each non-terminal children. For the root of the tree, $a_i=1$, $b_i=t_i$, $c_i=0$ and $d_i=-1$, leading to equation \ref{eq:eq3}. For the rest of the non-terminal nodes, the calculation of the vectors for each children is done in the following way:

\begin{itemize}

\item Sum operation. In this case, the output of the node is written as $o_i=x_i+y_i$, being $x_i$ and $y_i$ the outputs (semantics) of its two children. The equation for the first child becomes the following:

\begin{equation}
\label{eq:eqSumChild1}
\frac{1}{N}\sum_{i=1}^N{(\frac{a_i \cdot (x_i+y_i)-b_i}{c_i \cdot (x_i+y_i)-d_i})^2} = \frac{1}{N}\sum_{i=1}^N{(\frac{a_i \cdot x_i-(b_i-a_i \cdot y_i)}{c_i \cdot x_i-(d_i-c_i \cdot y_i)})^2}
\end{equation} 

which has the shape of equation \ref{eq:eq5} with $a_i'=a_i$, $b_i'=b_i-a_i\cdot y_i$, $c_i'=c_i$ and $d_i'=d_i-c_i \cdot y_i$. For the second child, the equation is very similar:

\begin{equation}
\label{eq:eqSumChild2}
\frac{1}{N}\sum_{i=1}^N{(\frac{a_i \cdot (x_i+y_i)-b_i}{c_i \cdot (x_i+y_i)-d_i})^2} = \frac{1}{N}\sum_{i=1}^N{(\frac{a_i \cdot y_i-(b_i-a_i \cdot x_i)}{c_i \cdot y_i-(d_i-c_i \cdot x_i)})^2}
\end{equation} 

which has the shape of equation \ref{eq:eq5} with $a_i'=a_i$, $b_i'=b_i-a_i\cdot x_i$, $c_i'=c_i$ and $d_i'=d_i-c_i\cdot x_i$.

If the operator sum is the root of the tree, then $a_i'=a_i=1$, $c_i'=c_i=0$, $d_i'=d_i=-1$, and $b_i=t_i$. In this case, the resulting equations for the two children are the following

\begin{equation}
\label{eq:eqSumRoot1}
\frac{1}{N}\sum_{i=1}^N{(x_i-(t_i-y_i))^2}
\end{equation} 

\begin{equation}
\label{eq:eqSumRoot2}
\frac{1}{N}\sum_{i=1}^N{(y_i-(t_i-x_i))^2}
\end{equation} 

and can be interpreted for each child as "move the target value substracting from it the value of the output of the other child" and apply the original equation. Thus, as was done in the root node, two spheres in the space are created, with centers in t-y (for the first child) and t-x (for the second child), i.e., the target values for the root of the tree has been moved to a different position for each of the children. If any tree is found which its outputs, applied to equations \ref{eq:eqSumRoot1} or \ref{eq:eqSumRoot2} has a lower value, then the MSE will be reduced. From another point of view, if a subtree is found inside one of these spheres, the corresponding first or second child of the root can be replaced by this new subtree. As a consequence, the semantic of the tree will move towards the target, having an improvement in the overall result. In general, from a shape given by equation \ref{eq:eq5} the sum operation creates two new identical shapes in other points.

Figure \ref{fig:FigAddMul} a) shows an example of a tree situated in p. This tree is $(p_1+p_2)$. The calculation of the equations for each child leads to having a similar shape, but translated according to the values of $p_1$ and $p_2$. The resulting shapes (in this case, spheres) still have the semantics in the border.

\item Substraction operation. This case is very similar to the previous one: the output of the node is written as $o_i=x_i-y_i$, being $x_i$ and $y_i$ the outputs (semantics) of its two children. The equation for the first child becomes the following:

\begin{equation}
\label{eq:eqSubChild1}
\frac{1}{N}\sum_{i=1}^N{(\frac{a_i \cdot (x_i-y_i)-b_i}{c_i \cdot (x_i-y_i)-d_i})^2} = \frac{1}{N}\sum_{i=1}^N{(\frac{a_i \cdot x_i-(b_i+a_i \cdot y_i)}{c_i \cdot x_i-(d_i+c_i \cdot y_i)})^2}
\end{equation} 

which has the shape of equation \ref{eq:eq5} with $a_i'=a_i$, $b_i'=b_i+a_i \cdot y_i$, $c_i'=c_i$ and $d_i'=d_i+c_i \cdot y_i$. For the second child, the equation is:

\begin{equation}
\label{eq:eqSubChild2}
\frac{1}{N}\sum_{i=1}^N{(\frac{a_i \cdot (x_i-y_i)-b_i}{c_i \cdot (x_i-y_i)-d_i})^2} = \frac{1}{N}\sum_{i=1}^N{(\frac{-a_i \cdot y_i-(b_i-a_i \cdot x_i)}{-c_i \cdot y_i-(d_i-c_i \cdot x_i)})^2}
 = \frac{1}{N}\sum_{i=1}^N{(\frac{a_i \cdot y_i-(a_i \cdot x_i-b_i)}{c_i \cdot y_i-(c_i \cdot x_i-d_i)})^2}
\end{equation} 

which has the shape of equation \ref{eq:eq5} with $a_i'=a_i$, $b_i'=a_i \cdot x_i-b_i$, $c_i'=c_i$ and $d_i'=c_i \cdot x_i-d_i$.

Again, if this operator is used as root of the tree, the sphere of the root will be moved to t+y for the first child and x-t for the second child, but the idea is the same: if a tree is found inside any of these spheres, the corresponding child can be replaced with this tree and the overall result will be improved. In general, the sum and substraction operations creates two new identical shapes in other points.

\begin{figure}
    \centerline{\includegraphics[width=12cm, height=4cm]{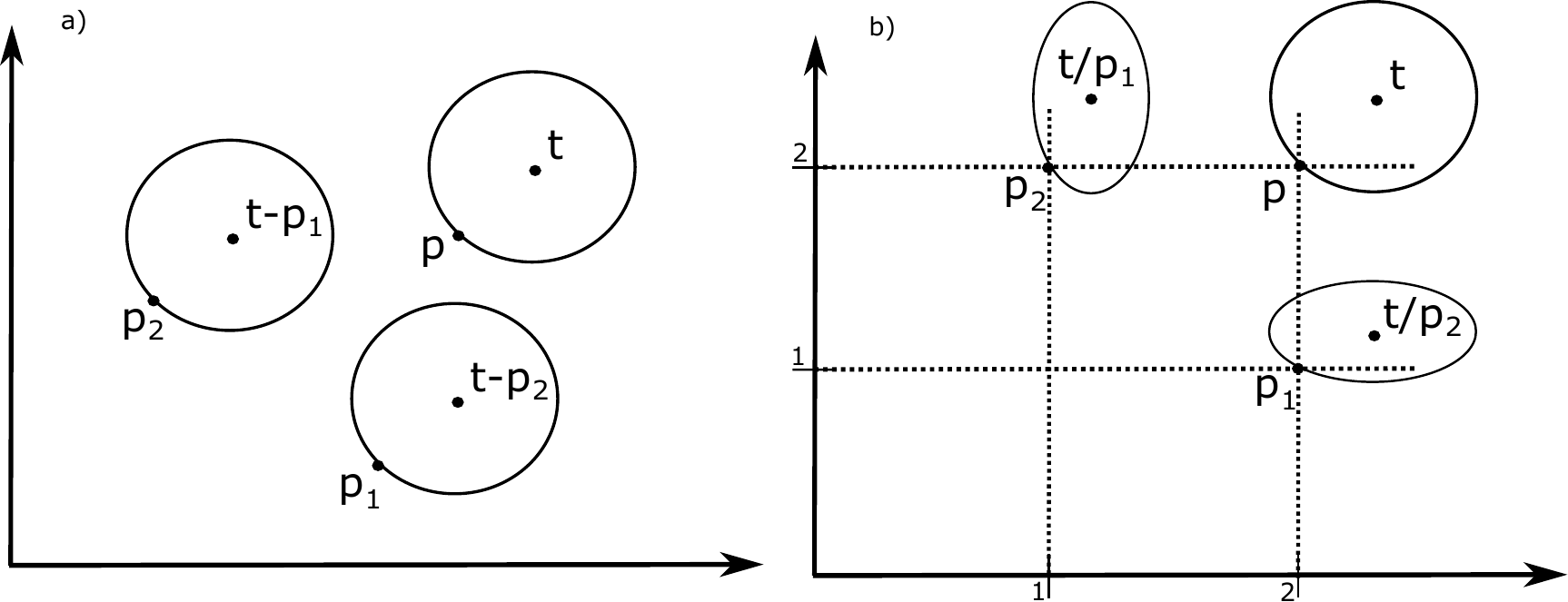}}
    \caption{Examples of calculating new shapes for the sum operation (a), and multiplication operation (b)}
    \label{fig:FigAddMul}
\end{figure}

\item Multiplication operation. In this case, the output of the node is written as $o_i=x_i\cdot y_i$, being $x_i$ and $y_i$ the outputs (semantics) of its two children. The equation for the first child becomes the following:

\begin{equation}
\label{eq:eqMulChild1}
\frac{1}{N}\sum_{i=1}^N{(\frac{a_i \cdot (x_i \cdot y_i)-b_i}{c_i \cdot (x_i \cdot y_i)-d_i})^2} = \frac{1}{N}\sum_{i=1}^N{(\frac{(a_i \cdot y_i) \cdot x_i-b_i}{(c_i \cdot y_i) \cdot x_i-d_i})^2}
\end{equation} 

which has the shape of equation \ref{eq:eq5} with $a_i'=a_i \cdot y_i$, $b_i'=b_i$, $c_i'=c_i \cdot y_i$ and $d_i'=d_i$. For the second child, the equation is very similar:

\begin{equation}
\label{eq:eqMulChild2}
\frac{1}{N}\sum_{i=1}^N{(\frac{a_i \cdot (x_i \cdot y_i)-b_i}{c_i \cdot (x_i \cdot y_i)-d_i})^2} = \frac{1}{N}\sum_{i=1}^N{(\frac{(a_i \cdot x_i) \cdot y_i-b_i}{(c_i \cdot x_i) \cdot y_i-d_i})^2}
\end{equation} 

If the multiplication operator is the root of the tree, then $a_i'=a_i=1$, $c_i'=c_i=0$, $d_i'=d_i=-1$, and $b_i=t_i$. In this case, taking into account that these equations allow the calculation of the MSE, consequently, the resulting equations for the two children are the following, :

\begin{equation}
\label{eq:eqMulRoot1}
\frac{1}{N}\sum_{i=1}^N{(y_i \cdot x_i-t_i)^2}=MSE
\end{equation} 

\begin{equation}
\label{eq:eqMulRoot2}
\frac{1}{N}\sum_{i=1}^N{(x_i \cdot y_i-t_i)^2}=MSE
\end{equation} 

These equations can be rewritten as

\begin{equation}
\label{eq:eqMulRoot1_2}
\sum_{i=1}^N{\frac{(x_i-\frac{t_i}{y_i})^2}{(\frac{SSE}{y_i})^2}}=1
\end{equation} 

\begin{equation}
\label{eq:eqMulRoot2_2}
\sum_{i=1}^N{\frac{(y_i-\frac{t_i}{x_i})^2}{(\frac{SSE}{x_i})^2}}=1
\end{equation} 

and therefore are the equations of ellipsis. These equations can be interpreted for each child as "move the target value through the line determined by the target vector, and shrink or extend the radius of the sphere in each dimension according to the other child outputs" and apply the original equation. Thus, the sphere of the root of the tree becomes ellipses in each of the two children. However, the reasoning is the same: if a new subtree is found inside these new shapes (ellipsis), then the result of the application of equation \ref{eq:eqMulRoot1} or \ref{eq:eqMulRoot2} will be equal to a lower MSE, and its semantic will move towards the target. In general, the multiplication operation creates new shapes in other points, which are distortions of this one.

Figure \ref{fig:FigAddMul} a) shows an example of a tree situated in p. This tree is $(p_1\cdot p_2)$, with $p_1=(1,2)$ and $p_2=(2,1)$. The calculation of the equations for each child leads to having the sphere translated and shrink according to the values of $p_1$ and $p_2$. The resulting shapes are ellipsis, and, as in the rest of the cases, still have the semantics in the border.

The use of the operations of sum, substraction and multiplication allows the building of complex trees, and having a semantic space with different N-dimensional ellipsoids. If a model is found inside one of these ellipsoids, the corresponding node can be changed with that model. When this is done, the semantic of the root of the tree get closer to the target, the MSE gets lower, the radius of the sphere of the root of the tree shrinks and the rest of the spheres/ellipses move from their places and shrinks in the same amount.

\item Division operator. In this case, the output of the node is written as $o_i=x_i/y_i$, being $x_i$ and $y_i$ the outputs (semantics) of its two children. The equation for the first child becomes the following:

\begin{equation}
\label{eq:eqDivChild1}
\frac{1}{N}\sum_{i=1}^N{(\frac{a_i \cdot (x_i/y_i)-b_i}{c_i \cdot (x_i/y_i)-d_i})^2} = \frac{1}{N}\sum_{i=1}^N{(\frac{\frac{a_i \cdot x_i-b_i \cdot y_i}{y_i}}{\frac{c_i \cdot x_i-d_i \cdot y_i}{y_i}})^2} = \frac{1}{N}\sum_{i=1}^N{(\frac{a_i \cdot x_i-b_i \cdot y_i}{c_i \cdot x_i-d_i \cdot y_i})^2}
\end{equation} 

which has the shape of equation \ref{eq:eq5} with $a_i'=a_i$, $b_i'=b_i\cdot y_i$, $c_i'=c_i$ and $d_i'=d_i\cdot y_i$. For the second child, the equation is:

\begin{equation}
\label{eq:eqDivChild2}
\frac{1}{N}\sum_{i=1}^N{(\frac{a_i \cdot (x_i/y_i)-b_i}{c_i \cdot (x_i/y_i)-d_i})^2} = \frac{1}{N}\sum_{i=1}^N{(\frac{\frac{a_i \cdot x_i-b_i \cdot y_i}{y_i}}{\frac{c_i \cdot x_i-d_i \cdot y_i}{y_i}})^2} = \frac{1}{N}\sum_{i=1}^N{(\frac{-b_i \cdot y_i-(-a_i \cdot x_i)}{-d_i \cdot y_i-(-c_i \cdot x_i)})^2} = \frac{1}{N}\sum_{i=1}^N{(\frac{b_i \cdot y_i-(a_i \cdot x_i)}{d_i \cdot y_i-(c_i \cdot x_i)})^2}
\end{equation} 

which has the shape of equation \ref{eq:eq5} with $a_i'=b_i$, $b_i'=a_i \cdot x_i$, $c_i'=d_i$ and $d_i'=c_i \cdot x_i$.

If this operator is the root of the tree, then $a_i'=a_i=1$, $c_i'=c_i=0$, $d_i'=d_i=-1$, and $b_i=t_i$. In this case, the resulting equations for the two children are the following:

\begin{equation}
\label{eq:eqDivRoot1}
\frac{1}{N}\sum_{i=1}^N{(\frac{x_i-t_i \cdot y_i}{y_i})^2} = \frac{1}{N}\sum_{i=1}^N{(\frac{1}{y_i} \cdot x_i-t_i)^2}
\end{equation} 

\begin{equation}
\label{eq:eqDivRoot2}
\frac{1}{N}\sum_{i=1}^N{(\frac{-t_i \cdot y_i+x_i}{y_i})^2} = \frac{1}{N}\sum_{i=1}^N{(\frac{x_i}{y_i}-t_i)^2}
\end{equation} 

In the first case, the effect is similar to the multiplication operation: the target value for the first child is moved along the line given by the target vector, and the radius is extended/shrink according to the values of the second child. In the second case, the sphere is turned into a completely different shape. Therefore, the division operation can transform ellipsoids into different shapes.

\end{itemize}

The third column of table \ref{tableExampleTree} shows the equations calculated for each of the nodes of the tree shown in fig. \ref{fig:figExampleTree}. The target vector used was (5,4,1).

The division operator is different from the rest for a very important reason: in the other three, the domain is all ${\rm I\!R}$. However, in the division operator the domain is restricted. This makes it have to be used with care. For example, figure \ref{fig:figExampleTree} shows an example of a tree with the division operator. The expression representing this tree is shown on eq. \ref{eq:eqExample}. However, it can be simplified to the following:

\begin{equation}
\label{eq:eqExample2}
\frac{x_3-2}{2\cdot x_3-1}
\end{equation}

However, both expressions are different, since the second is defined in any value of $x_3$ that belongs to ${\rm I\!R}-\{0.5\}$, while in the first one the values of $x_3$ that are not in the domain are 0, 2 and 0.5. Note that this last value is the same value in the first expression.

One could argue that the values that are not in the domain in the first equation are "cancelled" in successive divisions, i.e., giving rise to infinite values in the first division, and to 0 in the next. However, more complex cases may arise, as in the following equation:

\begin{equation}
\label{eq:eqExample3}
\frac{1}{\frac{3\cdot x_1}{x_3}-\frac{4\cdot x_4}{2\cdot x_3}}
\end{equation} 

When calculating the equations in each node of the tree, in the previous example shown in fig. \ref{eq:eqExample} it can be seen that the denominator of the equation corresponding to the variable $x_3$ has a root in 0.5, that is, the domain of that equation is the same as the domain of the equation \ref{eq:eqExample2} (simplified). In general, as you go down the equations in the tree, the domain of these equations will be equal to the domain of the first division operation performed in the tree. This effect is produced by the way the equation for the second child of the division operator is constructed, shown in eq. \ref{eq:eqDivChild2}. In that equation, this step

\begin{equation}
\label{eq:eqDivChildCorrected}
\frac{1}{N}\sum_{i=1}^N{(\frac{\frac{a_i \cdot x_i-b_i \cdot y_i}{y_i}}{\frac{c_i \cdot x_i-d_i \cdot y_i}{y_i}})^2} = \frac{1}{N}\sum_{i=1}^N{(\frac{a_i \cdot x_i-b_i \cdot y_i}{c_i \cdot x_i-d_i \cdot y_i})^2}
\end{equation} 

is correct for values of $y_i\ne0$. Therefore, for that node, any semantic in the second child with any $y_i=0$ is outside the domain of this node. For example, a node with a semantic of $(1,-2,0)$ can not be used as second child. Moreover, as the second child can be a tree, its semantic depends on its nodes and operations, since the result of these operations can lead to having a semantic with any $y_i=0$. For example, in the tree shown in fig. \ref{fig:figExampleTree}, the node 4 represents a division and thus its second child can not take the value of 0. If the variable $x_3$ takes the value of 2, then this denominator of the division operator of equation 4 will take a value of 0. Therefore, it is necessary to keep track of which values are not in the domain of a function, and, when applying new operators, operate with them consistently to calculate the new values that are not in the domain in the division operations.

Therefore, when calculating the equations it is necessary to take into account the domains of the operators already explored from the top of the tree. Since the +, - and * operators are defined in all ${\rm I\!R}$, only the division operator has to be taken into account. The way to do this is to extend the definition of the equations of each node. Until now, these were defined with four vectors, $a_i$, $b_i$, $c_i$ and $d_i$. In addition to these vectors, it is necessary to add a set of semantics $S$ that provoke any operation out of domain, in this case divisions by zero, in some upper node of the tree. As said, initially, that is, for the root node of the tree, $a_i=1$, $b_i=t_i$, $c_i=0$ and $d_i=1$, and also $S=\emptyset$.

As for the vectors $a_i$, $b_i$, $c_i$ and $d_i$, the values of the $S$ set must be calculated for each node. The process is similar to the one described: each non-terminal node makes the calculation of $a_i$, $b_i$, $c_i$, $d_i$ and $S$ for each children. Therefore, at the same time $a'_i$, $b'_i$, $c'_i$ and $d'_i$ are calculated for each child node, a new $S'$ set is calculated for this node. Given a node with an operation and a $S$ set, for each child the calculation method is the following:

\begin{enumerate}

\item $S'=\emptyset$

\item For each semantic $s\in S$, calculate $s'$ depending on the node operation. The way to calculate for one child is to perform the inverse of the node operation with respect to the other child. If x and y the two semantics of the first and second children respectively, the process is, for each operation and each child:

\begin{itemize}

\item Sum operation:

\begin{itemize}

\item First child: $s=s'+y \Rightarrow s'=s-y$.
\item Second child: $s=x+s' \Rightarrow s'=s-x$.

\end{itemize}

\item Substraction operation:

\begin{itemize}

\item First child: $s=s'-y \Rightarrow s'=s+y$.
\item Second child: $s=x-s' \Rightarrow s'=x-s$.

\end{itemize}

\item Multiplication operation:

\begin{itemize}

\item First child: $s=s'\cdot y \Rightarrow s'=s/y$.
\item Second child: $s=x\cdot s' \Rightarrow s'=s/x$.

\end{itemize}

\item Division operation:

\begin{itemize}

\item First child: $s=s'/y \Rightarrow s'=s\cdot y$.
\item Second child: $s=x/s' \Rightarrow s'=x/s$.

\end{itemize}

\end{itemize}

Since $s$, $s'$, $x$ and $y$ are vectors, the operations described are element-wise operations.

Once $s'$ is calculated, add it to $S'$.

\item In the case of the second child of the division operation, add the semantic $s'_i=0$ to $S'$, since this semantic is not in the domain of this operation.

\end{enumerate}

For each node, the set $S$ contains the semantics that make any operation of any previous node with a value out of domain, i.e., each $s$ semantic in $S$ has "forbidden" values. An important remark is that, if $s\in S$ has a semantic of "forbidden" values $s_i$, an invalid semantic $x$ will be a semantic in which any value $x_i=s_i$, i.e., they don't have to be equal all of the values.

The fourth column of table \ref{tableExampleTree} shows the S sets for each node of the tree shown in fig. \ref{fig:figExampleTree}. Nodes 1 and 2, before any division operation was performed, have $S=\emptyset$. Nodes 3, 4, 5 and 11, when only one division has been performed, have one element in $S$. Nodes 6, 7, 8 and 9, when 2 divisions have been performed, have 2 elements in $S$. Finally, when the calculation of the equations of node 10 is reached, 3 different divisions have been done, and therefore the number of elements in $S$ is 3. Note that each node used as divisor (nodes 3, 6 and 10) have the element (0,0,0) in $S$, meaning that no value of their semantics can be 0.

\begin{figure}
    \centerline{\includegraphics[width=6cm, height=5cm]{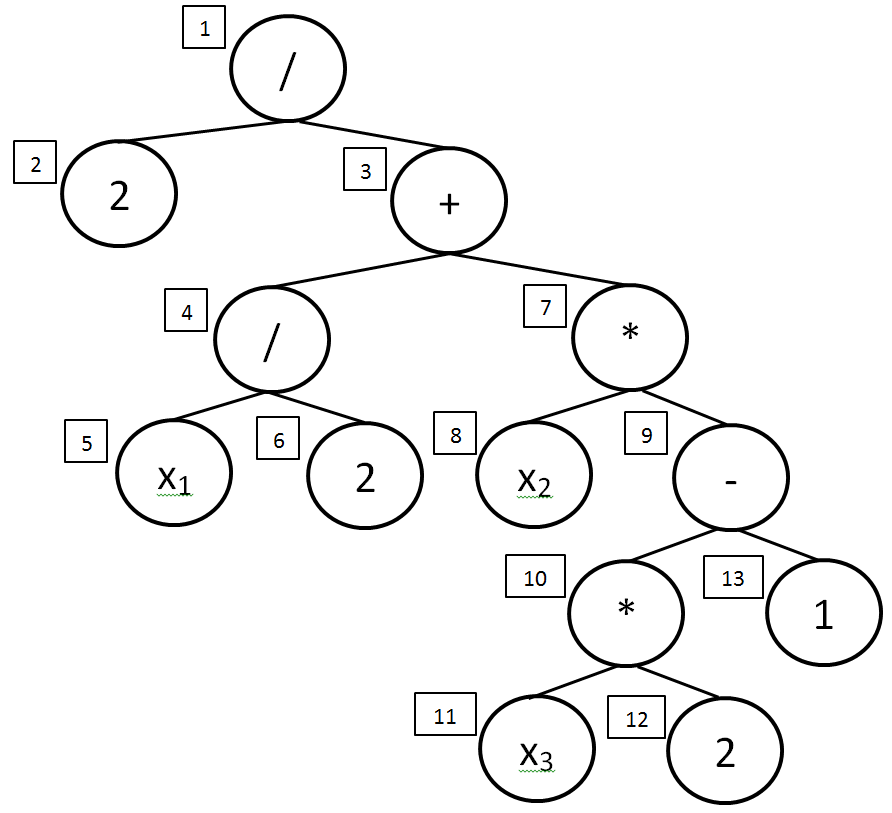}}
    \caption{Example of a tree with invalid operations}
    \label{fig:figExampleIncorrectTree}
\end{figure}

\begin{table}
\centering
\caption{Description of the tree of the example with invalid operations}
{
    \begin{tabular}{@{}ccccc@{}} 
    \toprule
        Node & \multicolumn{1}{c}{Semantic} & \multicolumn{1}{c}{S} \\
    \midrule
        1 & $(1.33, Inf, Inf)$ & $\emptyset$ \\
        2 & $(2, 2, 2)$ & $\emptyset$ \\
        3 & $(1.5, 0, 0)$ & $\{(0, 0, 0)\}$ \\
        4 & $(1.5, 1, 0)$ & $\{(0, 1, 0)\}$ \\
        5 & $(3, 2, 0)$ & $\{(0, 2, 0)\}$ \\
        6 & $(2, 2, 2)$ & $\{(0, 0, 0), (Inf, 2, NaN)\}$ \\
        7 & $(0, -1, 0)$ & $\{(-1.5, -1, 0)\}$ \\
        8 & $(0, 1, 0)$ & $\{(-1.5, 1, 0)\}$ \\
        9 & $(1, -1, 5)$ & $\{(Inf, -1, NaN)\}$ \\
        10 & $(2, 0, 6)$ & $\{(Inf, 0, NaN)\}$ \\
        11 & $(1, 0, 3)$ & $\{(Inf, 0, NaN)\}$ \\
        12 & $(2, 2, 2)$ & $\{(Inf, Inf, NaN)\}$ \\
        13 & $(1, 1, 1)$ & $\{(Inf, 1, NaN)\}$ \\
        \\
%    \bottomrule
    \end{tabular}
}
\label{tableExampleIncorrectTree}
\end{table}

Another example is shown in Fig. \ref{fig:figExampleIncorrectTree}, which has its semantics and S sets described on table \ref{tableExampleIncorrectTree}. Note that this tree is invalid, since divisions by zero are performed. thus, this tree will never be generated by this system. However, it is used here as an example for a better explanation of the following descriptions.
    
Special care has to be taken when calculating each semantic $s'$ from $s$ for the multiplication and division operators. In this case, divisions by 0 may occur in the following situations:

\begin{itemize}

\item In the case of the multiplication operation, for the first child when any element of $y_i$ (semantic of the second child) is 0 or, for the second child, when any element of $x_i$ (semantic of the first child) is 0, i.e., a multiplication of $0\cdot y_i$ or $x_i\cdot 0$ is being made. In both cases, the result of the multiplication is equal to 0, and the calculation of the new values out of domain are $s'=s/y$ (for the first child) and $s'=s/x$ (for the second). Two situations are possible:

\begin{itemize}

\item $s_i\ne 0$: the "forbidden value" is not 0. In this case, $s_i/x_i$ or $s_i/y_i$ equals to infinite ($Inf$). This means that any value is valid for $x_i$ or $y_i$ because $x_i\cdot 0=0\ne s_i$ or $0\cdot y_i=0\ne s_i$ and therefore they are never out of the domain. An example of this can be seen in Fig. \ref{fig:figExampleIncorrectTree} and table \ref{tableExampleIncorrectTree}, in the S set of node 9 (first value of the only element).

\item $s_i=0$: the "forbidden value" is 0. In this case, $s_i/x_i$ or $s_i/y_i$ equals to "not a number" ($NaN$). This means that there are not valid values for $x_i$ and $y_i$, since for any $x_i$ or $y_i$ values, $x_i\cdot 0=0=s_i$ or $0\cdot y_i=0=s_i$, and therefore they are always out of the domain. An example of this can be seen in Fig. \ref{fig:figExampleIncorrectTree} and table \ref{tableExampleIncorrectTree}, in the S set of node 9 (third value of the only element).

\end{itemize}

\item In case of the division operation, for the first child, the reasoning is the same as in the multiplication operation.

\item For the second child of the division operation, $s=x/s'$ the operation to calculate the new "forbidden" semantic is $s'=x/s$. Division by zero will take place when any $s_i=0$: the "forbidden value" of the result of the division operation is 0. Two situations may occur:

\begin{itemize}

\item $x_i\ne 0$. In this case, the result of this operation is $s'_i=x_i/s_i=Inf$, meaning that any $y_i$ value is valid, since $x_i/y_i\ne s_i=0$ if $x_i\ne 0$.

This can be argued, since a value of $y_i=0$ is not valid for this division. However, the calculation of each $s'$ corresponds to values out of the domain of previous division operators, not for this one. For the current division operation, as was already explained, the semantic $s'=0$ is added to $S$, and therefore the value of $y_i=0$ is not valid.

An example of this can be seen in Fig. \ref{fig:figExampleIncorrectTree} and table \ref{tableExampleIncorrectTree}, in the S set of node 6 (first value of the second element).

\item $x_i=0$. In this case, the result of the division operation is $x_i/y_i=0=s_i$. The calculation of $s'_i$ is the following: $s'_i=x_i/s_i=0/0=NaN$, meaning that there is not valid values for $y_i$, i.e., for any value $y_i$, $x_i/y_i=0=s_i$.

Again, these calculations are valid for $y_i\ne 0$. The case $y_i=0$ is excluded with the addition of the semantic $s'_i=0$ to $S$;

An example of this can be seen in Fig. \ref{fig:figExampleIncorrectTree} and table \ref{tableExampleIncorrectTree}, in the S set of node 6 (third value of the second element).

\end{itemize}

\end{itemize}

Therefore, for each $s\in S$, a value labeled as $Inf$ represents that any value is valid, while a value labeled as $NaN$ represents that any value is invalid. The operations between these values are done in the usual way, if $x\in {\rm I\!R}$:

\begin{itemize}

\item Any operation between $x$ and $NaN$ has $NaN$ as result. This means that if any value is out of domain in a specific node, then any value will also be out of domain in each children, no matter what operation is done in this node. An example of this can be seen in Fig. \ref{fig:figExampleIncorrectTree} and table \ref{tableExampleIncorrectTree}, in the S sets of nodes 10 and 13 (third value of the only element). Also, node 12 shows another example (third value of the only element).

\item Sum and substraction operations between $x$ and $Inf$ return $Inf$. This means that if any value is valid (the domain is ${\rm I\!R}$) for the result of the operation, then any value is also valid for each children. An example of this can be seen in Fig. \ref{fig:figExampleIncorrectTree} and table \ref{tableExampleIncorrectTree}, in the S sets of nodes 10 and 13 (first value of the only element).

\item $s'_i=Inf/x_i=Inf$ ($x_i=0$ or $x_i\ne 0$). This can happen in a when $s'$ is being calculated as one of the children in a multiplication operation. This means that if any value is valid as result of a multiplication operation ($Inf$), then each child can have any value. An example of this can be seen in Fig. \ref{fig:figExampleIncorrectTree} and table \ref{tableExampleIncorrectTree}, in the S set of node 12 (first and second values of the only element).

\item $s'_i=x_i/Inf=0$ ($x_i=0$ or $x_i\ne 0$). This can happen when $s'$ is being calculated as the second child in a division operation (i.e., the S set of the denominator) with the semantic of the first child $x_i$. This means that even any value (denoted as $Inf$) is valid as result of a division operation in which the numerator is a valid number, the denominator must different from 0 (0 is not in the domain).

\item $s'_i=Inf\cdot y_i=Inf$ when $x_i\ne 0$. This can happen when $s'$ is being calculated as the first child in a division operation (i.e., the S set of the numerator) with the semantic of the denominator $y_i\ne 0$. This means that if the result of a division operation can be any value ($Inf$) and the denominator is different to 0, then the first child can take any value in ${\rm I\!R}$.

\item $s'_i=Inf\cdot y_i=NaN$ when $y_i=0$. This can happen when $s'$ is being calculated as the first child in a division operation (i.e., the S set of the numerator) with the semantic of the denominator $x_i\ne 0$. This means that even if the result of a division operation can be any value, if the second child evaluates to 0, then there is no valid value for the first child.

\item Operations between $NaN$ and/or $Inf$ values will not happen because these operations take place between the semantics of the S set, which may contain $NaN$ and $Inf$ values, and semantics of the nodes of the tree, which may not. The definition of the S set is done to prevent the semantics of the nodes from having $NaN$ and $Inf$ values.

\end{itemize}

Any semantic $x$ to be applied the equation \ref{eq:eq5} of any node in order to compute the MSE from that node must be first checked with its $S$ set. For each semantic $s$ in $S$, if there is any value in which $s_i=x_i$ or $s_i=NaN$, then that semantic can not be used because it would lead to having out-of-domain values in preceeding nodes. In the case in any which $s_i=NaN$, since there are no valid values, for simplicity reasons, the whole $S$ set is labeled as $NaN$.

With this method, once the tree has been evaluated, the equation can be calculated for each node. Therefore, before a tree is going to be improved, each node has to be evaluated, and this information has to be stored on its corresponding node. This evaluation process goes from the bottom of the tree to the top. After it, the values of $a_i$, $b_i$, $c_i$, $d_i$ and $S$ of each node have to be calculated. This process goes from the top, with values of $a_i=1$, $c_i=0$, $d_i=-1$, $b_i=t_i$ and  $S=\emptyset$, to the bottom of the tree, following the described equations.

Once these values have been found for each node, the search for subtrees that can substitute a node begins. For any subtree that could substitute a node, first its semantic is checked with the set $S$ of the node. If all of the values are valid, then the semantic is evaluated on the equation \ref{eq:eq5} of that node. This subtree can replace the node if the result of the following equation is positive:

\begin{equation}
\label{eq:eqOptimize}
MSE - \frac{1}{N}\sum_{i=1}^N{(\frac{a_i \cdot o_i-b_i}{c_i \cdot o_i-d_i})^2}
\end{equation} 

If there are several subtrees that could replace a node, the selected subtree will be the one with the highest positive value in this equation. The search for subtrees has 4 different methods: search for constants (constant search), search for variables (variable search), search for constants combined with variables (constant-variable search) and search for constants combined with expressions (constant-expression search). These searches do not take place in those nodes in which in which $S=NaN$, since there are no valid values. However, since this method always builds correct trees, no nodes with $S=NaN$ are expected. The following subsections discuss each of these searches.

\begin{figure}
    \centerline{\includegraphics[width=12cm, height=8cm]{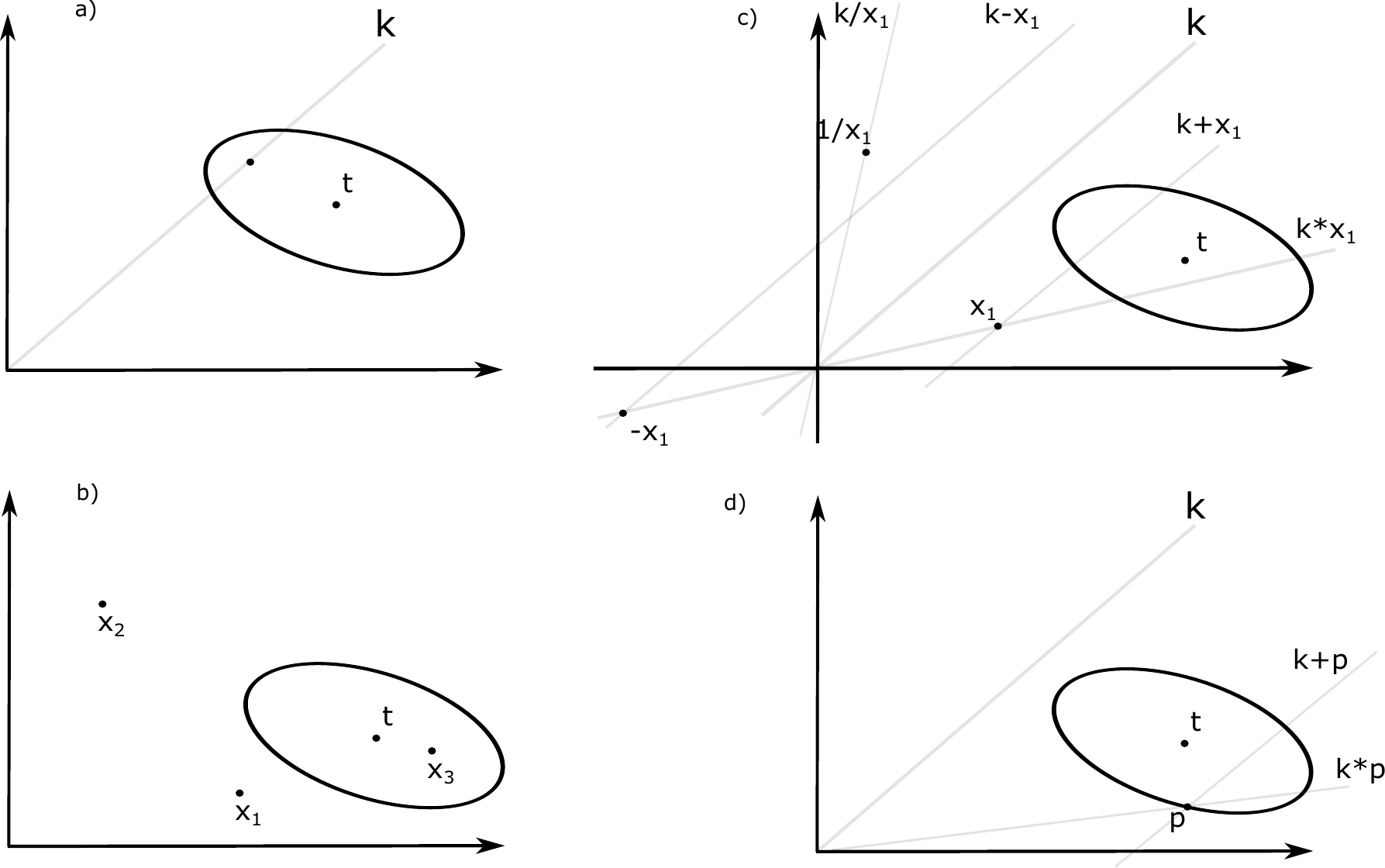}}
    \caption{Examples of a) constant search, b) variable search, c) constant-variable search, d) constant-expression search }
    \label{fig:FigSearches}
\end{figure}

\subsection{Search for constants}

\label{searchForConstants}

One of the biggest problems in GP is constant generation. In the first approaches, constant generation was left to the evolutionary process. An ephemeral random constant was included into the terminal set, so each time it was selected in the building of a tree, a random constant in a predefined interval was generated. The building of a useful value was left to the evolutionary process, by successive combination of these random constants. Even there are some approaches using gradient descend \cite{7257017}\cite{GPFactorVariables}, this technique is still slow, demanding a high number of useless operations, being very inefficient.

Within a tree, constants take place as a terminal node. Therefore, they can be seen as trees with a single node, representing its constant value, and therefore they are also considered as a model, with a semantic. The particularity of the semantic of a constant k is that all of the elements of the vector semantic take the same value: $o_i=k$. In other words, the semantics of all of the constants are situated on the line $span(1,1,...,1)$.

Given a node of the tree, with its corresponding equation, the objective is to find the constant that maximizes equation \ref{eq:eqOptimize}, i.e., the constant k that minimizes the following equation:

\begin{equation}
\label{eq:eqConstant}
\frac{1}{N}\sum_{i=1}^N{(\frac{a_i \cdot o_i-b_i}{c_i \cdot o_i-d_i})^2} = \frac{1}{N}\sum_{i=1}^N{(\frac{a_i \cdot k-b_i}{c_i \cdot k-d_i})^2}
\end{equation} 

The way to find this constant is to calculate the derivative of this expression, set it equal to zero, and calculate the value k. Being the original equation a sum of squared expressions, any value of k that makes the derivative equal to 0 will be a minimum, since equation \ref{eq:eqConstant} does not have any maximum. The derivative of this expression is the following equation:

\begin{equation}
\label{eq:eqDerivativeGeneral}
\frac{2}{N}\sum_{i=1}^N{(b_i \cdot c_i-a_i \cdot d_i)\frac{a_i \cdot k-b_i}{(c_i \cdot k-d_i)^3}}
\end{equation} 

In general, calculating the values in which this expression becomes 0 is a time-consuming task, because it involves building a 3N-order polynomial and finding its roots, with N possibly being very high. Also, there are many possible values for k. However, this calculation can be simplified in some common situations:

\begin{itemize}

\item $c_i=0$ and $a_i=0$ (i=1,...,N). In this case, the function is constant and has no minimum.

\item $c_i=0$ (i=1,...,N) and any $d_i=0$. In this case, the function can not be calculated and therefore there is no minimum.

\item $c_i=0$ and $d_i\ne0$ (i=1,...,N). In this case, the only minimum value of k is given by:

\begin{equation}
\label{eq:eqDerivative1}
k=\frac{\sum_{i=1}^N{\frac{a_i \cdot b_i}{d_i^2}}}{\sum_{i=1}^N{\frac{a_i^2}{d_i^2}}}
\end{equation} 

This equation can be simplified in the following situations:

\begin{itemize}

\item $a_i$ are constants ($a_i=k_a$) and $d_i$ are constants ($d_i=k_d$). The case with $k_d=-1$ happens when no division has been performed yet. The minimum k is given by:

\begin{equation}
\label{eq:eqDerivative2}
k=\frac{1}{k_a\cdot N}\sum_{i=1}^N{b_i}
\end{equation} 

\item $a_i$ are constants ($a_i=k_a$) and $d_i$ are not constants. The minimum k is given by:

\begin{equation}
\label{eq:eqDerivative3}
k=\frac{1}{k_a}\frac{\sum_{i=1}^N{\frac{b_i}{d_i^2}}}{\sum_{i=1}^N{\frac{1}{d_i^2}}}
\end{equation} 

\item $a_i$ are not constants and $d_i$ are constants ($d_i=k_d$). The minimum k is given by:

\begin{equation}
\label{eq:eqDerivative4}
k=\frac{\sum_{i=1}^N{a_i \cdot b_i}}{\sum_{i=1}^N{a_i^2}}
\end{equation} 

\end{itemize}

\item Any $c_i=0$ and $d_i=0$ (i=1,...,N). In this case, the function can not be calculated and therefore has no minimum.

\item $c_i\ne0$ and $d_i=0$ (i=1,...,N). In this case, the only minimum value of k is given by:

\begin{equation}
\label{eq:eqDerivative5}
k=\frac{\sum_{i=1}^N{\frac{b_i^2}{c_i^2}}}{\sum_{i=1}^N{\frac{a_i \cdot b_i}{c_i^2}}}
\end{equation} 

As happens with equation \ref{eq:eqDerivative1}, this equation can be simplified for the cases in which $b_i$ and/or $c_i$ are constants.

\item $c_i$ and $d_i$ are constants $c_i=k_c$ and $d_i=k_d$ (i=1,...,N), and $k_c\ne0$ and $k_d\ne0$. In this case, the only minimum value of k is given by:

\begin{equation}
\label{eq:eqDerivative6}
k=\frac{\sum_{i=1}^N{b_i \cdot (b_i \cdot k_c-a_i \cdot k_d)}}{\sum_{i=1}^N{a_i \cdot (b_i \cdot k_c-a_i \cdot k_d)}}
\end{equation} 

\end{itemize}

In any of these cases, once the minimum $k$ has been found, it has to be checked with the $S$ set. For each semantic $s\in S$, if any value $s_i=k$, then this search is unsuccessful.

If none of these situations take place, as was stated, the finding of the minimum values can be a time-demanding task. However, in this work we propose the alternative solution of evaluating a set of points that can have a low value in equation \ref{eq:eq5}. These values are the zeros on each term inside the sum. These values are given by $b_i/a_i$. In general, the minimum value of equation \ref{eq:eq5} will not be any of these values. However, one of them will take a close value. In order to avoid rounding errors, in this calculations those zeros with values close to any poles ($d_i/c_i$) are excluded.

Therefore, in the general case the process is the following:
\begin{enumerate}

\item Take the N $z_i=b_i/a_i$ zero values.

\item Compare each $z_i$ with all of the values of each $s\in S$. If there is any coincidence, delete $z_i$, since it is not valid.

\item Exclude those values which are too close to any root of the denominator of the equation \ref{eq:eq5}.

\item Evaluate the remaining $z_i$ with equation \ref{eq:eq5} and select the one with the lowest value (lowest MSE) as k.

\end{enumerate}

Once a value of k has been found for a node, if the result of equation \ref{eq:eqOptimize} is positive, then this search was successful and a single-node terminal tree representing this constant can substitute this node in the tree, leading to an improvement in the MSE.

This process allows the creation and refinement of constants. However, this is not limited to changing the value of one constant for another (i.e., one terminal node representing a constant for another terminal node representing a constant). It is important to highlight that this process of finding a constant can be performed for each node of the tree to be improved, whether it is a constant, variable or a non terminal node. The user may decide on which nodes he wants this search to be performed. Therefore, the node selected to be replaced by a constant can be a non-terminal. In this case, if this search is successful, the tree is being reduced, having the result a lower number of nodes.

In general, it has been found that, as result of this search, most of the times a constant is changed into another. Therefore, this search technique is very useful to refine the constant values of the nodes of the tree. Sometimes a variable or a non-terminal node (thus reducing the tree) is changed by a constant. However, since these situations occur very few times, much computational time can be saved by performing this search only on constant nodes, and exploring the rest of the nodes only when needed. As in the rest of the searches, the user can decide whether he wants to perform this search only on constant nodes and maybe skip an interesting improvement, or perform a full search on all of the nodes of the tree.

Figure \ref{fig:FigSearches} a) shows and example of a constant search. The shape given by equation \ref{eq:eq5} has an intersection with the grey line defined as span{(1,1,...,1)}. Any constant node is on this line. Therefore, any constant node on this line and inside the ellipsis would give a lower MSE than the given tree. This method returns the constant that returns the lowest MSE for this shape.

\subsection{Search for variables}

This process is easier to calculate than the previous one. As happens with constants, each variable is represented by a terminal node, with a semantic. Therefore, variables are also points in the search space.

Given a node of the tree, with its equation \ref{eq:eq5} and its $S$ set, the first to do is to check among all of the variables if their semantics are valid with the $S$ set. For those variables with valid semantics, the equation \ref{eq:eqOptimize} is evaluated with their semantics. Again, a positive value means a reduction in MSE. These are variables inside the shape of that node in the semantic space. The selected variable will be the one with the highest reduction in MSE.

As in the previous search process, any node of a tree can be selected to be changed with a variable, being this node constant, variable or non-terminal. Therefore, the number of nodes in the tree can not grow with this method, as in the previous one, and the tree can be reduced. As in the previous search, the user may decide on which nodes he wants to perform this search. Obviously, if this search on being performed on variable nodes, those variables equal to the nodes being explored are excluded.

Also, as happened in the previous search, it has been found that this search is successful most of the times on constant nodes. Therefore, computational time can be saved if this search is performed only on constant nodes, and exploring the rest of the nodes only when needed.

Those variables with constant values (the same values for each datapoint) are excluded from this computation, since they are constants and could be optimised by the previous method.

Figure \ref{fig:FigSearches} b) shows and example of a variable search. The variable $x_3$ was found to be inside the shape given by equation \ref{eq:eq5}. Therefore, the given tree can be replaced by this variable, having an improvement in MSE. Variables $x_1$ and $x_2$ do not lead to improving MSE.

\subsection{Search for variables combined with constants}

\label{sec:constantVariableSearch}

With the two search methods already described, the number of nodes of the tree can become lower. However, often there is the necessity to find larger and more complex trees to get closer to the solution. This subsection allows the finding of simple subtrees with 3 nodes to replace another node of the tree. If this node to be replaced is a terminal node, then the tree will grow.

The idea behind this search method is a combination of the two previous methods. After the calculation of each equation on the nodes of the tree, it is possible not to find a constant or a variable that improves the MSE (a constant or a variable inside any of the shapes in the space). However, although a variable $x$ is not inside any of the shapes, the models $k+x$, $k-x$, $k\cdot x$ or $k/x$ may be inside one of the shapes, for any value of $k$. These four expressions represent, each of it, a line in the semantic space. This search method will look for the intersection of one of these lines with each of the shapes. Looking for intersections is equivalent to looking for the best value of k. Therefore, this third method proposes the search of a variable combined with a constant, with one of the for arithmetic operations. For each variable x, the possibilites are:

\begin{itemize}

\item Sum operation. Although the variable $x$ represents a single point in the search space, the expression $(k+x)$ represents a line in the search space. This line goes parallel to the line in which constants are situated. Therefore, if the variable $x$ is not inside any of the shapes, any section of the line $(k+x)$ may be inside of the shapes. The expression $(x+k)$ is represented by a tree with three nodes: a non-terminal node representing the sum, and two nodes, with the constant $k$ and the variable $x$.

The objective here is, given a node with its equation calculated, to find a constant value that maximizes equation \ref{eq:eqOptimize} for the semantic of the tree $(k+x)$. To do this, a new equation is calculated from the $a_i$, $b_i$, $c_i$, $d_i$ and $S$ values of the equation of this node. As the constant is going to be situated as the first child of the sum operation, this new equation is calculated as in equation \ref{eq:eqSumRoot1}, with $a_i'=a_i$, $b_i'=b_i-a_i \cdot x_i$, $c_i'=c_i$ and $d_i'=d_i-c_i \cdot x_i$, where $x_i$ are the values of the variable $x$ for each pattern. Also, $S'$ is calculated from $S$ with the method previously described for the sum operator. With this new equation, the constant optimization process described in section \ref{searchForConstants} is performed. As a result, an improvement value is returned, a positive one indicates that this node can be replaced with $(k+x)$.

\item Minus operation. Similarly with the previous case, the expression $(k-x)$ represents a line the space, parallel to the constant line, and parallel to $(k+x)$ line. Note that in the previous case the position of the constant and variable is indifferent: $(k+x)$ and $(x+k)$ leads to the same expression. However, in this case the constant must be the first argument of the minus operation, and the variable the second. Otherwise, we would be again in the previous case. to keep coherence, in all of these four cases the constant is going to be the first argument.

The process is similar in all of the four cases: given a node with an equation represented by $a_i$, $b_i$, $c_i$, $d_i$ and $S$, a new equation is calculated, this time from equation \ref{eq:eqDivChild1}: with $a_i'=a_i$, $b_i'=b_i+a_i \cdot x_i$, $c_i'=c_i$, $d_i'=d_i+c_i \cdot x_i$, and $S'$ calculated from $S$ with the method previously described for the substraction operation Once this equation was calculated, the constant optimization process is performed.

\item Multiplication operation. In this case, the expression $(k\cdot x)$ represents a line in the search space in which the vector x is included, i.e., all of the vectors in $(k\cdot x)$ are collinear to x. The objective here is the same: find the value of k that minimizes equation \ref{eq:eq5}.

Given a node of the tree with an equation represented by $a_i$, $b_i$, $c_i$, $d_i$ and $S$, a new equation is calculated, this time from equation \ref{eq:eqMulChild1}: $a_i'=a_i \cdot x_i$, $b_i'=b_i$, $c_i'=c_i \cdot x_i$, $d_i'=d_i$, and $S'$ calculated from $S$ with the method previously described for the multiplication operation. With this equation, the previous constant search process is undergone, having as result a value of reduction in MSE.

\item Division operation. In this case, the expression $(k/x)$ represents a line in the search space in which the vector given by the values $1/x_i$ is included, i.e., all of the vectors in $(k/x)$ are collinear to the vector given by $1/x_i$. With the same objective as in the previous cases, and given a node, from the equation of this node a new one is calculated, with equation \ref{eq:eqDivChild1}: $a_i'=a_i$, $b_i'=b_i \cdot x_i$, $c_i'=c_i$, $d_i'=d_i \cdot x_i$, and $S'$ calculated from $S$ with the method previously described for the division operation. The same constant optimization is performed.

As in the minus operation, this operation does not allow changing the order of the children. If $(x/k)$ was chosen instead of $(k/x)$, then an operation similar to the previous one (constant-variable search with multiplication operation) will be being performed, having as result the value of $1/k$.

\end{itemize}

In this search, for each of the four operations, a set $S'$ has to be calculated from the set $S$ of the node. If $S'=NaN$, then the constant search for that operation, node and variable does not take place, since there are not possible values for the constant. A particular case happens in the division operator when the variable has any 0 in its semantic. In this case, the result of the calculation of $S'$ is $NaN$, excluding those variables with 0 values in semantics from being the second child of a division operation.

This process can be done for each node of the tree, whether it is a constant, variable or non-terminal, each variable and each of the four operations. As a result, the combination selected is the one that returns a higher value in equation \ref{eq:eqOptimize}, in case it is higher than 0. This combination states which node can be changed and which 3-node tree insert in its place. The node to be changed can be terminal or non-terminal, so this search process can lead to having the tree increase or decrease the number of nodes. However, it has been found that most of the times this search is successful only on terminal nodes (constants and variables), and sometimes on non-terminal nodes. As in previous searches, performing this search only on terminal nodes may save computational time.

Figure \ref{fig:FigSearches} c) shows and example of a constant-variable search. The shape given by equation \ref{eq:eq5} does not have an intersection with the grey line defined as span{(1,1,...,1)}, and there are not variables inside this shape. Therefore constant and variables searches would not be successful. However, from the variable $x_1$ four different lines can be defined: $k+x_1$, $k-x_1$, $k*x_1$ and $k/x_1$. In this case, the grey lines $k+x_1$ and $k*x_1$ have an intersection with the shape, so in both cases a constant can be calculated to have an expression $k+x_1$ or $k*x_1$ that improves MSE. The given tree would be replaced with this expression.

As in the previous search, those variables with constant values (the same values for each datapoint) are excluded from this computation.

It is important to highlight that this search has to be carefully used so it does not "hide" a constant search. For instance, the branch "$(3.2 + x_3)$" can be selected for this search. As a result, this node could be changed by "$(3.5 + x_3)$" in this constant-variable search. However, in practice what was done is to change the constant node from a value of 3.2 to a value of 3.5. For this reason, for a specific variable and each of the four operations (+,-,*,/), this search is not performed on nodes representing the same operation, with a constant as first child and the same variable as second child.

Moreover, even with the described precaution, it is possible to use this operation to "hide" a constant search. Another example could be in the tree $(3.4 + x_2)$, in which $x_2$ could be replaced by $(0.1 + x_2)$. This has as result the tree $(3.4 + (0.1 + x_2))$, which has a semantic equivalent to $(3.5 + x_2)$, and could be generated by a constant search. Another example could be $((1.2\cdot x_1) + x_1)$. In this tree, the second $x_1$ could be changed into $(0.5\cdot x_1)$, and the resulting tree would be $((1.2\cdot x_1) + (0.5\cdot x_1))$, which has a semantic equivalent to $(1.7\cdot x_1)$ or $((0.7\cdot x_1) + x_1)$. This last tree could be found from the original one, by performing a constant search on the constant leaf.

In general, it is hard to determine when a constant-variable search can "hide" a constant search. This situation makes the trees unnecessarily large, with possibly a high number of nodes (see section \ref{sec:constantOptimization}). If a constraint to the size of the tree is set, then the algorithm could be prematurely stop because of having too many nodes that could be simplified.

One way to avoid this problem is, before this search takes place, optimize all of the constants. This process is described in section \ref{sec:constantOptimization}, and it is based on constant search. 

Another possibility is to perform a constant search when this search is done. With this approach, a modification of a constant and a constant-variable search that modifies the same constant will have the same MSE reduction. The first of them (constant search) would be the chosen to modify the tree.

\subsection{Search for expressions combined with constants}

\label{sec:constantExpressionSearch}

The previous search allows the tree to grow by replacing terminal nodes with small branches. However, once a part of the tree has been built, it will unlikely be modified, and its modification involves the deletion of a branch and substitution by a constant, variable or constant-variable branch. This can be undesirable, since that deleted part of the tree has proved to be useful. Therefore, a method that allows the modification of a branch without deleting it is desirable.

This search allows the tree to grow by modifying a non-terminal node of the tree. This modification is done as in the previous cases: finding a better branch and replacing the whole subtree with this branch. However, in this case, the branch contains the previous subtree, so it it not deleted but modified. For instance, a node $2.5/(3+x_2)$ can be replaced by the branch $(3.4 + 2.5/(3+x_2))$, in which the second child is the previous subtree.

The idea behind this search method is similar to constant-variable search. In this case, the node selected represents a point $p$ in the semantic space in the border of the MSE shape. There may be a constant $k$ that makes $(k+p)$ inside the MSE shape, thus improving the MSE. As in the previous search, $(k+p)$ is the line that goes through p, parallel to the constant line (1,1,...,1). Similarly, there may be a constant $k$ that makes $(k*p)$ inside the MSE shape. $(k*p)$ is the line that goes through p and (0,0,...,0). In both cases, finding a value for k means finding a point in the corresponding line inside the MSE shape. Other approaches such as $(k-p)$ or $(k/p)$ are not explored, because they do not begin in the border of the MSE shape and thus are not likely to return an improvement in MSE. In the first case, $(k-p)$, the line goes though $-p$ instead of $p$, and in the second case, $(k/p)$, the line goes through $1/p$ instead of $p$.

As opposite to the previous searches, this search is performed only on non-terminal nodes, because if it is performed on a terminal node:

\begin{itemize}

\item If this node is a constant, then a constant search is being performed.

\item If this node is a variable, then a constant-variable search is being performed.

\end{itemize}

The behaviour of this search is strongly similar to the constant-variable search. In the case of performing a $(k+p)$ search on a node $p$ with its equation, a constant value that maximizes equation \ref{eq:eqOptimize} for the semantic of the tree $(k+p)$ has to be found. To do this, a new equation is calculated from the $a_i$, $b_i$, $c_i$, $d_i$ and $S$ values of the equation of this node. As the constant is going to be situated as the first child of the sum operation, this new equation is calculated as in equation \ref{eq:eqSumRoot1}, with $a_i'=a_i$, $b_i'=b_i-a_i\cdot p_i$, $c_i'=c_i$, $d_i'=d_i-c_i\cdot p_i$, where $p_i$ are the values of the node $p$ for each pattern, and $S'$ is calculated from $S$ with the method described for the sum operation. With this new equation, the constant optimization process described in section \ref{searchForConstants} is performed. As a result, an improvement value is returned, a positive one indicates that this node can be replaced with $(k+p)$.

If a $(k\cdot p)$ search is performed on a node with an equation represented by $a_i$, $b_i$, $c_i$, $d_i$ and $S$, a new equation is calculated, this time from equation \ref{eq:eqMulChild1}: $a_i'=a_i\cdot p_i$, $b_i'=b_i$, $c_i'=c_i\cdot p_i$, $d_i'=d_i$, and $S'$ is calculated from $S$ with the method described for the multiplication operation. With this equation, a constant search process is undergone, having as result a value of reduction in MSE.

With this search, the size of tree is always increased in two nodes.

Figure \ref{fig:FigSearches} d) shows and example of a constant-expression search. In this case, the node is represented by the semantic p. From this point, two lines can be defined: $k+p$ and $k\cdot p$, having both of them a part inside the shape. Thus, a constant can be calculated for each line, and the tree can be replaced by either $k+p$ or $k\cdot p$. Only sum and multiplication operations are considered because on $k-p$ and $k/p$ the nodes $-p$ and $1/p$ are not tangent to the shape and it is less likely to find a line with an intersection with the shape.

As happens with constant-variable search, constant-expression search can hide a constant search. For example, if the tree is $(3+2\cdot x_1)$ and the operation $(k+p)$ is used, then the result would be $(k+(3+2\cdot x_1))$, which has a semantic equivalent to $(k+3)+(2\cdot x_1)$, and could be generated by a constant search on the first constant leaf.

One way to partially avoid this situation is not to perform constant-expression searches with the operation $(k+p)$ on non-terminal nodes with the operation "+" or "-", and one of the children is a constant. In the same way, constant-expression searches with the operation $(k\cdot p)$ on non-terminal nodes with the operation "*" or "/", and one of the children is a constant should be avoided too.

Even with these precautions, a constant-expression search can hide a constant search. For example, the tree $((2-x_4) + (4\cdot x_4))$ with the operation $(k+p)$, would result $k+((2-x_4)+(4\cdot x_4))$, which has a semantic equivalent to $((k+2)-x_4)+(4\cdot x_4)$. This last tree could be found by a constant search.

Another example could be the tree $(3+(2\cdot x_1))$, in which the node where the constant-expression search takes place is $p=(2\cdot x_1)$. In this case, a $(k+p)$ constant-expression search could take place and $p$ could be replaced by $(k+(2\cdot x_1))$. In this example, the final tree would be $(3+(k+(2\cdot x)))$, which has a semantic equivalent to $((3+k)+(2\cdot x))$. This tree could be found by means of a constant search.

Therefore, the same situation as in previous search happens. It is hard to know in advance when a constant-variable search or a constant-expression search will hide a constant search. As was already stated, this leads to the problem of having too large trees, and too early stopping of the algorithm. The same two solutions can be applied: performing an optimization of all of the constants before these searches, or perform a constant search at the same time.

\subsection{Deletion of parts of the tree}

\label{sec:deletionOfPartsOfTheTree}

The deletion of a branch of the tree implies replacing its father with its "sibling" branch. For instance, in the tree $(2\cdot x_1 + 3\cdot x_2)$, the deletion of the branch $3\cdot x_2$ leads to have $(2\cdot x_1 + ?)$ and then replace the root with the first child, having as result $2\cdot x_1$. In this example, this is equivalent to having performed a constant search in the branch $3\cdot x_2$ and having as result an improvement in MSE with the constant 0, leading to the tree $(2\cdot x_1 + 0)$. If the root were the operators of -, * or /, the corresponding constants would be 0, 1 and 1.

Therefore, in all of the cases deleting a branch of the tree is similar to performing a constant search in that branch, and having as result one of those constants, which is very unlikely to happen. For this reason, the deletion of parts of the tree is not considered in this method, because it can be performed by constant, variable and constant-variable searches.

\subsection{Optimization of constants}

\label{sec:constantOptimization}

As the tree is being built, each time a constant is generated, its value is the best for that tree. However, as the improvement process keeps going on, the tree will be modified. This makes that this previously calculated constant value will not be the best for the new tree and, therefore, this tree does not return the best possible result with that structure. This happens with all of the constants of the tree. Therefore, each time the tree is modified, an optimization of the constants could be performed in order to find their new best values. Also, it has been found that, after a tree has been modified, constant search almost always leads to have reduction in MSE for several iterations of only constant search.

For this reason, a constant optimization process is proposed. This process is simply based on examining the tree, taking all of the constant nodes and perform successively the following steps on each of them:

\begin{enumerate}

\item Calculate the equation in this node.
\item Perform a constant search, having as result a value of MSE reduction.
\item If this reduction of MSE is higher than a value (see section \ref{sec:algorithm} for details of this parameter), then modify the constant of the node with the result of the search.

\end{enumerate}

Once these steps are executed on a constant node, the process goes into the following one. When the last constant node has been processed, this process begins again with the first. This process finishes when none of the constant nodes can be optimized. Note that before that point is reached, it is possible that each node has been optimized several times.

A different constant optimization process could be considered, in which in each iteration a constant search is performed on all of the constant nodes, and the one with the highest reduction (in case it is higher than the parameter value) is modified. However, this would need the calculation of all of the equations of the constant nodes in each iteration. Since the calculation of an equation needs the calculation of the equations of the previous nodes and the constant nodes are always leaves of the tree, this implies the calculation of a high number of equations of the tree. This can be a time-demanding task. For this reason, the described approach is proposed instead, which is much more efficient, since only a small number of equations need to be calculated in each iteration. However, if the user wishes to perform a constant optimization process in which in each iteration the constant to be changed is the one with the highest MSE reduction, this can be done by repeately executing constant searches on constant nodes.

%For this reason, a constant optimization process is proposed. This process is simply based on performing several constant searches with two remarks:
%\begin{itemize}
%\item This search is performed only on constant nodes, which are the values to be optimized.
%\item If a constant value was changed in the previous constant search, this value is not being explored in the next constant search, in order to save computational time.
%\end{itemize}
% These consecutive constant searches are performed until the improvement in MSE is lower than a pre-fixed amount. 

As it is obvious, if there is only one constant in the tree, then only one iteration is performed.

The optimization of constants can be very useful to prevent some successful searches that in practice only modify the value of a constant, making an artificial growth of the tree. For instance, in the tree $(2+x)$ a constant-variable search can be performed at the variable $x$, having as result the tree $(2+(3+x))$. This tree returns the same results as $(5+x)$, which could be obtained simply with a constant search on the constant $2$. In this example, the constant-variable search result was hiding a constant search, making the tree unnecessarily larger. If a constraint to the size of the tree is used (see section \ref{sec:constraints}), then this growth of the tree makes a premature stop of the algorithm. An additional problem of having too large trees is that the computational time is increased because all of the searches are applied in a larger number of nodes.

As was already described, one way to prevent these situations is, before the selected searches take place, perform an optimization of all of the constants of the tree. If this is done, obviously constant search on constants should not be performed after the optimization of constants.

\subsection{Constraints to the tree}

\label{sec:constraints}

As one of the objectives of this work is to find simple expressions easy to analyse by a human, it is interesting to limit the complexity of these expressions. With this objective, two constraints are used: height of the tree and number of nodes. The user may make use of any of these two, or none.

Both restrictions make effect in the constant-variable and constant-expression searches described in section \ref{sec:constantVariableSearch}. In the other two searches there is no need to apply complexity constraints, because in  both searches the result would be a tree with the same or less complexity. Only on constant-variable and constant-expression searches the complexity of the tree is increased.

The application of these two constraints is very straightforward. If the maximum height constraint is being used, instead of performing these two searches in all of the nodes of the tree, they will be done only in the nodes whose depth is lower than the maximum height. If the maximum number of nodes (n) constraint is being used, the tree has r nodes and each node of the tree represents a subtree with s nodes, then the constant search is performed only on those nodes that meet the constraint $n-r+s\geq3$. The constant-expression search is performed if $n-r\geq2$

These two constraints are closely related. Since only binary operations, arithmetic functions, are used, the height constraint sets a limit to the number of nodes. The number of nodes of a tree of height h can be up to $2^{h-1}$. Moreover, in this tree these $2^{h-1}$ must be balanced, so a height limit is a constraint to the number of nodes, but also to the structure of these nodes in the tree. Also, it was already demonstrated that a binary tree with n nodes has as average a height of $2\sqrt{\pi n}$ \cite{FLAJOLET1982171}. Therefore, to allow the building of a tree with n nodes without any structure constraint, the maximum height should be higher than $2\sqrt{\pi n}$, which would lead to having more than $2^{2\sqrt{\pi n}-1}$ nodes, which is a very large number. For this reason, the height limit is not used, and the only constraint is the number of nodes.

Setting a limit on the complexity of the tree has two interesting features. First, it allows the obtaining simple easy-to-understand expressions. Second, it allows the obtaining models with better generalization abilities and controlling overfitting. The described system has the ability to grow until a good enough result on the training set is found. However, an excessive growth, resulting in a very large tree and a very complex model, may lead to overfitting the training set. for this reason, the user may set a limit to this growth in the maximum number of nodes that the tree can have. Thus, this parameter will be important and the experiments will be performed with it.

\subsection{Algorithm}

\label{sec:algorithm}

The method proposed in this section allows the creation of trees with a low MSE value. However, differently from GP and GSGP in which many different trees are created, in this method a simple tree is continuously improved.

The algorithm begins with a simple initial tree made of a single node representing a constant. This constant is the point in the constant line closer to the target point. To calculate it, set the constraint that the vectors k and k-t must be perpendicular, and therefore the point product $<k,k-t>=0$. Developing this expression leads to

\begin{equation}
\label{eq:eqInitialConstant}
0=\sum_{i=1}^N{k \cdot (k-t_i)}=\sum_{i=1}^N{k^2}-\sum_{i=1}^N{k \cdot t_i}=N \cdot {k^2}-k \cdot \sum_{i=1}^N{t_i}
\end{equation} 

And therefore $k=\frac{1}{N}\sum_{i=1}^N{t_i}$ is the average value of the targets. As this constant node is already the best value, it will not be optimised due to the constant search process. This value could also be obtained with a constant search from equation \ref{eq:eqDerivative1} ($a_i=1$, $b_i=t_i$, $c_i=0$, $d_i=-1$, $S=\emptyset$).

After this initial tree has been created, the iterative process begins. On each iteration, the constant search, variable search, constant-variable and constant-expression search can be performed on any nodes, whether they are constant, variable or non-terminal nodes, following a specific strategy. For instance, some possible strategies could be:

\begin{enumerate}

    \item Perform the four searches at the same time, and substitute the corresponding node with the result of the strategy with a higher reduction in MSE, in case it is positive. This is the strategy that makes a full search, however, it is also the most time-demanding strategy, since much computational time is wasted in performing operations that will not lead to improving the tree. IT is important to remark that in this strategy constant search is performed in constants, together with constant-variable search and constant-expression search, to avoid the mentioned problem of "hiding" changes in constants and tree growing.
    
    \item Perform the four searches at the same time, however, in constant search only perform this search in variables and non-terminal nodes. After any modification is done to the tree, perform the constant optimization process. This optimization ensures that the next constant-variable and constant-expression searches do not "hide" a change in constants.
    
    These two strategies can be seen as similar in the sense that in the second one a constant optimization step is performed on each iteration, while in the first one a constant search on the constant nodes of the tree is performed on each iteration. Therefore, in the first strategy successive iterations may perform a constant refinement similar to the constant optimization performed in the second strategy. However, this constant refinement performed in the first strategy will be stopped when any other search returns a higher MSE reduction, while in the second strategy the constant refinement is stopped when no MSE reduction is found in the constant optimization process.
    
    \item In order to save computational time, those searches that are found not to be successful most of the times can be avoided. Also, the searched can be performed sequentially, and when one of the searches is successful, the rest are not performed. This strategy proposes to run the following steps on each iteration:
    
    \begin{enumerate}
    
    \item Perform variable search only on constant nodes.
    
    \item If the previous search was not successful, perform constant-expression search (this search is always only performed on non-terminal nodes).
    
    \item If any of the two previous searches were not successful, perform constant-variable search on terminal nodes.

    \item If none of the three previous searches was successful, perform together:
    
    \begin{itemize}
    
    \item Constant search on variable and non-terminal nodes.
    \item Variable search on variable and non-terminal nodes.
    \item Constant-variable search on non-terminal nodes.
    
    \end{itemize}
    
    \item If any of the previous searches was successful, perform a constant optimization.
    
    \end{enumerate}
    
    The objective of these search is to perform those searches that are unlikely to be successful only when the rest are not successful. This usually happens when the tree has reached the maximum limit. As it can be seen, those searches that are performed when this happens are those that allow the tree to become smaller.

    \item In a similar way to the first and second strategies, this fourth strategy is like the third, but without the constant optimization process at the end. Instead of it, the first step of this strategy is a constant search on constant nodes (refinement of constant values). If a constant is modified, then no further searches are performed and this iteration is finished. If no constant is modified, then a strategy similar to strategy 3 is executed, except for the final constant optimization, which is not performed.

\end{enumerate}

These four strategies were used in the experiments because they describe two different situations: an exhaustive strategy in which in each iteration all of the searches are performed in all of the nodes, and a more light strategy in which in each iteration the while tree is examined only when no previous search was successful. In both strategies, their variants with and without constant optimization are explored too.

However, the user may decide to use other strategies he may think to be useful. Other examples are the following:

\begin{itemize}
    \item Perform the searches consecutively. On each of them, modify the tree accordingly.
    \item Perform constant and/or variable search and, if no modification is done to the tree, constant-variable search. If not modification is done, perform constant-expression search.
    \item Perform constant search. If no modification is done, perform variable search. If no modification is done in this second search, perform constant-expression search. If no modification is done in this third search, perform constant-variable search.
    
\end{itemize}

Constant search is the less computational expensive search, while constant-variable search is the most time consuming. So, a good strategy should keep a balance between high improvements and fast computation.

This process is iteratively performed until a stopping criteria is met. These criteria can be configured by the user. This work proposes the use of the following criteria:

\begin{itemize}
    \item The number of iterations exceeds a fixed value.
    \item The MSE in the current tree reaches the goal value set by the user.
    \item The tree could not be improved in the last iteration. A hyperparameter defining the minimum improvement to consider that a search has been successful is needed.
\end{itemize}

Therefore, this system can be configured with a low number of hyperparameters:

\begin{itemize}
    \item Maximum number of iterations to be executed. Default value: infinite.
    \item Goal in MSE. Default value: 0.
    \item Minimum of improvement in the MSE for any search. A search is found to be successful if the reduction in MSE is positive and higher to the previous MSE value multiplied by this parameter. For example, if this parameter takes a value of $10^{-2}$, then a search will be successful only if the reduction in MSE is greater than 1\% of the MSE. This value is also applied to the constant optimization process described in section \ref{sec:constantOptimization}. Default value: $10^{-6}$.
    %\item Maximum height of the tree. Default value: infinite,
    \item Maximum number of nodes of the tree. Default value: infinite,
    \item Strategy to be used on each iteration.
\end{itemize}

Although the execution of this system involves a great number of calculations, these can be efficiently performed with element-wise operations and parallel executions. element-wise operations can be performed on the evaluation of the equations of each node and in the expressions of the search for constants. Parallel code execution can be performed in the three search processes, because they involve the evaluation of functions in all of the nodes of the tree. Therefore, the use of Single Instruction, Multiple Data (SIMD) programming \cite{Cardoso:2017:ECH:3131535} can be of a great benefit for this algorithm.

An important feature of this algorithm is that it is deterministic, i.e., each time it is run with the same dataset it will return the same result. This makes that the number of times it has to be run for each parameter configuration is only one.

This algorithm was implemented in Julia \cite{bezanson2017julia}. The source code of this technique will be provided so all of the experiments described in this paper can be repeated, and the system can be used by anyone to perform their own experiments or make use of it in their researches.

%%%%%%%%%%%%%%%%%%%%%%%%%%%%%%%%%%%%%%%%%%%%%%%%%%%%%%%%%%%%%%%%%%%%%%%%%%%%%%%%

\section{Example of equation development}
\label{sec:Example}

This section describes the application of this system in order to develop an easy and well-known expression: Newton's law of universal gravitation \cite{newton1687philosophiae}. This equation measures the force of attraction between two masses. This force is proportional to the product of both masses, and inversely proportional to the square of their distance. This equation is given by \ref{eq:eqNewton}, in which $M_1$ and $M_2$ are the masses, $d$ is the distance, and $G=6.67392 \cdot 10^{-11}$ is the gravitational constant.

\begin{equation}
\label{eq:eqNewton}
F=G \cdot \frac{M_1 \cdot M_2}{d^2}
\end{equation}

This equation is of application to arbitrarily large or small masses and/or distances, i.e., it can be applied either to the force between small particles and big planets. In this example, this second case will be explored, in order to show that the system can return good results with arbitrarily different values, since masses and distances belong to different ranges.

In order to apply this equation, 1000 random patterns were built. Each pattern is composed by:

\begin{itemize}
    \item $x_1$: Mass of the first planet: a random number between $10^{23}$ and $10^{25}$
    \item $x_2$: Mass of the second planet: a random number between $10^{23}$ and $10^{25}$
    \item $x_3$: Distance between both planets: a random number between $10^{8}$ and $10^{12}$
    \item target: the result of the application of equation \ref{eq:eqNewton}
\end{itemize}

With this dataset, the system was configured with the following hyperparameters:

\begin{itemize}
    \item Maximum height of the tree: Infinite
    \item Maximum number of nodes: Infinite
    \item Strategy: The first strategy described on section \ref{sec:algorithm}
    \item Goal in MSE. This parameter is found to be very important in this example, since we are working with values in a very wide range, from very small ($G$) to very large ($x_1$ and $x_2$). The targets were also in a large scale, from $10^{12} N$ to $10^{22} N$. Due to the limitation of the use of floating point values, even when 64 bits are used for their storage, when working with such a wide range, unaccuracies may occur due to operating with numbers on very different scales. For instance, the operation $10^{20}+10^{-10}-10^{20}$ returns an incorrect value of 0. This may result that, even the correct expression has been obtained, the MSE calculated is not 0 due to unaccuracies, and the system keeps trying to improve this expression with inaccurate calculations. For this reason, a goal MSE of 0 was not used in this example. Instead of it, the goal used was the average value of the targets. 
\end{itemize}

Table \ref{tableNewton} shows a summary of an execution of this system with the described dataset. As was described, before the execution is performed, an initial tree with a single constant node is built. In this execution, this constant had as value $9.185106275464827\cdot10^{20}$. From this initial tree, 4 iterations were performed until an expression with a MSE below the goal was found. All of the operations performed were constant-variable searches, and the nodes selected for substitutions were the constants. This example also shows the ability of this system to develop the constants needed for the expressions to be returned. For this reason, the constants are shown on this table with a large number of decimals.

Finally, as a result of this execution, the expression $((((6.673919999999998\cdot10^{-11}/x_3)\cdot x_1)\cdot x_2)/x_3)$ is returned by the system, which is similar to equation \ref{eq:eqNewton}.

\begin{table}
\centering
\caption{Summary of the execution for developing Newton's law of universal gravitation}
{
    \begin{tabular}{@{}ccccc@{}} 
    \toprule
    Iteration & \multicolumn{1}{c}{Node selected} & \multicolumn{1}{c}{Resulting expression} & \multicolumn{1}{c}{MSE} \\
              & \multicolumn{1}{c}{for substitution} & \multicolumn{1}{c}{} \\
    \midrule
        0 & - & $9.185106275464827\cdot10^{20}$ & $2.7645\cdot10^{43}$ \\
        1 & $9.185106275464827\cdot10^{20}$ & $(1.1526861538137098\cdot10^{30}/x_3)$ & $2.1717\cdot10^{43}$ \\
        2 & $1.1526861538137098\cdot10^{30}$ & $((434063.5718753305\cdot x_2)/x_3)$ & $1.9183\cdot10^{43}$ \\
        3 & $434063.5718753305$ & $(((3.932029293203675\cdot10^{-19}\cdot x_1)\cdot x_2)/x_3)$ & $5.1733\cdot10^{42}$ \\
        4 & $3.932029293203675\cdot10^{19}$ & $((((6.673919999999998\cdot10^{-11}/x_3)\cdot x_1)\cdot x_2)/x_3)$ & $1.3863\cdot10^{12}$
%    \bottomrule
    \end{tabular}
}
\label{tableNewton}
\end{table}

%%%%%%%%%%%%%%%%%%%%%%%%%%%%%%%%%%%%%%%%%%%%%%%%%%%%%%%%%%%%%%%%%%%%%%%%%%%%%%%%

\section{Experiments}
\label{sec:Experiments}

\begin{figure}
\centering

    \begin{subfigure}[b]{.33\linewidth}
        \centering
        \includegraphics[width=.99\textwidth]{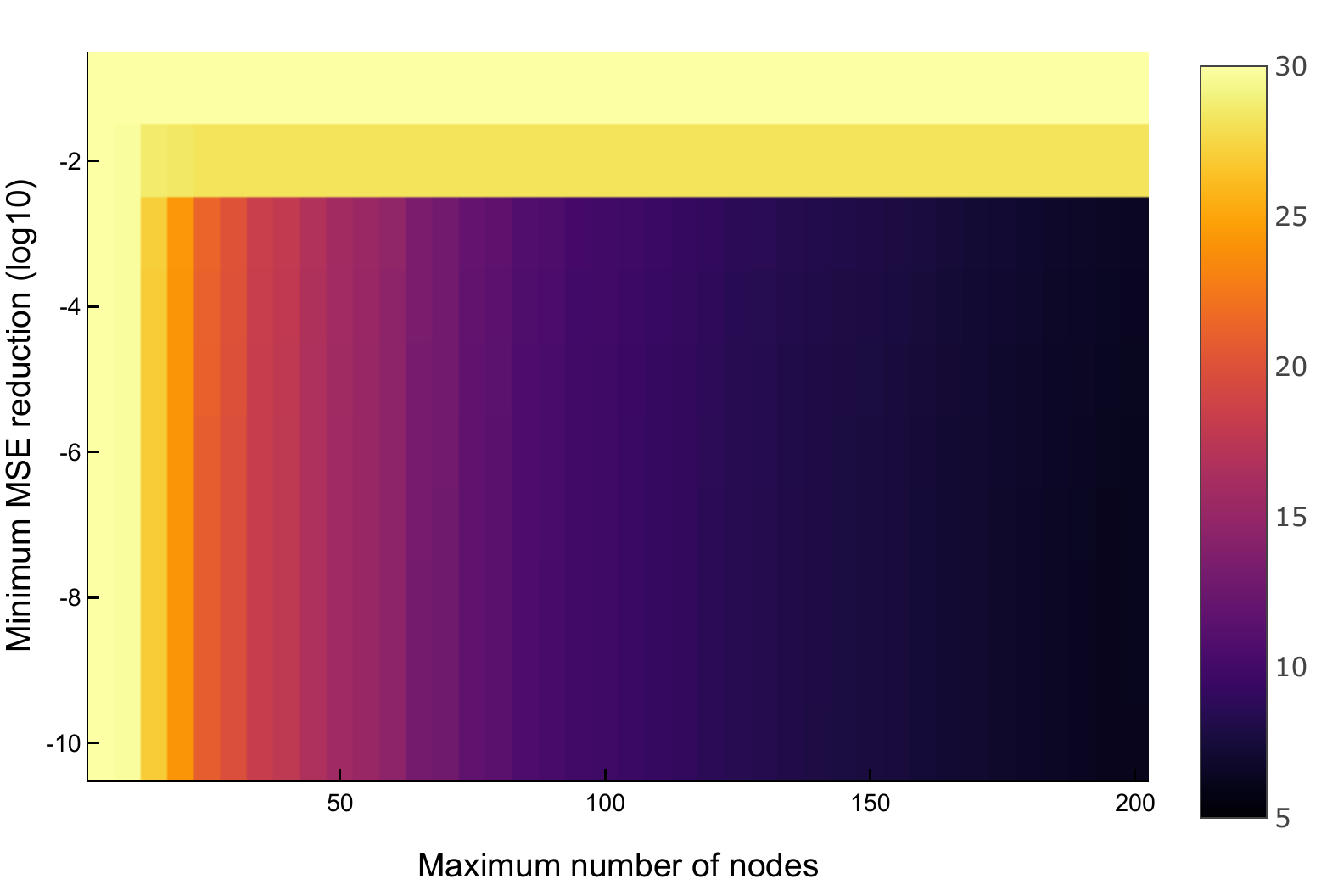}
        \caption{Mean training, strategy 1}
        \label{fig:housing_strategy1_numNodes_heatmapMSEMeanTraining}
    \end{subfigure}
    \begin{subfigure}[b]{.33\linewidth}
        \centering
        \includegraphics[width=.99\textwidth]{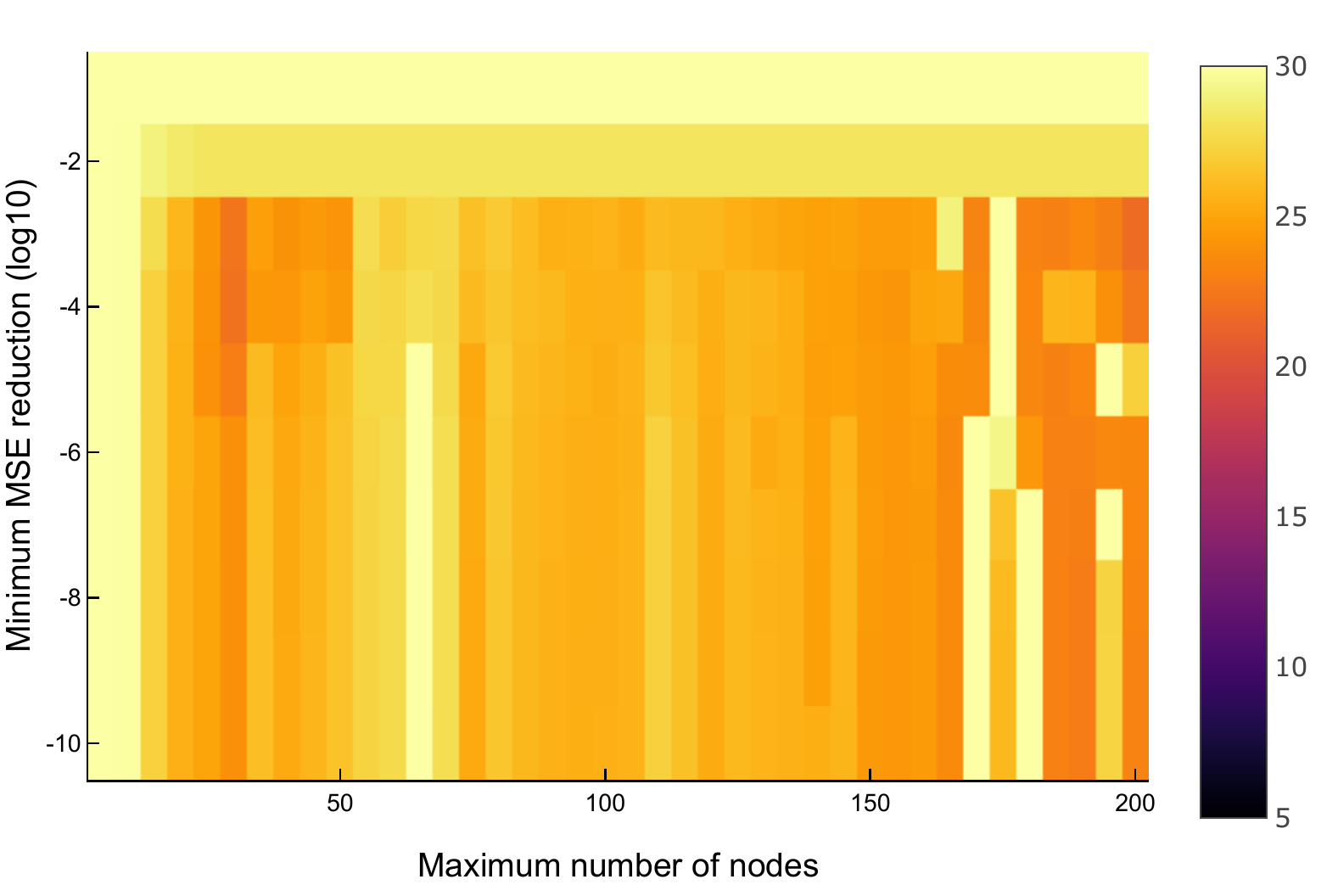}
        \caption{Mean test, strategy 1}
        \label{fig:housing_strategy1_numNodes_heatmapMSEMeanTest}
    \end{subfigure}
    \begin{subfigure}[b]{.33\linewidth}
        \centering
        \includegraphics[width=.99\textwidth]{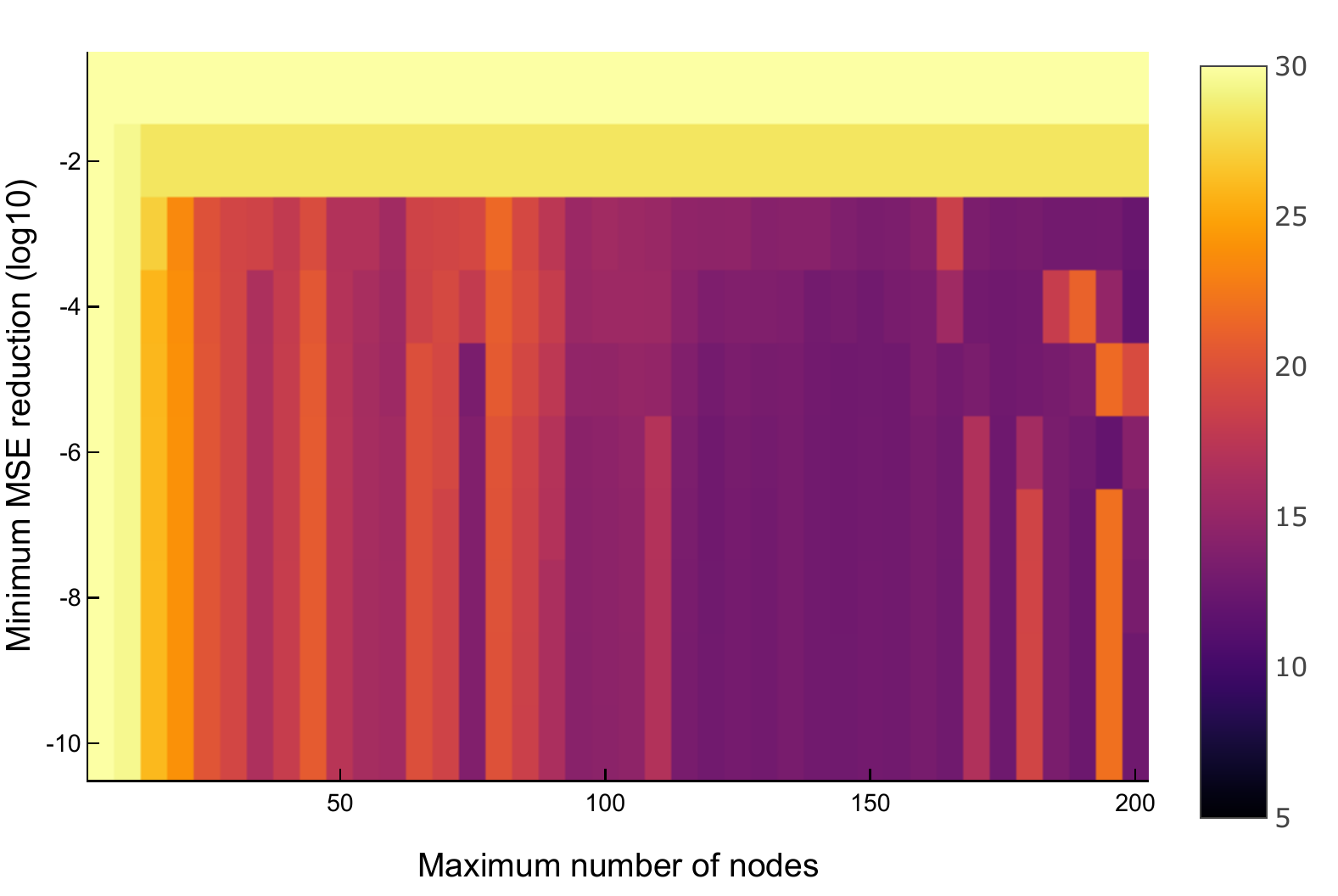}
        \caption{Median test, strategy 1}
        \label{fig:housing_strategy1_numNodes_heatmapMSEMedianTest}
    \end{subfigure}
    % To put more figures below, just add \\
    \\
    \begin{subfigure}[b]{.33\linewidth}
        \centering
        \includegraphics[width=.99\textwidth]{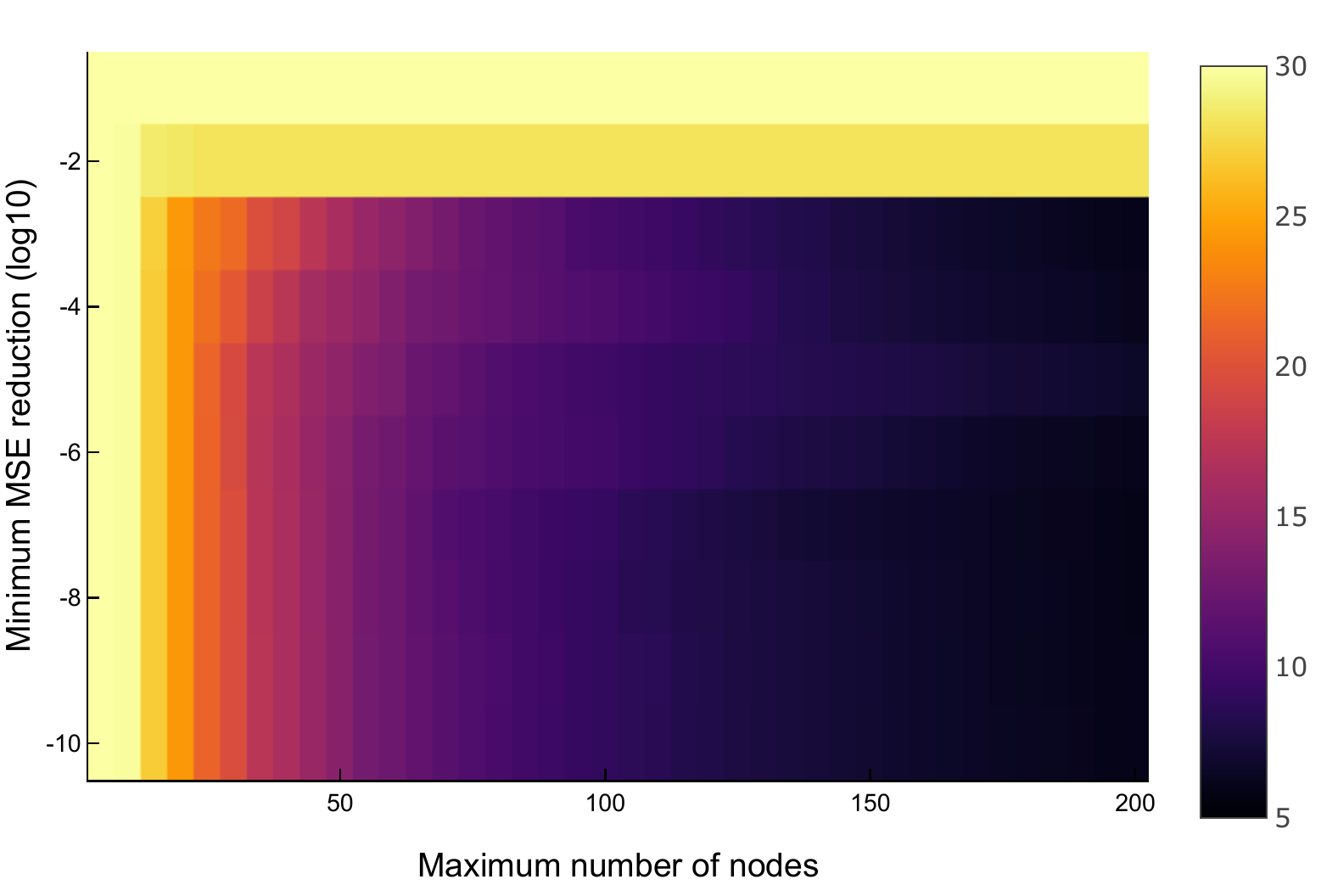}
        \caption{Mean training, strategy 2}
        \label{fig:housing_strategy2_numNodes_heatmapMSEMeanTraining}
    \end{subfigure}
    \begin{subfigure}[b]{.33\linewidth}
        \centering
        \includegraphics[width=.99\textwidth]{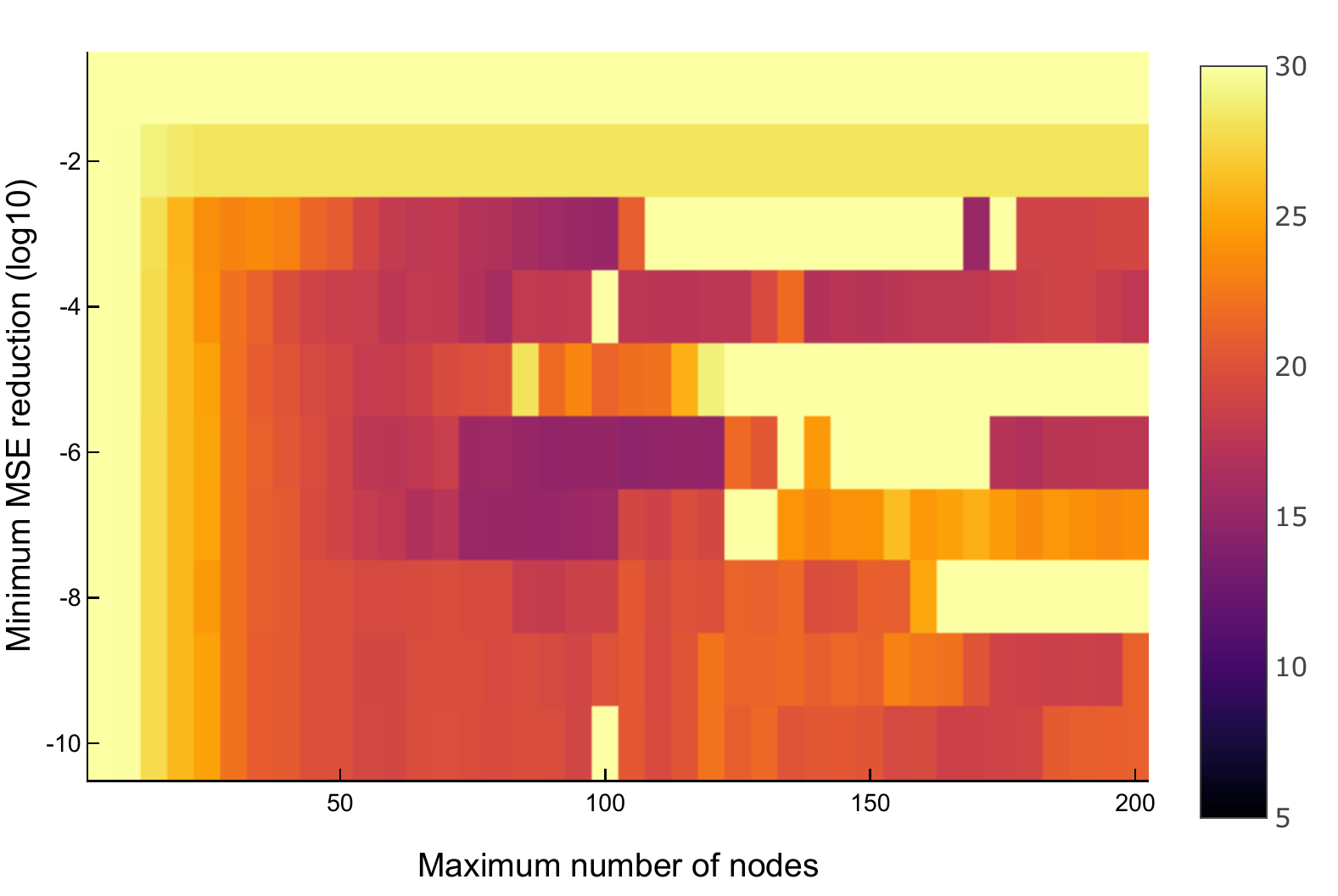}
        \caption{Mean test, strategy 2}
        \label{fig:housing_strategy2_numNodes_heatmapMSEMeanTest}
    \end{subfigure}
    \begin{subfigure}[b]{.33\linewidth}
        \centering
        \includegraphics[width=.99\textwidth]{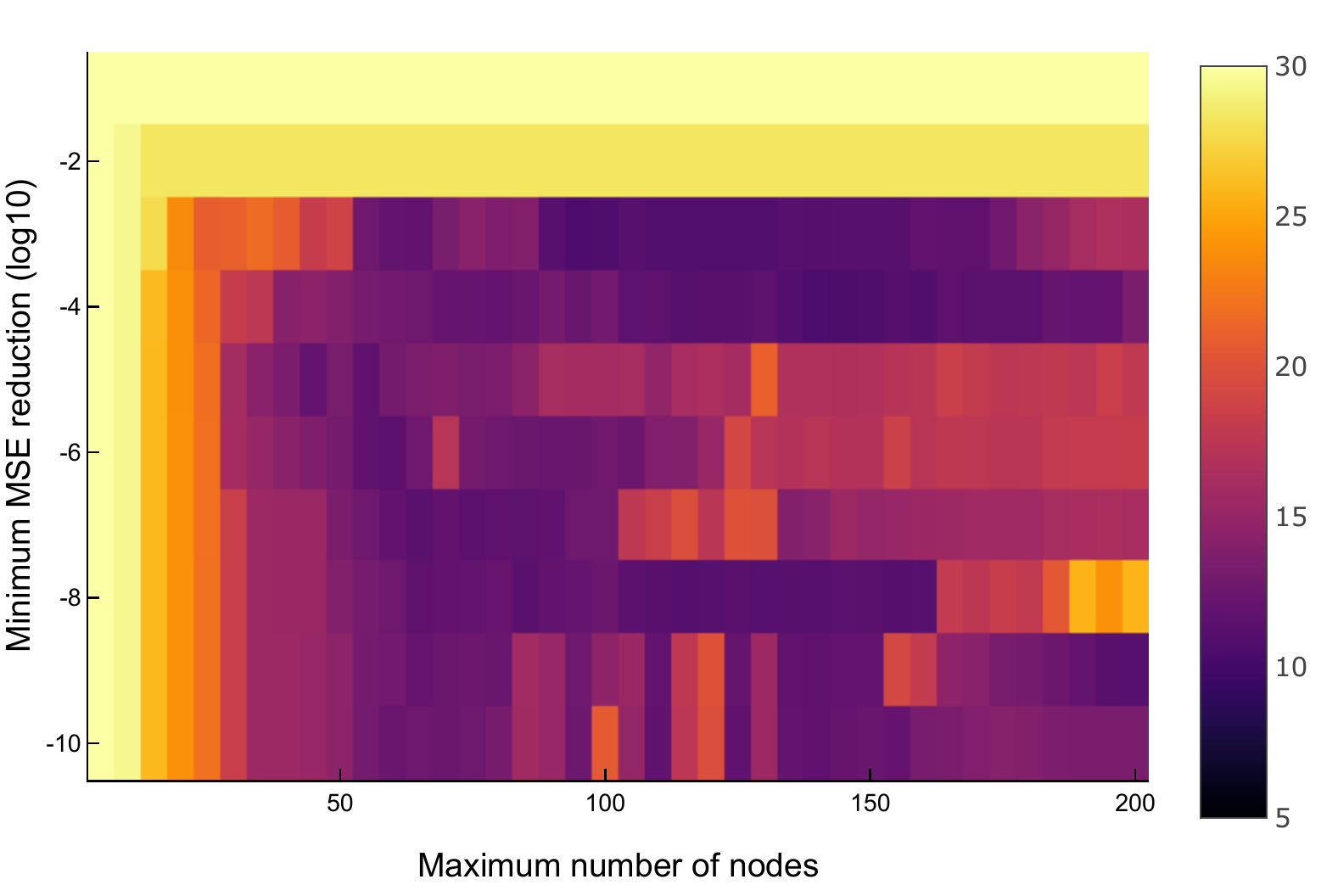}
        \caption{Median test, strategy 2}
        \label{fig:housing_strategy2_numNodes_heatmapMSEMedianTest}
    \end{subfigure}
    \\
    \begin{subfigure}[b]{.33\linewidth}
        \centering
        \includegraphics[width=.99\textwidth]{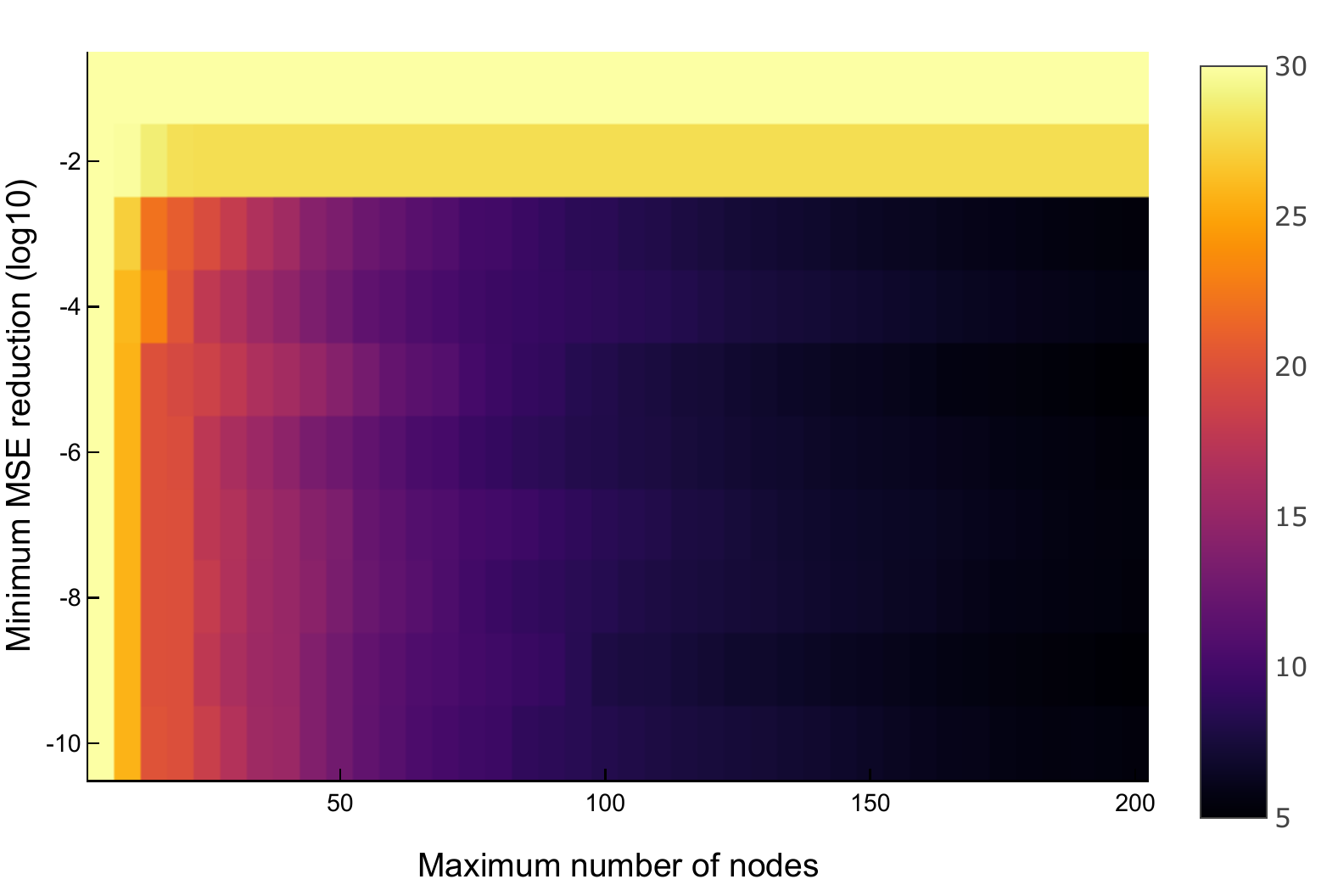}
        \caption{Mean training, strategy 3}
        \label{fig:housing_strategy3_numNodes_heatmapMSEMeanTraining}
    \end{subfigure}
    \begin{subfigure}[b]{.33\linewidth}
        \centering
        \includegraphics[width=.99\textwidth]{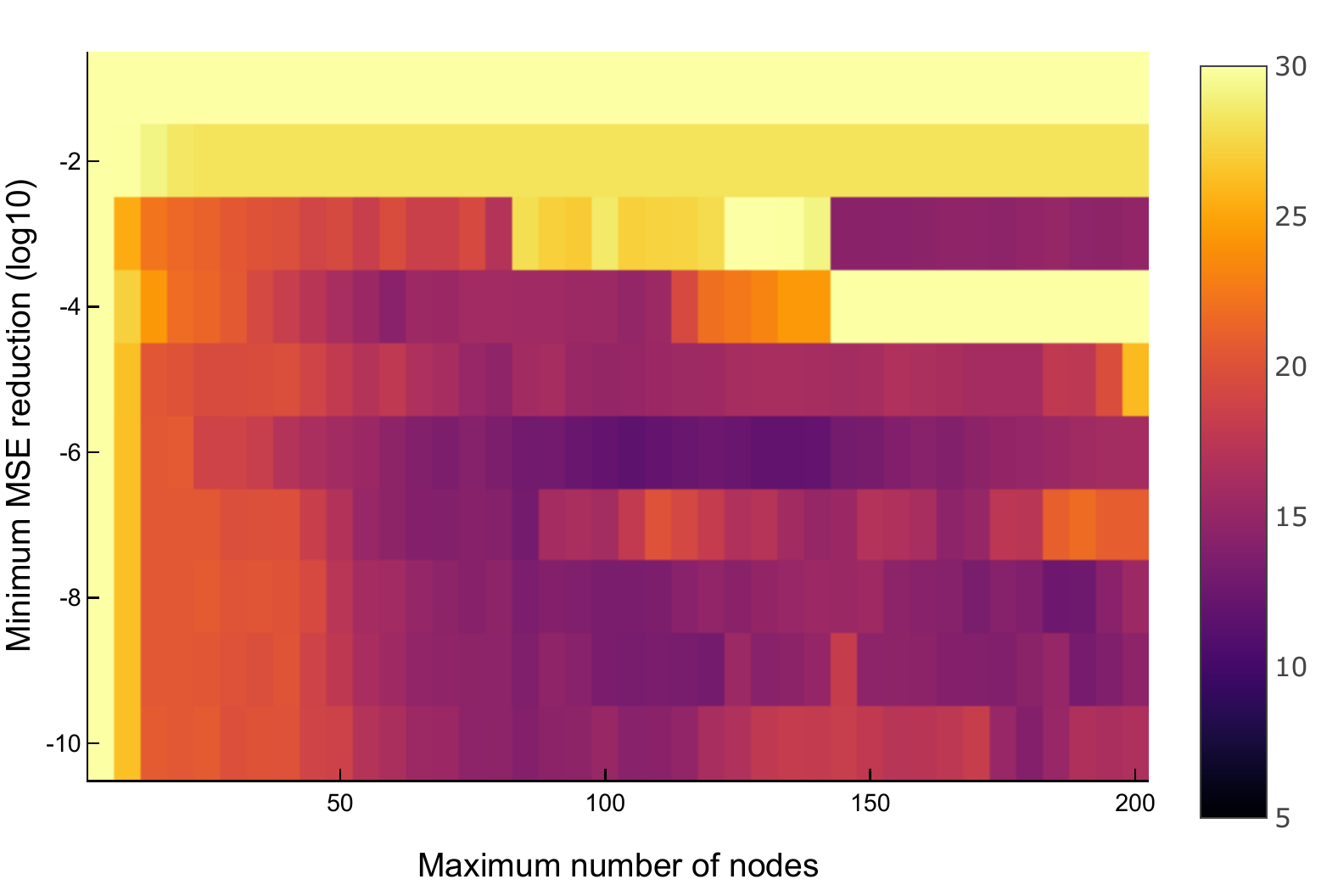}
        \caption{Mean test, strategy 3}
        \label{fig:housing_strategy3_numNodes_heatmapMSEMeanTest}
    \end{subfigure}
    \begin{subfigure}[b]{.33\linewidth}
        \centering
        \includegraphics[width=.99\textwidth]{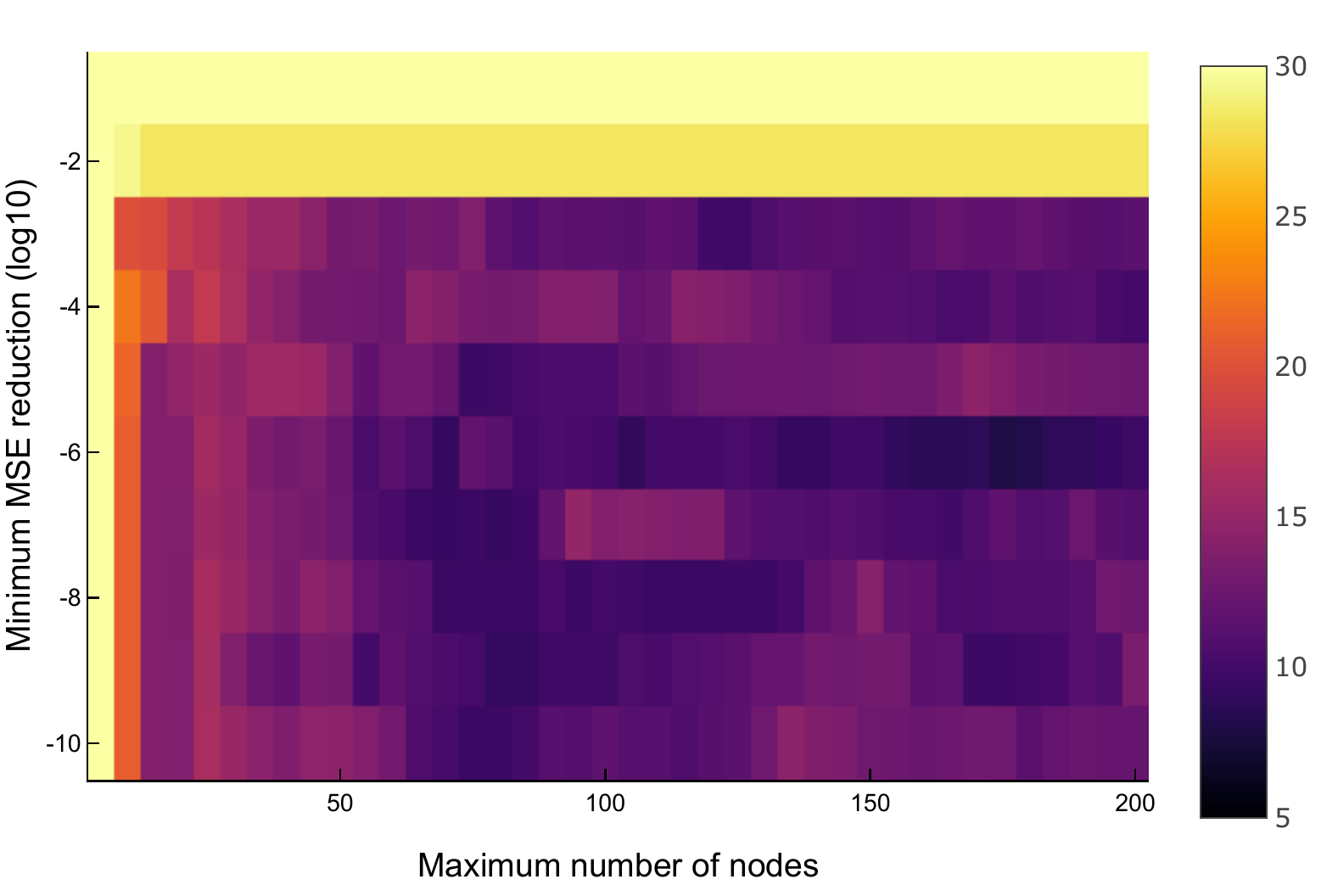}
        \caption{Median test, strategy 3}
        \label{fig:housing_strategy3_numNodes_heatmapMSEMedianTest}
    \end{subfigure}
    \\
    \begin{subfigure}[b]{.33\linewidth}
        \centering
        \includegraphics[width=.99\textwidth]{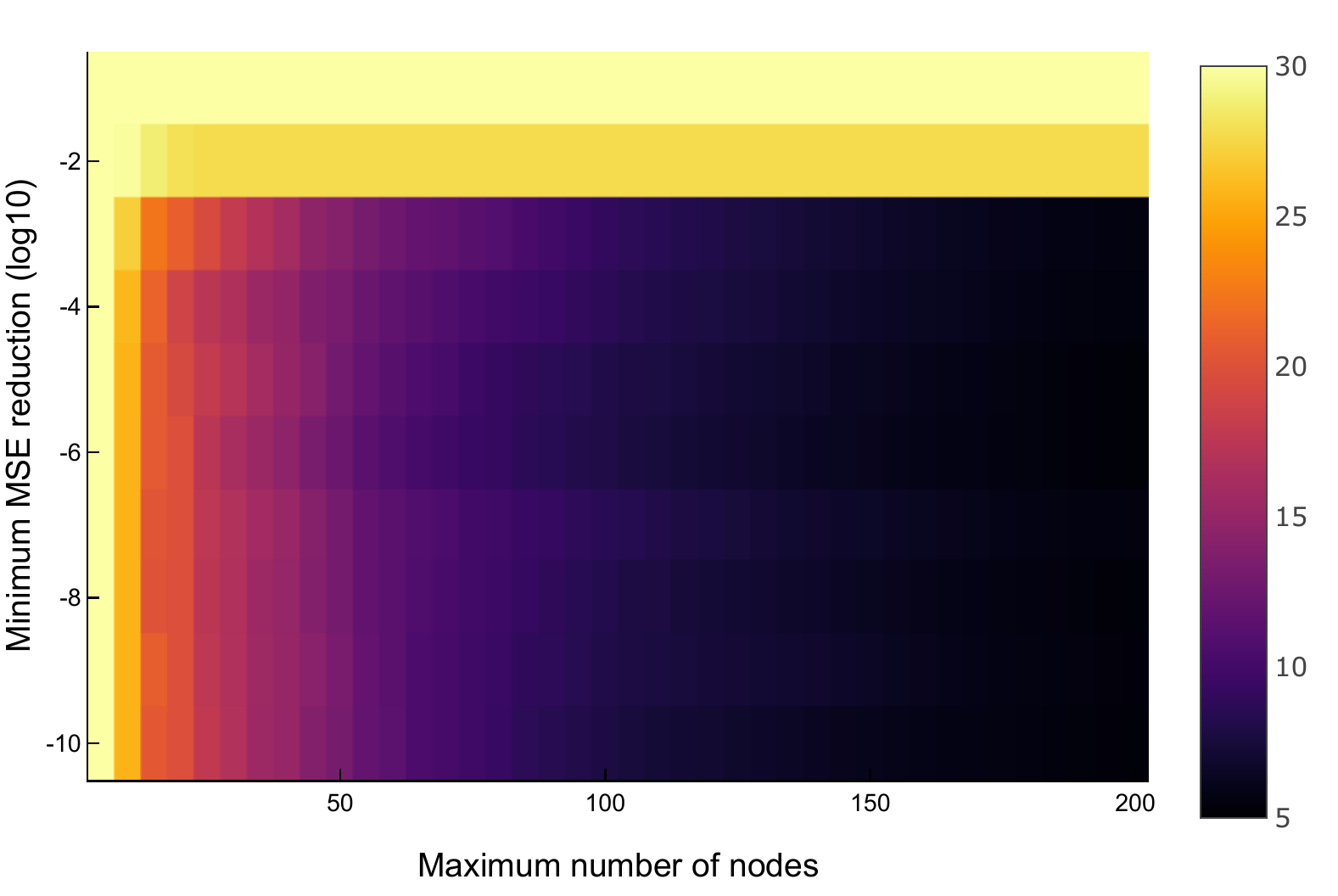}
        \caption{Mean training, strategy 4}
        \label{fig:housing_strategy4_numNodes_heatmapMSEMeanTraining}
    \end{subfigure}
    \begin{subfigure}[b]{.33\linewidth}
        \centering
        \includegraphics[width=.99\textwidth]{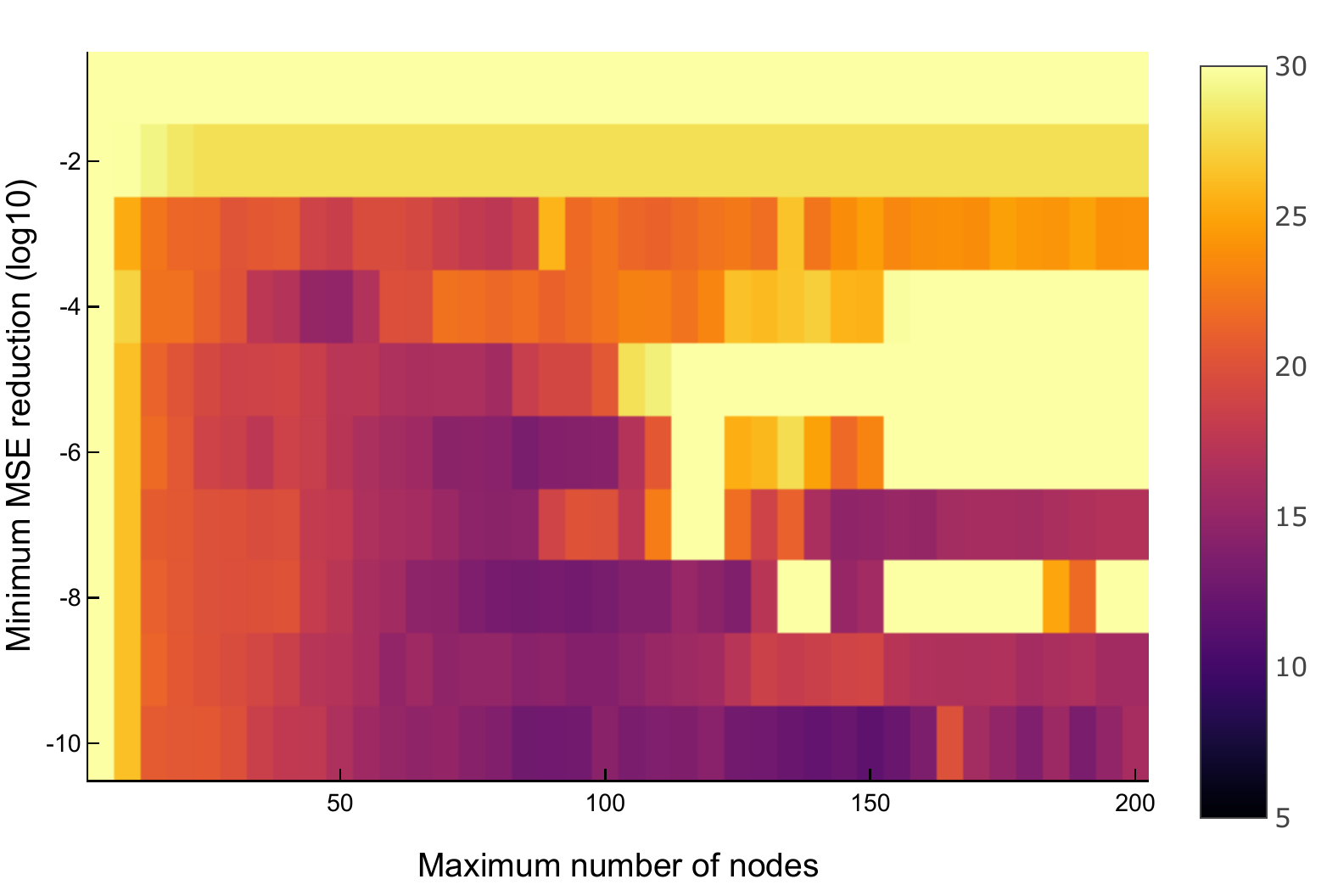}
        \caption{Mean test, strategy 4}
        \label{fig:housing_strategy4_numNodes_heatmapMSEMeanTest}
    \end{subfigure}
    \begin{subfigure}[b]{.33\linewidth}
        \centering
        \includegraphics[width=.99\textwidth]{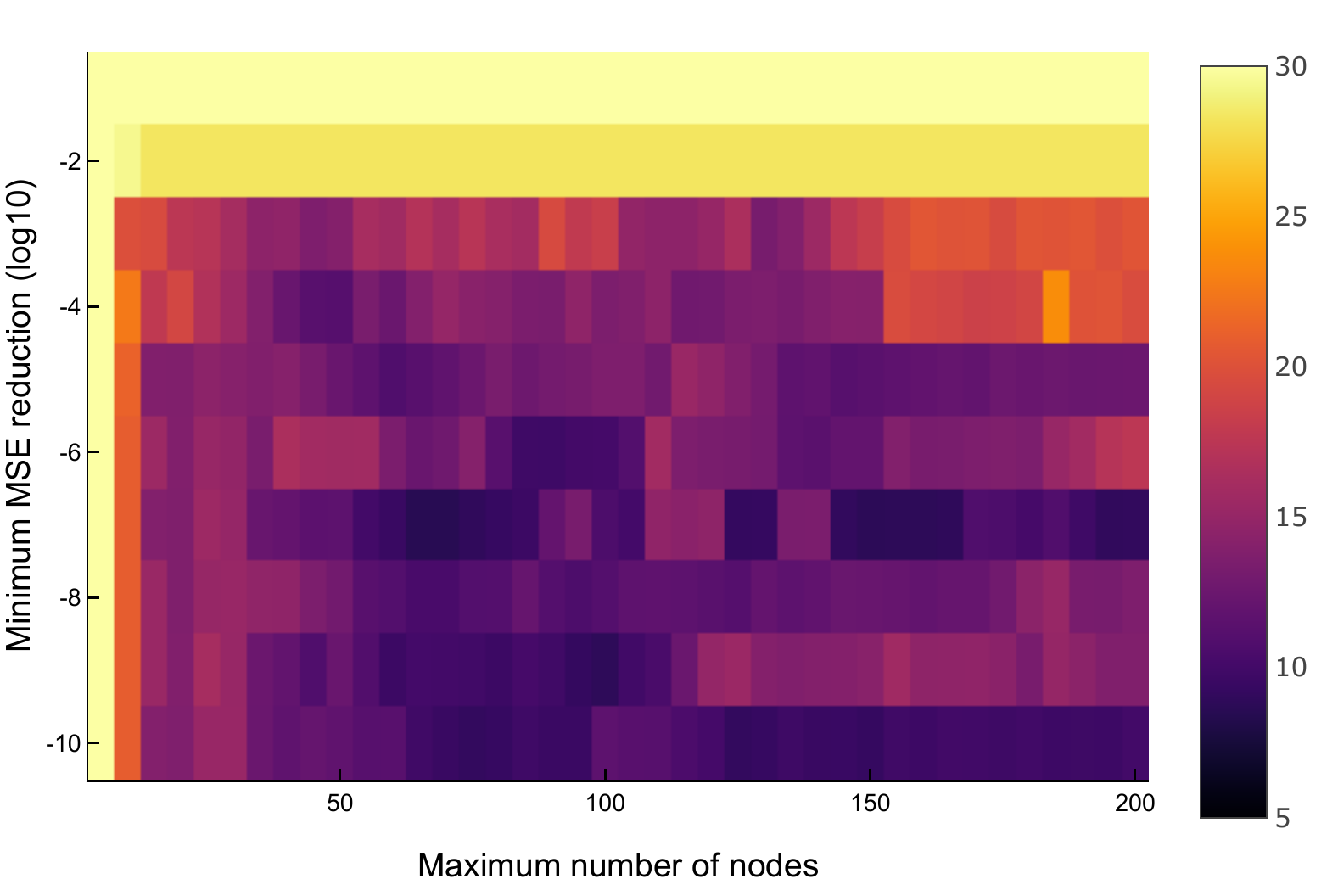}
        \caption{Median test, strategy 4}
        \label{fig:housing_strategy4_numNodes_heatmapMSEMedianTest}
    \end{subfigure}

    \caption{MSE results in training and test for the housing problem}
    \label{fig:housing_numNodes_heatmapMSE}
\end{figure}

\begin{figure}
    \begin{subfigure}[b]{.33\linewidth}
        \centering
        \includegraphics[width=.99\textwidth]{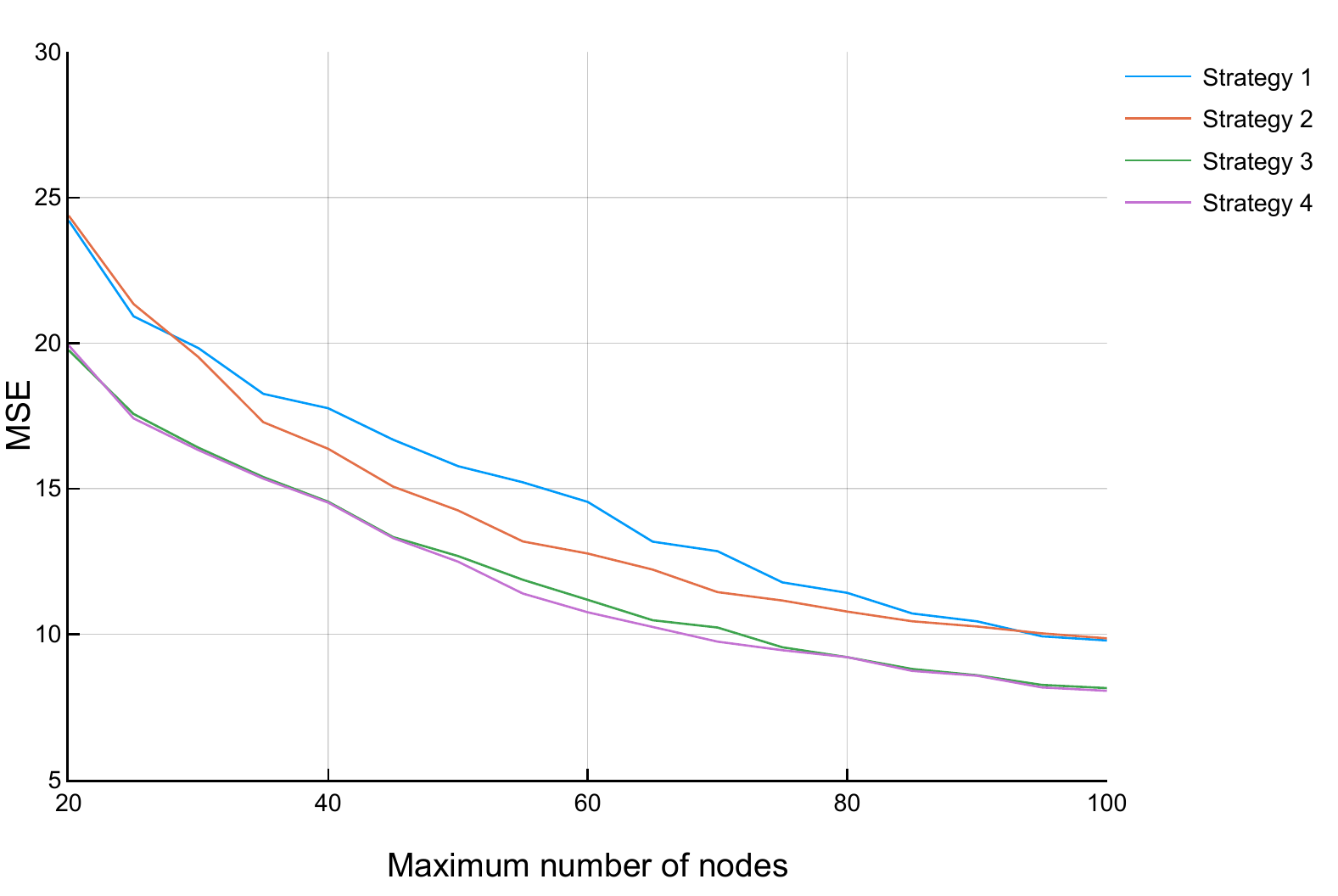}
        \caption{Mean training}
        \label{fig:housing_numNodes_plotMSEMeanTraining}
    \end{subfigure}     \begin{subfigure}[b]{.33\linewidth}
        \centering
        \includegraphics[width=.99\textwidth]{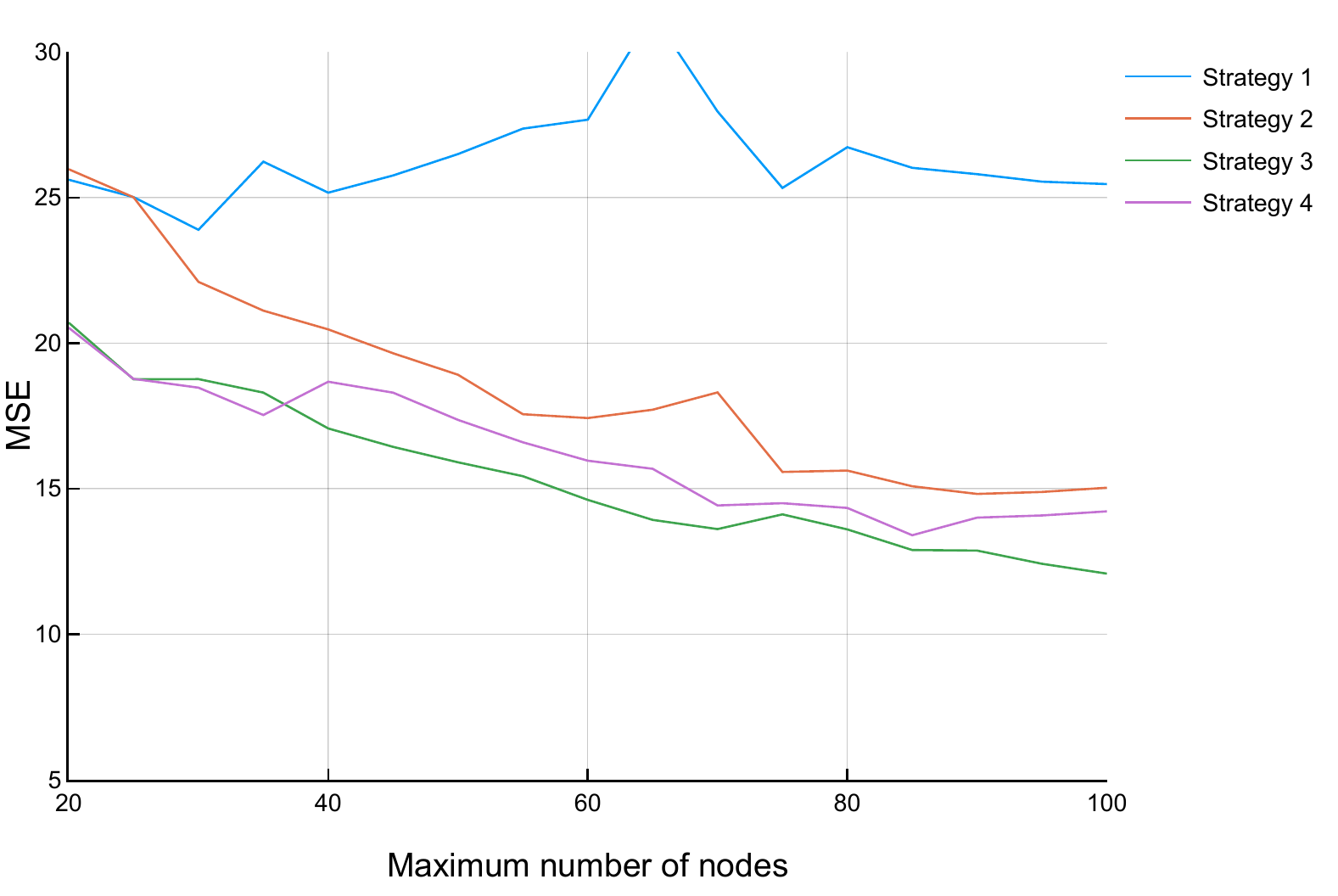}
        \caption{Mean test}
        \label{fig:housing_numNodes_plotMSEMeanTest}
    \end{subfigure}
    \begin{subfigure}[b]{.33\linewidth}
        \centering
        \includegraphics[width=.99\textwidth]{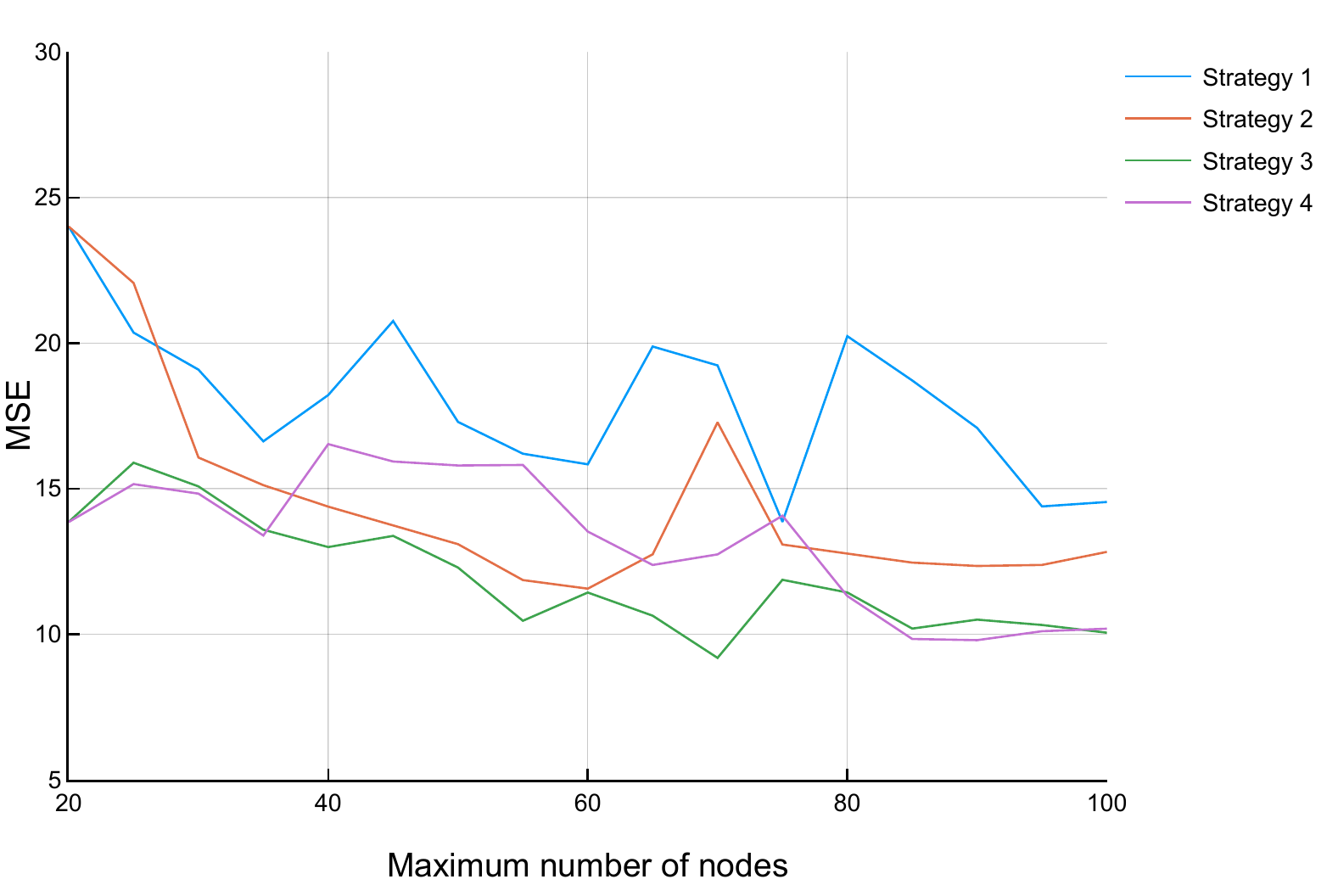}
        \caption{Median test}
        \label{fig:housing_numNodes_plotMSEMedianTest}
    \end{subfigure}
    \caption{MSE results in training and test for the housing problem}
    \label{fig:housing_numNodes_plotMSE}
\end{figure}

\begin{table}
\centering
\caption{Configurations that returned the best MSE in the training sets}
{
    \begin{tabular}{@{}ccccc@{}} 
    \toprule
    Strategy & \multicolumn{1}{c}{Minimum average} & \multicolumn{1}{c}{Minimum MSE} & \multicolumn{1}{c}{Maximum} & \multicolumn{1}{c}{Test} \\
              & \multicolumn{1}{c}{MSE training} & \multicolumn{1}{c}{reduction} & \multicolumn{1}{c}{number of nodes} & \multicolumn{1}{c}{MSE} \\
    \midrule
        1 & $6.218$ \tiny$(0.733)$\normalsize & $10^{-10}$ & 195 & $27.366$ \tiny$(18.874)$\normalsize \\
        
        2 & $5.954$ \tiny$(0.621)$\normalsize & $10^{-8}$ & 200 & $50.707$ \tiny$(64.974)$\normalsize \\
        
        3 & $4.933$ \tiny$(0.485)$\normalsize & $10^{-5}$ & 200 & $26.096$ \tiny$(38.273)$\normalsize \\
        
        4 & $5.217$ \tiny$(0.561)$\normalsize & $10^{-6}$ & 200 & $132.638$ \tiny$(341.447)$\normalsize
%    \bottomrule
    \end{tabular}
}
\label{housing_summaryTraining}
\end{table}

\begin{table}
\centering
\caption{Configurations that returned the best MSE in the test sets}
{
    \begin{tabular}{@{}ccccc@{}} 
    \toprule
    Strategy & \multicolumn{1}{c}{Minimum average} & \multicolumn{1}{c}{Minimum MSE} & \multicolumn{1}{c}{Maximum} & \multicolumn{1}{c}{Training} \\
              & \multicolumn{1}{c}{MSE test} & \multicolumn{1}{c}{reduction} & \multicolumn{1}{c}{number of nodes} & \multicolumn{1}{c}{MSE} \\
    \midrule
        1 & $21.782$ \tiny$(17.435)$\normalsize & $10^{-3}$ & 200 & $6.496$ \tiny$(0.629)$\normalsize \\

        2 & $14.579$ \tiny$(7.026)$\normalsize & $10^{-6}$ & 105 & $9.565$ \tiny$(1.374)$\normalsize \\
        
        3 & $11.644$ \tiny$(6.624)$\normalsize & $10^{-6}$ & 105 & $7.902$ \tiny$(0.556)$\normalsize \\

        4 & $11.725$ \tiny$(6.075)$\normalsize & $10^{-10}$ & 150 & $6.145$ \tiny$(0.645)$\normalsize
%    \bottomrule
    \end{tabular}
}
\label{housing_summaryTest}
\end{table}

\begin{table}
\centering
\caption{Summary of the results with a MLP}
{
    \begin{tabular}{@{}ccc@{}} 
    \toprule
    Epoch & \multicolumn{1}{c}{Training MSE} & \multicolumn{1}{c}{Test MSE} \\
    \midrule
        4347 & $4.827$ \tiny$(0.125)$\normalsize & $12.399$ \tiny$(6.869)$\normalsize \\
        49926 & $1.739$ \tiny$(0.065)$\normalsize & $19.791$ \tiny$(14.779)$\normalsize
%    \bottomrule
    \end{tabular}
}
\label{housing_summaryMLP}
\end{table}

The previous section showed that the system described in section \ref{sec:Model} can be successfully used as tool for developing mathematical expressions. These mathematical expressions can also be used as a ML model. This section describes the experiments carried out with this objective, on a real-world problem. This is a well-known problem, in which the objective is to predict the median value of a home in the area of Boston Mass. The information was collected by the U.S Census Service and is available  from the StatLib archive (http://lib.stat.cmu.edu/datasets/boston). This dataset was originally described at \cite{Harrison1978HedonicPA}. From this publication, it has been extensively used throughout the literature as  benchmark. This dataset has only 506 data points, and 13 input variables.

%The experiments carried out in this part have as objective to show the applicability of this technique. In this sense, as shown on section \ref{sec:constraints}, it is important to limit the complexity of the trees. For this reason, the experiments will be focused on studying the impact of the two parameters described in section \ref{sec:constraints} that limit the size of the tree: maximum height and maximum number of nodes. These two hyperparameters will be studied separately.

The experiments carried out in this part have as objective to show the applicability of this technique. In this sense, as shown on section \ref{sec:constraints}, it is important to limit the complexity of the trees. For this reason, the experiments will be focused on studying the impact of the parameters described in section \ref{sec:constraints} that limits the size of the tree: maximum number of nodes.

Also, the hyperparameter of minimum improvement in MSE plays an important role. For this reason, it was chosen for performing the experiments. The values chosen for this parameter are $10^{-1}$, $10^{-2}$, $10^{-3}$, $10^{-4}$, $10^{-5}$, $10^{-6}$, $10^{-7}$, $10^{-8}$, $10^{-9}$ and $10^{-10}$.

The experiments done in this section were run with the four strategies explained in section \ref{sec:algorithm}.

For each parameter value studied, a 10-fold cross-validation was performed. Since this system is deterministic, only one training in each fold was done. The results shown here are the average of the training results and average and median values of the test results. In order to correctly compare the results, in all of the experiments performed, the same training and test sets were used.

An important measure of the performance of this system is the computational time. All of the the experiments were run on a Intel(R) Xeon(R) CPU E5-2650 v3, with a frequency of 2.30GHz. A measure of time was performed on each of the experiments performed.

A grid search technique \cite{raschka2015python} was used to find the best combination of minimum MSE improvement and maximum number of nodes. Figure \ref{fig:housing_numNodes_heatmapMSE} shows 12 heatmaps with the MSE results of these experiments. On each heatmap figure, in the x axis the maximum number of nodes is situated, and in the y axis the minimum reduction is located. In order to show this graph more easily, this hyperparameter is shown as $log_{10}$.  Also, those MSE values higher than 35 or lower than 5 were cropped to those limits.

First, figures \ref{fig:housing_strategy1_numNodes_heatmapMSEMeanTraining}, \ref{fig:housing_strategy2_numNodes_heatmapMSEMeanTraining}, \ref{fig:housing_strategy3_numNodes_heatmapMSEMeanTraining} and \ref{fig:housing_strategy4_numNodes_heatmapMSEMeanTraining} show, for each strategy, the heatmaps for the average training MSEs obtained for the different values of the two hyperparameters. For each combination of the two hyperparameters, these figures show the average of the 10 MSE training values obtained from the cross-validation process. As it was expected, as the maximum number of nodes is increased, the complexity of the resulting expression grows too, and it is able to better fit the training set, having a lower MSE. However, this has the risk of overfitting the training set. For this reason, figures \ref{fig:housing_strategy1_numNodes_heatmapMSEMeanTest}, \ref{fig:housing_strategy2_numNodes_heatmapMSEMeanTest}, \ref{fig:housing_strategy3_numNodes_heatmapMSEMeanTest} and \ref{fig:housing_strategy4_numNodes_heatmapMSEMeanTest} show, for each strategy, the average results of the 10 folds in the test sets. In some cases very high values in MSE were obtained. This shows that this system can be very sensitive to outliers. This also has a big impact when calculating the average of 10 test values. Therefore, in order to minimize this impact, median results on the test sets can be reported. These results are shown on figures \ref{fig:housing_strategy1_numNodes_heatmapMSEMedianTest}, \ref{fig:housing_strategy2_numNodes_heatmapMSEMedianTest}, \ref{fig:housing_strategy3_numNodes_heatmapMSEMedianTest} and \ref{fig:housing_strategy4_numNodes_heatmapMSEMedianTest}, one for each strategy. As it can be seen, the best results in test were not obtained with a very high number of nodes, i.e., with very high complex expressions. It seems that a number of nodes between 20 and 100 returns the best results in test for this problem. Regarding the minimum MSE reduction hyperparameter, no clear tendency can be concluded from these results.

Figures on \ref{fig:housing_numNodes_plotMSE} show a comparison between the MSE results in the different strategies. These figures show the average values in training (fig. \ref{fig:housing_numNodes_plotMSEMeanTraining}), test (fig. \ref{fig:housing_numNodes_plotMSEMeanTest}) and median values in test (fig. \ref{fig:housing_numNodes_plotMSEMedianTest}). A minimum MSE reduction of $10^{-6}$ was chosen for the four strategies. Also, a maximum number of nodes between 20 and 100 was used. These figures clearly show the reduction of MSE as the number of nodes and the complexity of the expressions grow. Also, it seems that the best results are obtained with the third and fourth strategies.

\begin{figure}
\centering
    \begin{subfigure}[b]{.49\linewidth}
        \centering
        \includegraphics[width=.99\textwidth]{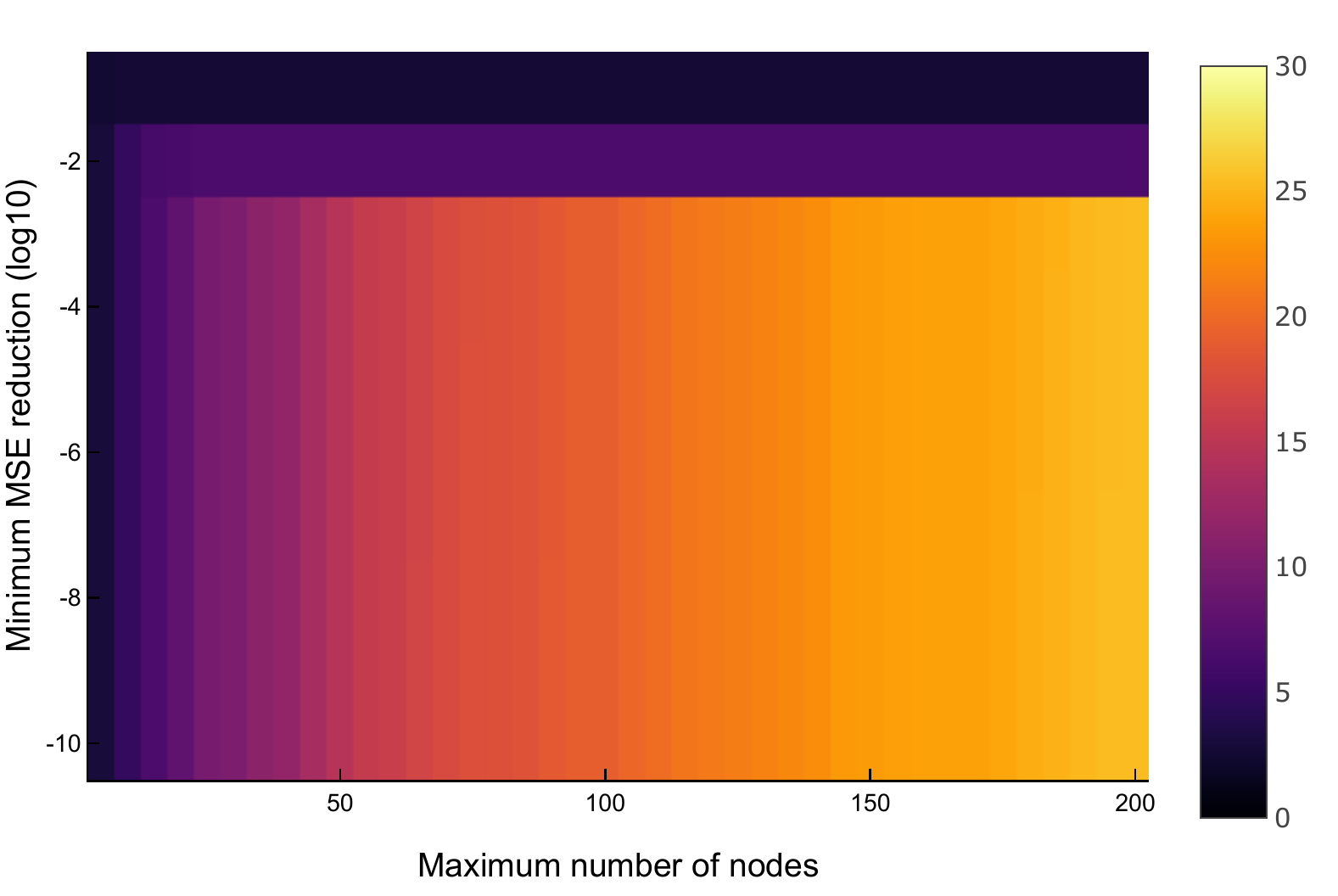}
        \caption{Strategy 1}
        \label{fig:housing_strategy1_numNodes_heatmapHeight}
    \end{subfigure}
    \begin{subfigure}[b]{.49\linewidth}
        \centering
        \includegraphics[width=.99\textwidth]{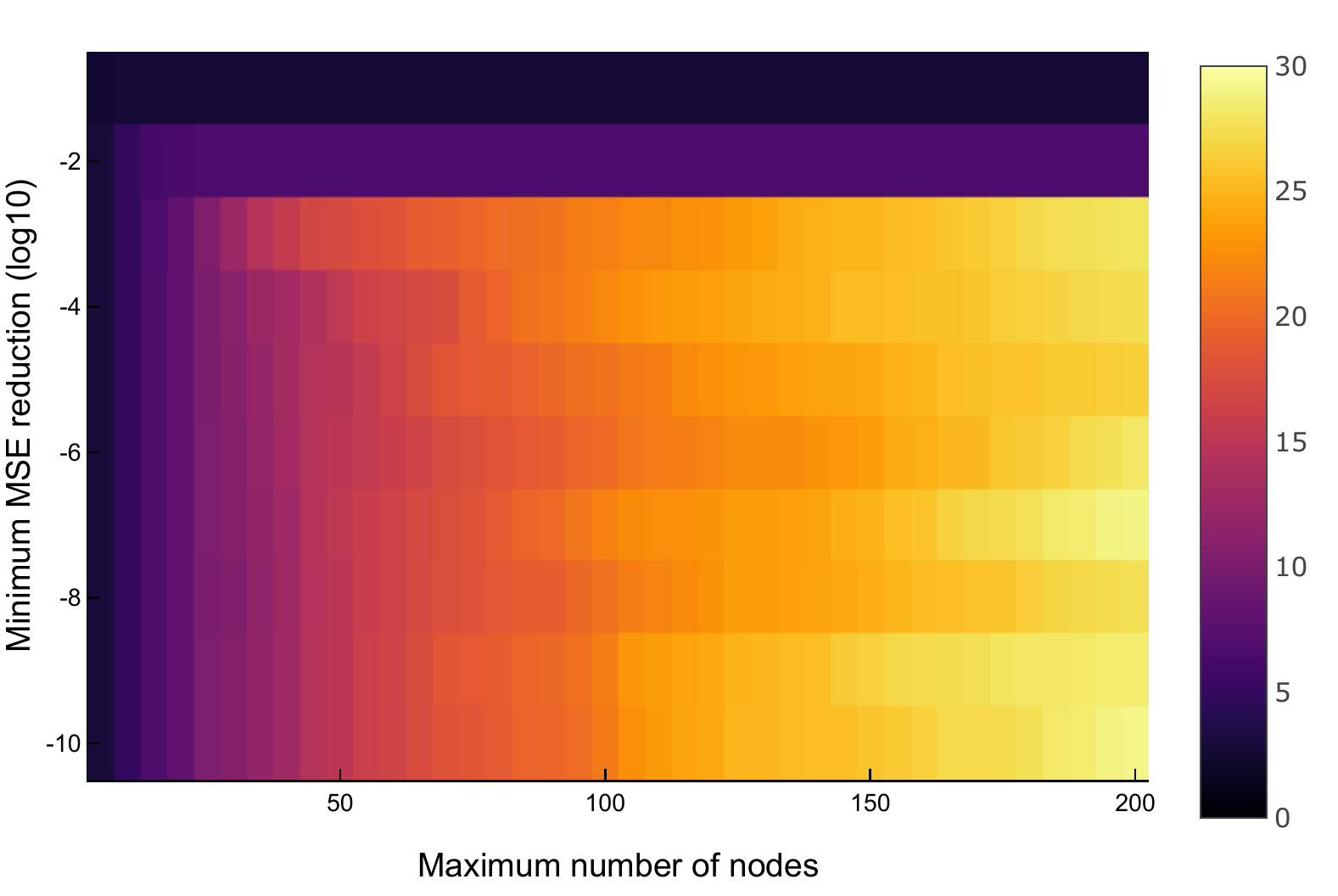}
        \caption{Strategy 2}
        \label{fig:housing_strategy2_numNodes_heatmapHeight}
    \end{subfigure}
    % To put more figures below, just add \\
    \\
    \begin{subfigure}[b]{.49\linewidth}
        \centering
        \includegraphics[width=.99\textwidth]{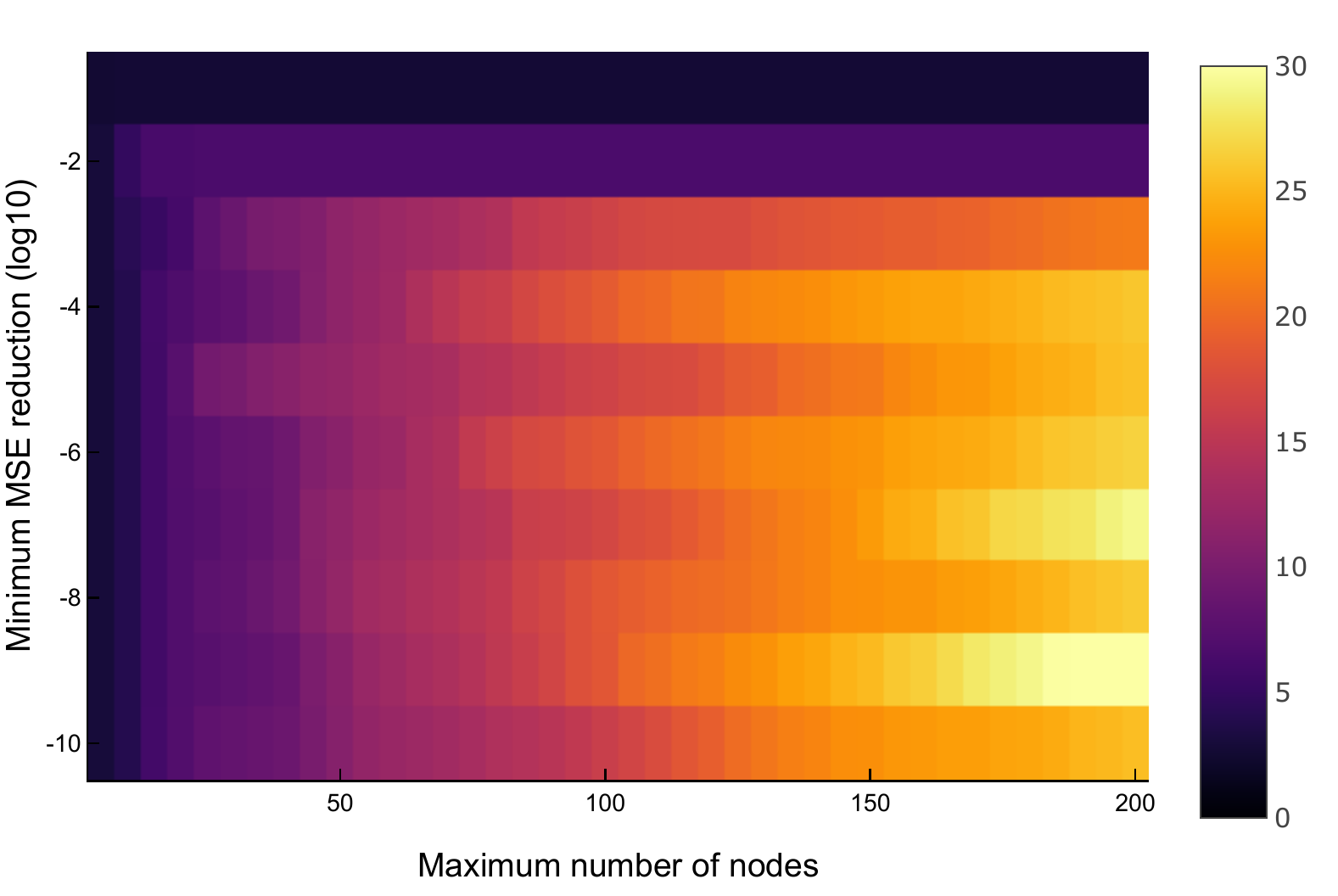}
        \caption{Strategy 3}
        \label{fig:housing_strategy3_numNodes_heatmapHeight}
    \end{subfigure}
    \begin{subfigure}[b]{.49\linewidth}
        \centering
        \includegraphics[width=.99\textwidth]{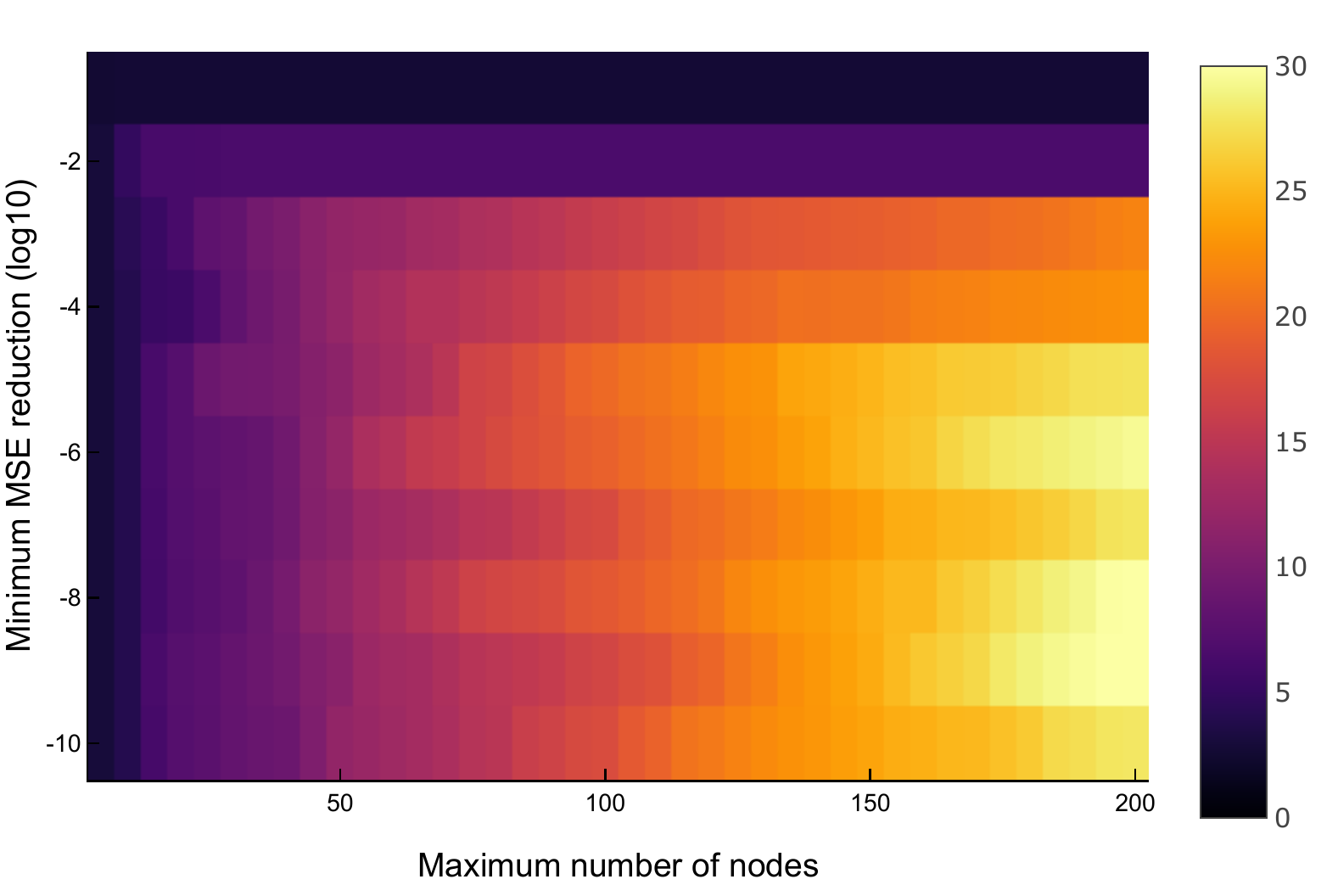}
        \caption{Strategy 4}
        \label{fig:housing_strategy4_numNodes_heatmapHeight}
    \end{subfigure}
    \caption{Averages heights for the housing problem}
    \label{fig:housing_numNodes_heatmapHeight}
\end{figure}

Another interesting information is obtained when examining the height of the resulting trees. Figure \ref{fig:housing_numNodes_heatmapHeight} shows 4 heatmaps, one for each strategy, with the average heights of the 10 trees obtained in the 10-fold for each hyperparameter configuration. As was expected, higher number of nodes leads to having higher heights. However, these heights are not in the scale of $log_2(n)$, being n the number of nodes. This shows that the system, with a limit of the number of nodes, finds better results with very unbalanced trees rather than with balanced trees. This fact supports the idea of not using an hyperparameter to limit the height of the tree. %, as was explored in section \ref{sec:experimentsHeight}.

\begin{figure}
\centering
    \begin{subfigure}[b]{.49\linewidth}
        \centering
        \includegraphics[width=.99\textwidth]{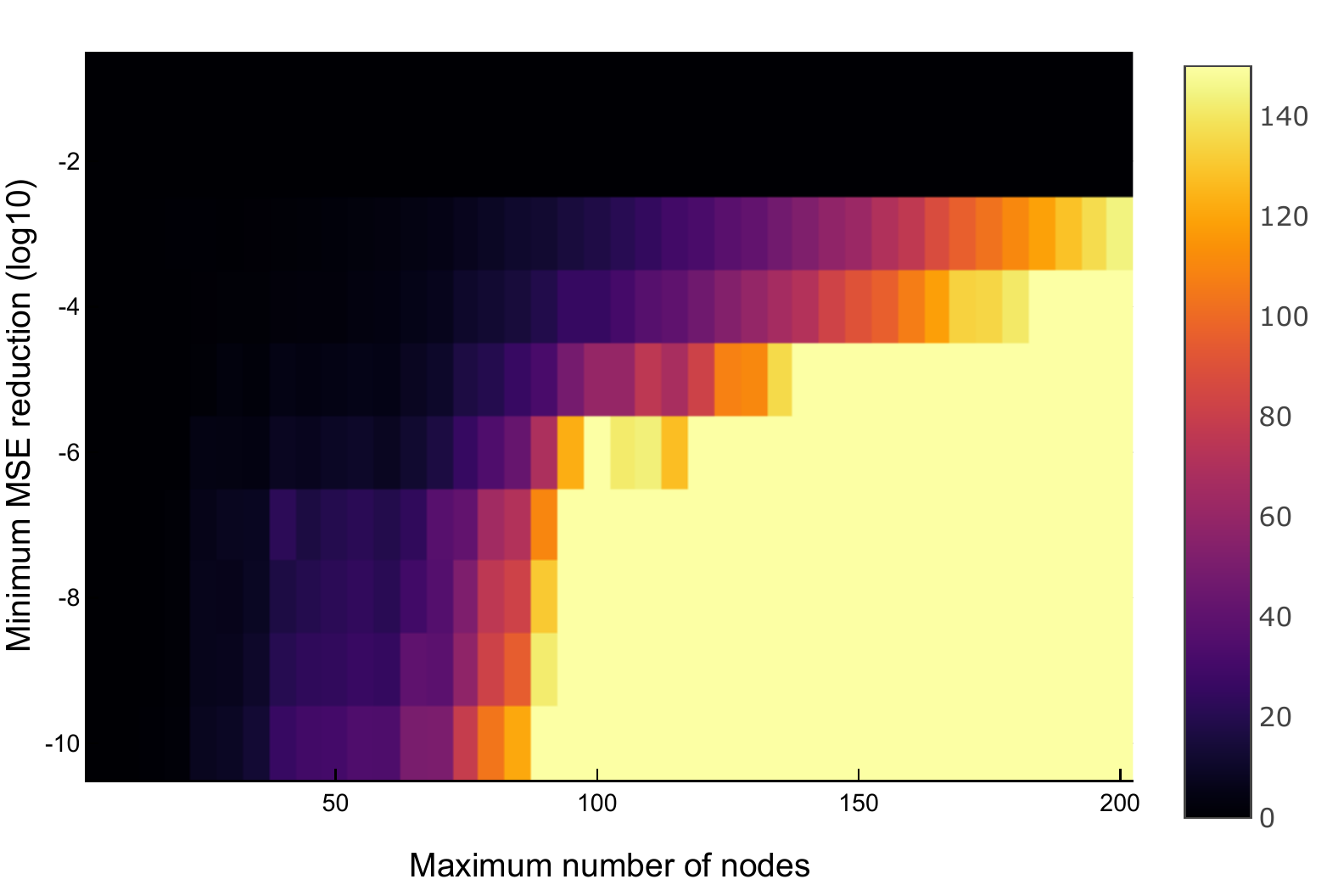}
        \caption{Strategy 1}
        \label{fig:housing_strategy1_numNodes_heatmapMeanTime}
    \end{subfigure}
    \begin{subfigure}[b]{.49\linewidth}
        \centering
        \includegraphics[width=.99\textwidth]{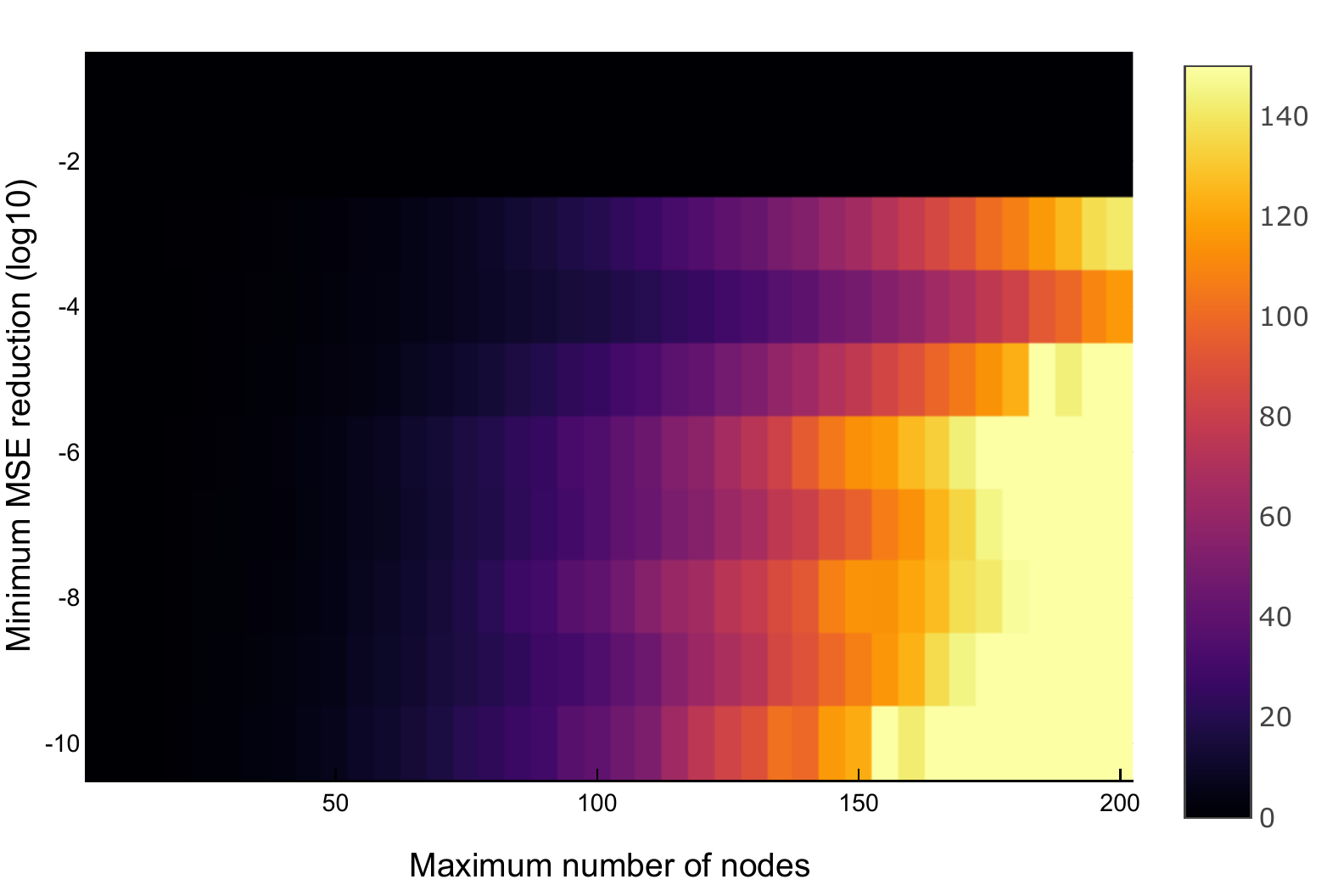}
        \caption{Strategy 2}
        \label{fig:housing_strategy2_numNodes_heatmapMeanTime}
    \end{subfigure}
    % To put more figures below, just add \\
    \\
    \begin{subfigure}[b]{.49\linewidth}
        \centering
        \includegraphics[width=.99\textwidth]{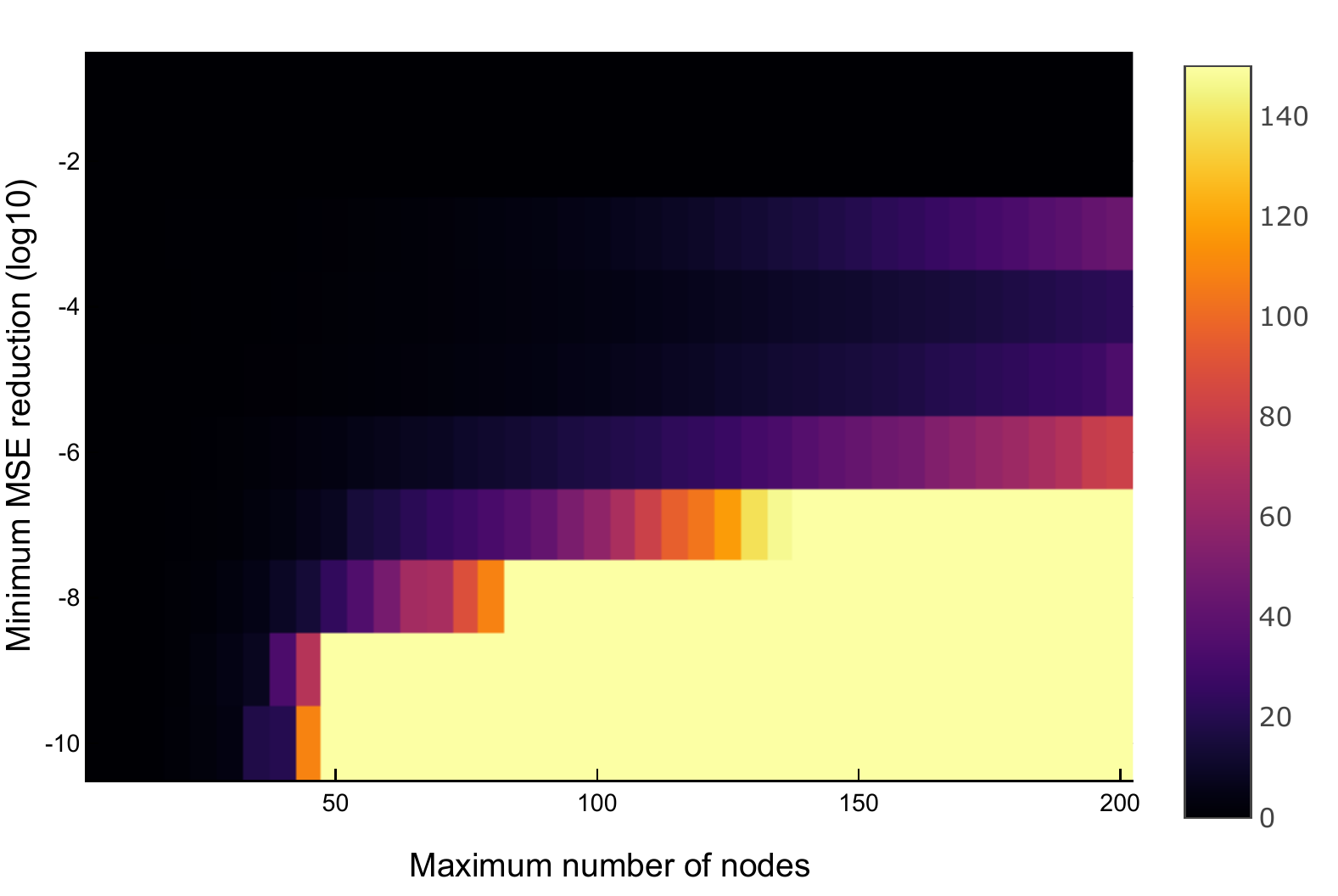}
        \caption{Strategy 3}
        \label{fig:housing_strategy3_numNodes_heatmapMeanTime}
    \end{subfigure}
    \begin{subfigure}[b]{.49\linewidth}
        \centering
        \includegraphics[width=.99\textwidth]{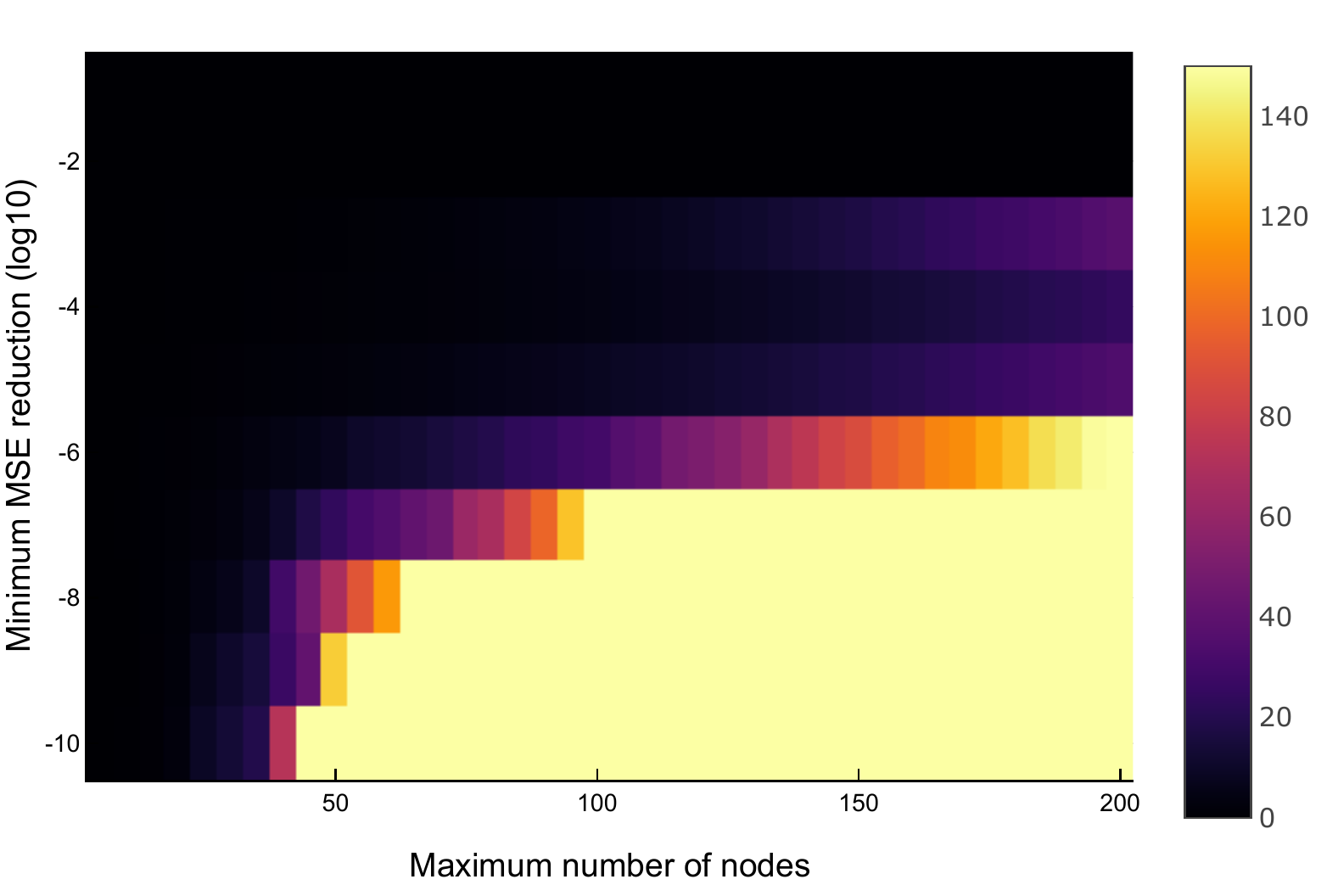}
        \caption{Strategy 4}
        \label{fig:housing_strategy4_numNodes_heatmapMeanTime}
    \end{subfigure}
    \caption{Average time for the housing problem}
    \label{fig:housing_numNodes_heatmapTime}
\end{figure}

\begin{figure}
    \centerline{\includegraphics[width=10cm, height=6cm]{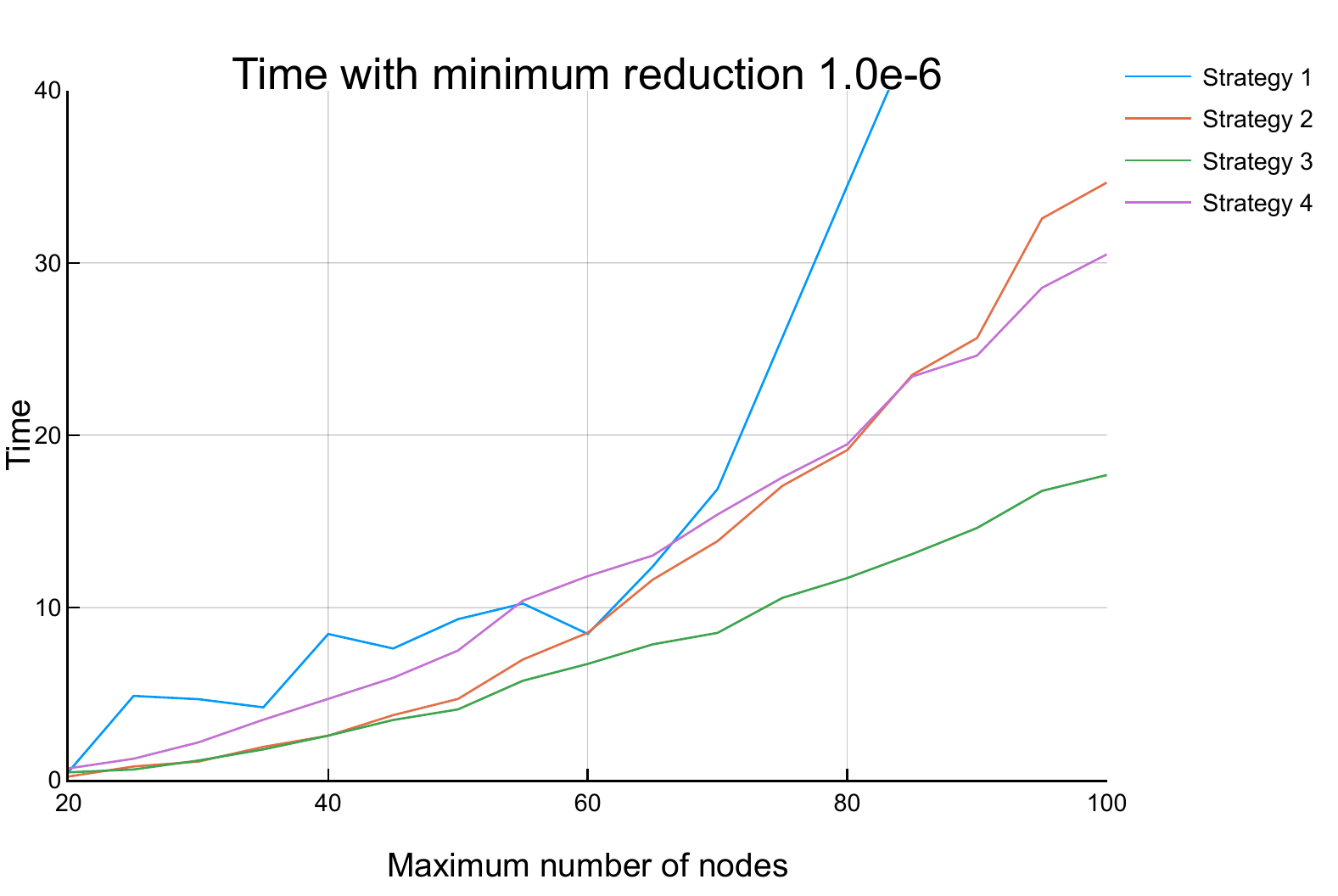}}
    \caption{Average computational time for different maximum number of nodes values and a minimum MSE reduction of $10^{-6}$}
    \label{fig:housing_numNodes_plotTime}
\end{figure}

Finally, figures in \ref{fig:housing_numNodes_heatmapTime} show, for each strategy, the computational time (in seconds) for each hyperparameter configuration. The results of this measures were cropped when the time was higher than 150 seconds. As was expected, as the trees are higher, the computational time is higher too, because a larger number of nodes have to be explored. Also, these figures show that the computational time increases with lower minimum MSE reduction. Therefore, for fast executions higher values of this parameter are preferred. With respect to the strategies, the third and fourth strategies are sensitively faster than the first and second. In comparison with figure \ref{fig:housing_numNodes_plotMSE}, figure \ref{fig:housing_numNodes_plotTime} shows the computational time for the same hyperparameters values. This graph shows the increase of time as the number of nodes is increased, and the difference in the speed of each strategy.

\begin{figure}
    \centerline{\includegraphics[width=10cm, height=6cm]{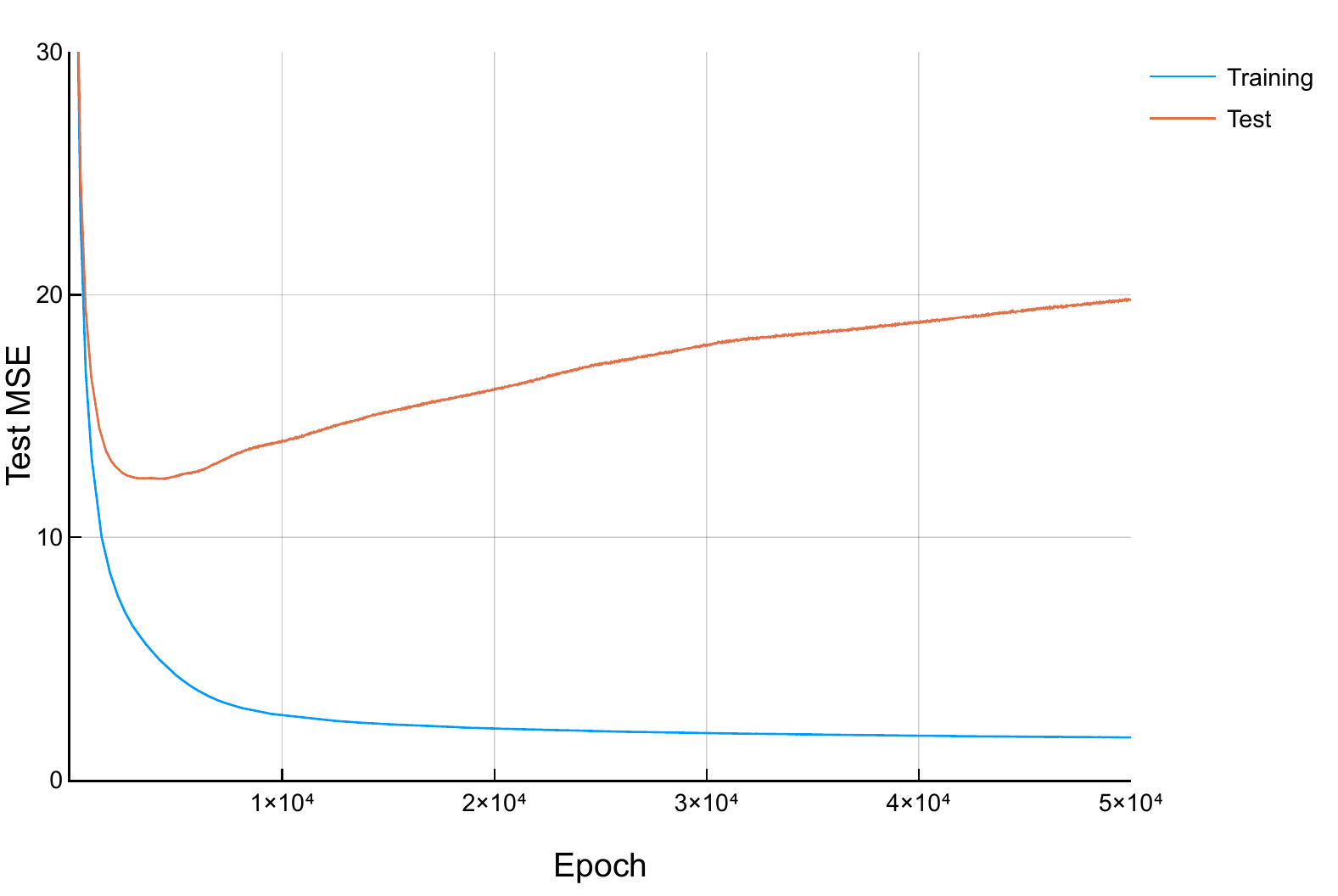}}
    \caption{Results of the MLP training process}
    \label{fig:housing_MLP}
\end{figure}

\subsection{Discussion}

\label{sec:discussion}

As it can be seen on these graphs, the system is able to return good results in training and test. An interesting feature is that the number of nodes needed to obtain these results is not very high (around 100 nodes), as opposed to other systems like GSGP, which returns much more complex expressions. Table \ref{housing_summaryTraining} shows a summary of the configurations that returned the best training results for each strategy. As was expected, the best training results were obtained with the most complex trees (highest number of nodes). This table also shows the test results returned by these configurations. Even the training results show a great ability to fit the training set, these configurations did not return the best results in test. Table \ref{housing_summaryTest} shows the same information, but this time for the configurations that returned the best results in the test set.

However, in order to correctly measure the generalization ability of this system, it has been compared with one of the most popular Machine Learning systems: Artificial Neural Networks (ANNs). An ANN with two hidden layers, with 10 neurons in the first hidden layer and 3 neurons in the second hidden layer was used. This topology was chosen to be complex enough in order to allow the ANN to fit the training data at least as good as this system, and compare the generalization abilities of both systems. An ADAM optimizer was used, with a learning rate of 0.01. The same 10-fold training/test sets were used for the training of the ANN. However, since this process is not deterministic, for each set the training process was repeated 50 times and averaged. In each training, 50000 epochs were done. Figure \ref{fig:housing_MLP} shows the average training process for the 10 folds. This graph shows the training and test results for each epoch. As it is usual, while the training results are still improving, the test results reach a point in which begin to worsen. Finally, the training process is stopped when the maximum number of epochs is reached. In this point, the ANN has overfit the training set, with a bad result in the test set. Table \ref{housing_summaryTest} shows a summary of the best results in the training set, reached at epoch 49926, and in the test set, at epoch 4347. As it can be seen, the training result is much lower than the ones obtained with the system described in this paper, because a complex topology was used, leading to having a complex system. However, as table \ref{housing_summaryTraining} suggests, better training results could be obtained with a larger number of nodes.

The test results of table \ref{housing_summaryMLP} can be compared with the ones shown in table \ref{housing_summaryTest}. To perform this comparison, a paired two-tailored sample t-test was used between the best test results of each strategy and the best test results of the ANN. In each of this four tests, a significance level of $\alpha=0.05$ was used. As result, all of the four tests returned that the mean results are not significantly different. Therefore, the system described in this paper is able to return test results as good as an ANN.

Another important hyperparameter was analysed: the minimum MSE improvement. From the resulting graphs, it is clear that this parameter should not take very high values ($10^{-1}$). However, there is no clear relation between this value and the obtained results measured in MSE. However, lower values in this parameter make computational time much longer. Joining these two considerations, values not very low ($10^{-10}$) and not very low ($10^{-1}$) should be given to this parameter. Since computational time increases with lower minimum MSE reductions, but MSE in test does not seem to significatively improve, leaving a higher value to this value seems to be a good choice. For this reason, the selected value was $10^{-6}$ for the four strategies.

Regarding the four strategies used for the experiments, the first one returns the worst results, while being also the one that takes the highest computational time. Therefore, performing all of the searches in all of the nodes does not seem to be a good choice, since some of these searches are unlikely to be successful in many nodes. Strategy 2 returns better results than strategy 1, which supports the idea that it is convenient to optimize the constants of the tree before making any modification to its structure. However, this strategy still takes too much time, because in each iteration all of the searches are performed on all of the nodes. Finally, strategies 3 and 4 return the best results. As figure \ref{fig:housing_numNodes_plotTime} shows, those searches that use the constant optimization process (searches 2 and 3) take less time to perform the training than their counterparts without constant optimization (searched 1 and 4 respectively).

Regarding the height of the trees, the graph with the tree height shows that the resulting trees are not balanced, and therefore setting a tree height limit would not be a good alternative.

%%%%%%%%%%%%%%%%%%%%%%%%%%%%%%%%%%%%%%%%%%%%%%%%%%%%%%%%%%%%%%%%%%%%%%%%%%%%%%%%

\section{Conclusions}

\label{sec:seccionConclusions}

This work presents a novel technique for Symbolic Regression. In this field, the most used technique is GP, which is based on evolutionary processes. Thus, this field had an important lack of mathematical-based methods. The technique presented in this work allows the obtaining of mathematical expressions that can model an input-output relationship.

Also, the expressions obtained by this method can have a limit in complexity. This allows the obtaining of expressions that can be easily analysed by humans, in contrast with other techniques such as GSGP, that return very large expressions. The analysis of these expressions is usually one of the objectives of Symbolic Regression.

Results on section \ref{sec:Experiments} show that this technique can return good results in real-world problems. The results in the configurations with a high number of nodes show a very small MSE. This shows the capacity of the system. However, for generalization purposes, setting a limit on the complexity allows the obtaining expressions with good generalization. Also, the computational time has been measured, and this system has shown to return good results in short time. The strategy that returned the lowest MSE in test and fastest results only took some seconds to build the tree.

An additional advantage of this system used for Machine Learning purposes is that the returned model is an standard equation and thus it can be used in any programming language with no need to import any ML library. Moreover, this expression can be used in any system other than programming environments. For instance, it can be easily used in a spreadsheet as opposed to other systems such as Neural Networks.

%%%%%%%%%%%%%%%%%%%%%%%%%%%%%%%%%%%%%%%%%%%%%%%%%%%%%%%%%%%%%%%%%%%%%%%%%%%%%%%%

\section{Future Works}
\label{sec:seccionFutureWorks}

This work opens a wide new research field in Symbolic Regression. As was described throughout in the paper, many research works are still to be done by the research community. Some of the possible developments could be:

\begin{itemize}

    \item In the constant search, find an easy expression or method to compute the minimum of equation \ref{eq:eq5} in the case when $c_i$ and $d_i$ are vectors. One possibility could be to use gradient descent \cite{snyman2018practical} to minimize this function.
    
    \item Perform a comparison of different search strategies in different problems to find a strategy that behaves better in most of them.

    \item A new constant optimization algorithm could be proposed. In the one described in this paper, each constant is alternatively optimised in different iterations, which needs the computation of the equations of the corresponding nodes in each iteration. An alternative could be to change the value of several (or all) constants in each iteration, thus making this process faster.

    \item As the limit of the complexity of the tree has been proven to be an important factor, new ways of limiting this complexity can be found. For instance, setting a limit to the number of sum, minus, multiplication or division operations that can be used.

    \item Limiting the complexity of the models is a common way to avoid overfitting. However, other methods such as the use of a validation set can also be used. This possibility could be explored, with the advantage of not needing to set the limit of number of neurons or height of the tree.
    
    \item In order to obtain expressions even easier to be understandable by humans, information about the structure of the desired expressions could be give. For instance, many times the desired expression is a division of two expressions, with no other division performed in these two parts. This structure, as well as any other could be given to the system. This could have as additional feature that the search process would be sped up.
    
    \item An interesting possibility could be to extend the variable search not only with variables but also with any subtrees. Those parts of the tree that are going to be replaced with another could be stored in a structure like a "node pool" and be used in the search later. The idea is that if once they were useful, they might become useful again later, when the tree is modified.
    
\end{itemize}

Also, this work could also be a basis to new ways of combining models not being necessarily mathematical expressions. Any model (for instance, an ANN) can be represented in the semantic space as one point, that could fall inside one of the shapes. Therefore, the tree being developed could combine mathematical expressions with other type of models. This could be easily done with a search similar to the variable search, that could be called "model search", and a "constant-model search" In this sense, this technique would allow the building of ensembles of models.

\section*{Acknowledgments}

The experiments described in this work were performed on computers in the Supercomputing Center of Galicia (CESGA). Daniel Rivero and Enrique Fernández-Blanco would also like to thank the support provided by the NVIDIA Research Grants Program.

\section*{Disclosure statement}
No potential conflict of interest was reported by the authors

\bibliographystyle{unsrt}
\bibliography{references} 

\end{document}